\def\year{2020}\relax
\documentclass[letterpaper]{article} %
\usepackage{aaai20}  %
\usepackage{times}  %
\usepackage{helvet} %
\usepackage{courier}  %
\usepackage[hyphens]{url}  %
\usepackage{graphicx} %
\usepackage{graphicx}  %
\usepackage{subfig}
\usepackage{amsmath}
\usepackage{algorithm}
\usepackage[noend]{algpseudocode}
\usepackage{nicefrac}       %
\usepackage{booktabs}       %
\usepackage{amssymb}
\usepackage{multirow}
\usepackage{pbox}

\usepackage{color}
\definecolor{fullred}{rgb}{0.95,.0,.1}
\newcounter{cmt}

\title{Particle Filter Recurrent Neural Networks}

\author{Xiao Ma\thanks{equal contribution}, Peter Karkus$^*$, David Hsu, Wee Sun Lee\\
National University of Singapore\\
\{xiao-ma, karkus, dyhsu, leews\}@comp.nus.edu.sg}
 
\begin{document}

\maketitle

\begin{abstract}
	Recurrent neural networks (RNNs) have been extraordinarily successful for prediction with sequential data.  To tackle highly variable and multi-modal real-world data, we introduce  \textit{Particle Filter Recurrent Neural Networks} (PF-RNNs), a new RNN family that  explicitly models uncertainty in its internal structure: while an  RNN relies on a  long, deterministic latent state vector, a PF-RNN maintains a latent \textit{state distribution}, approximated as a set of particles. For effective learning, we provide a fully differentiable particle filter algorithm that updates the PF-RNN latent state distribution according to the Bayes rule.  Experiments  demonstrate that the proposed PF-RNNs outperform the corresponding standard gated RNNs on a synthetic robot localization dataset and 10 real-world sequence prediction datasets for text classification, stock price prediction, etc. %
	
\end{abstract}

\section{Introduction}

Prediction with sequential data is a long-standing challenge in machine learning. It has many applications, e.g.,  object tracking~\cite{blake1997condensation},
speech recognition \cite{xiong2018microsoft},
and decision making under uncertainty~\cite{somani2013despot}. For effective prediction, predictors require ``memory'', which summarizes and tracks information in the input sequence. The memory state is generally not observable, hence the need for a \emph{belief}, {i.e.}, a posterior state distribution that captures the sufficient statistic of the input for making predictions.  Modeling the belief manually is often difficult. Consider the task of classifying news text---treated as a sequence of words---into categories, such as politics, education, economy, etc. %
It is difficult to handcraft the belief representation and dynamics for accurate classification.

State-of-the-art sequence predictors often use recurrent neural networks (RNNs), which \textit{learn} a vector $h$ of deterministic time-dependent latent variables as an approximation to the belief. Real-world data, however, are highly variable and often multi-modal. To cope with the complexity of uncertain real-world data and achieve better belief approximation, one could increase the length of latent vector $h$, thus increasing the number of network parameters and the amount of data required for training.

\begin{figure*}[t]
	\centering
	\begin{tabular}{c c c}
		\includegraphics[width=0.4\linewidth]{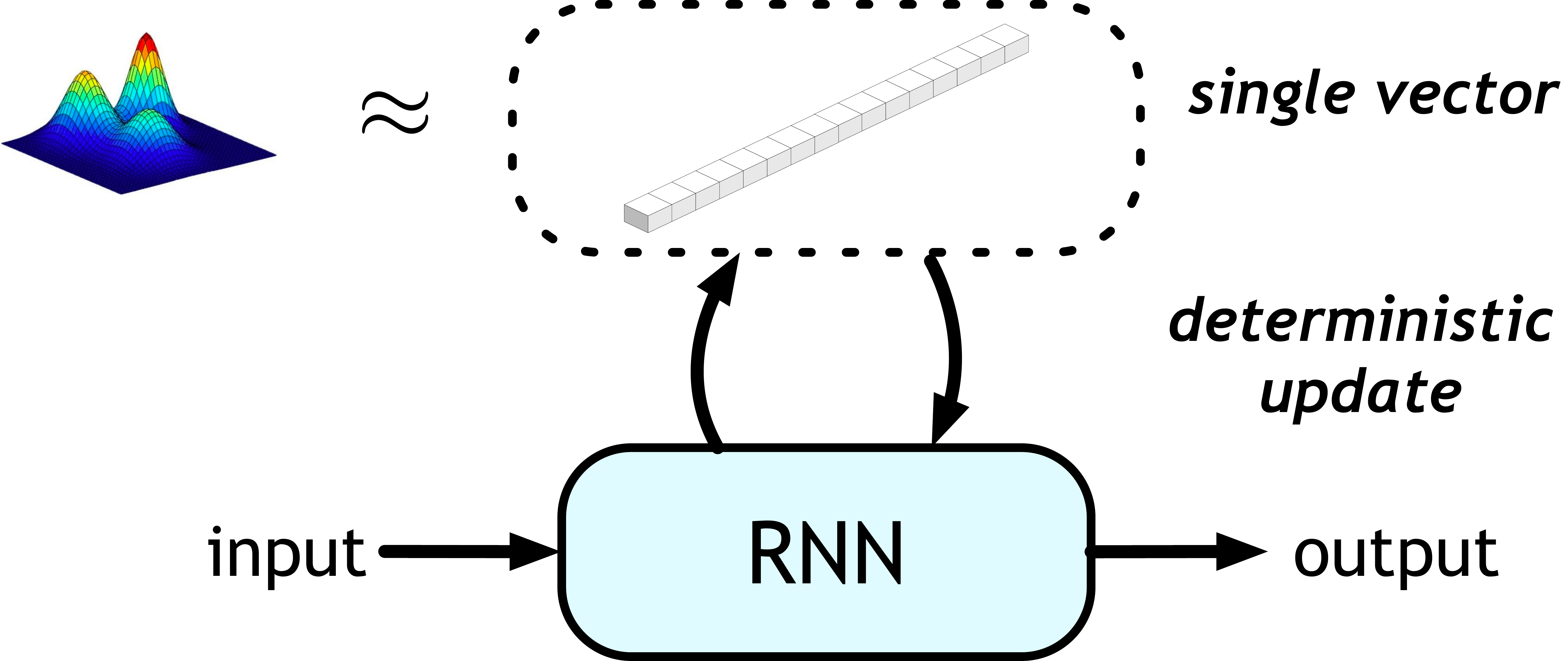} &
		 &
		\includegraphics[width=0.4\linewidth]{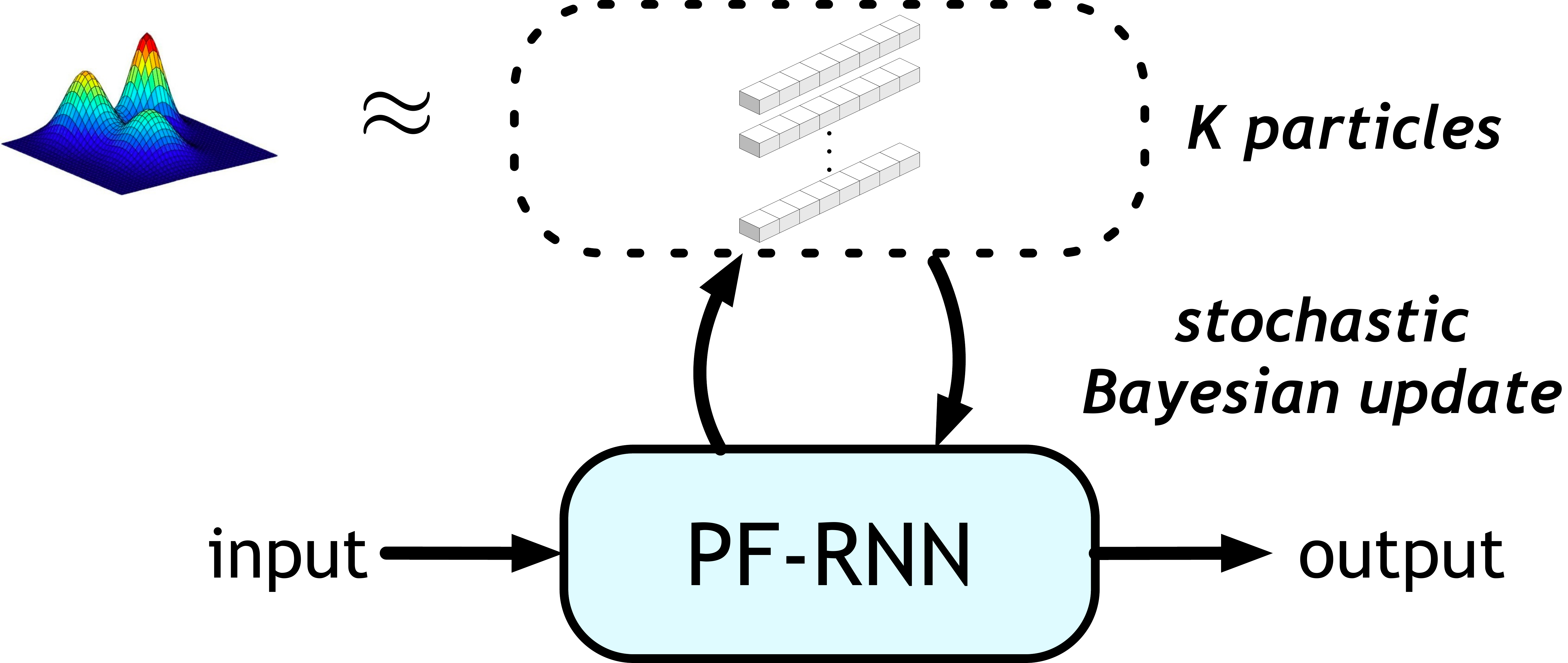}
	\end{tabular}
	\centering
	\caption{\textbf{A comparison of RNN and PF-RNN}. An RNN approximates the belief as a long latent vector and updates it with a deterministic nonlinear function. A PF-RNN approximates the belief as a set of weighted  particles and updates them with the stochastic particle filtering algorithm.}
	\label{fig:fig1_comparison}
\end{figure*}

We introduce  \textit{Particle Filter Recurrent Neural Networks} (PF-RNNs), a new family of RNNs that seeks to improve belief approximation without lengthening the latent vector $h$,  thus reducing the data required for learning.
\textit{Particle filtering}~\cite{del1996non} is a model-based belief tracking algorithm. It approximates the belief as a set of sampled states that typically have well-understood meaning. PF-RNNs borrow from particle filtering the idea of approximating the belief as a set of weighted particles, and combine it with the powerful approximation capacity of RNNs.
PF-RNN approximates the variable and multi-modal belief as a set of weighted latent vectors $\{h^1, h^2, \ldots\}$ sampled from the same distribution. 
Like standard RNNs, PF-RNNs follow a model-free approach: PF-RNNs' latent vectors are learned distributed representations, which are not necessarily interpretable.   
As an alternative to the Gaussian based filters, e.g., Kalman filters, particle filtering is a non-parametric approximator that offers a more flexible belief representation~\cite{del1996non}; it is also proven to give a tighter \textit{evidence lower bound} (ELBO) in the data generation domain~\cite{burda2015importance}.
In our case, the approximate representation is trained from data to optimize the prediction performance. For effective training with gradient methods, we employ a fully differentiable particle filter algorithm that maintains the latent belief. 
See Fig.~\ref{fig:fig1_comparison} for a  comparison
of RNN and PF-RNN.

We apply the underlying idea of PF-RNN to gated RNNs, which are easy to implement and have shown strong performance in many sequence prediction tasks.
Specifically, we propose PF-LSTM and PF-GRU, the particle filter extensions of  Long Short Term Memory (LSTM)~\cite{hochreiter1997long} and Gated Recurrent Unit (GRU)~\cite{cho2014learning}. PF-LSTM and PF-GRU 
serve as drop-in replacements for LSTM and GRU, respectively. They aim to learn a better belief representation from the same data, 
though at a greater computational cost.

We evaluated  PF-LSTM and  PF-GRU on 13 data sets: 3~synthetic dataset for systematic understanding and 10 real-world datasets with different sample sizes for performance comparison. 
The experiments show that our  PF-RNNs outperform the corresponding standard RNNs with a comparable number of parameters. Further, the PF-RNNs achieve the best results on almost all datasets when there is no restriction on the number of model parameters used.\footnote{The code is available at https://github.com/Yusufma03/pfrnns}
\section{Related Work}
There are two general categories for prediction with sequential data: model-based and model-free.
The model-based approach includes, e.g.,  the well-known hidden Markov models (HMMs) and the dynamic Bayesian networks (DBNs) \cite{murphy2002dynamic}.   
They rely on handcrafted state representations with well-defined semantics, e.g., phonemes in speech recognition.
Given a model, one may perform belief tracking according to the Bayes' rule. The main difficulty here is the state space and  the computational complexity of belief tracking grow exponentially with the number of state dimensions. 
To cope with this difficulty, particle filters represent the belief as a set of sampled states and perform approximate inference.
Alternatively, the model-free approach, such as RNNs, approximates the belief as a latent state vector,  learned directly from data, and updates it through a deterministic nonlinear function, also learned from data.

The proposed PF-RNNs build upon RNNs and combine their powerful data-driven approximation capabilities with the sample-based belief representation and approximate Bayesian inference used in particle filters. 
Related sample-based methods have been applied to generative models.
Importance sampling is used to improve variational auto-encoders \cite{burda2015importance}. This is extended to sequence generation \cite{le2017auto}
and to  reinforcement learning  \cite{igl2018deep}. 
Unlike the earlier works that focus on generation, we combine  RNNs and particle filtering for sequence prediction.
PF-RNNs are trained \textit{discriminatively}, instead of generatively, with the target loss function on the model output. 
As a result,  PF-RNN training prioritizes target prediction over data generation which may be irrelevant to the prediction task. 

PF-RNNs exploit the general idea of embedding algorithmic priors~\cite{karkus2019differentiable}, in this case, filtering algorithms, in neural networks and train them discriminatively~\cite{jonschkowski2016end,karkus2018particle}. 
Earlier work embeds a particle filter in an RNN for learning belief tracking, but follows a model-based approach and relies on handcrafted belief representation~\cite{jonschkowski2018differentiable,karkus2018particle}.
PF-RNNs retain the model-free nature of RNNs and exploit their powerful approximation capabilities to learn belief representation directly from data. Other work explicitly addresses belief representation learning with RNNs~\cite{gregor2018temporal,guo2018neural}; however, they do not involve Bayesian belief update or particle filtering. 

\section{Particle Filter Recurrent Neural Networks}

\subsection{Overview}
The general sequence prediction problem is to predict an output sequence, given an input sequence. 
In this paper,  we focus on predicting the output $y_t$ at   time $t$, given the input history $x_1,x_2,\ldots,x_t$. 

Standard RNNs handle sequence prediction by maintaining a deterministic latent state $h_t$ that captures the sufficient statistic of the input history, and updating $h_t$ sequentially given new inputs. Specifically, RNNs update $h_t$ with a  deterministic nonlinear function learned from data.
The predicted output $\hat{y}_t$  is another nonlinear function of the latent state $h_t$, also learned from data.

To handle highly variable, noisy real-world data, one key idea of PF-RNN is to capture the sufficient statistic of the input history in the latent \textit{belief}  $b(h_t)$ by forming multiple hypotheses over $h_t$. 
Specifically, PF-RNNs approximate $b(h_t)$ by a set of $K$ weighted particles $\{ (h_t^i,w_t^i) \}_{i=1}^K$, for latent state $h_t^i$ and weight $w_t^i$;  the particle filtering algorithm is used to  update the particles according to the Bayes rule.
The Bayesian treatment of the latent belief naturally captures the stochastic nature of real-world data.
Further, all particles share the same parameters in a PF-RNN.   The number of particles thus does not affect the number of PF-RNN network parameters. 
Given a fixed training data set, we expect that increasing the number of particles improves the belief approximation and leads to better learning performance,  but at the cost of greater computational complexity. 

Similar to RNNs, we used learned functions for updating latent states and predict the output $y_t$ based on the averaged particle: $\hat{y}_t = f_\mathrm{out}(\bar{h}_t)$,
where $\bar{h}_t = \sum_{i=1}^K w_t^i h_t^i$ and $f_\mathrm{out}$ is a  task-dependent prediction function.
For example, in a robot localization task, $f_\mathrm{out}$ maps $\bar{h}_t$ to a robot position. In a classification task, $f_\mathrm{out}$ maps $\hat{y}_t$ to a vector of class~probabilities.

For learning, we embed a particle filter algorithm in the network architecture and train the network \textrm{discriminatively} end-to-end. Discriminative training aims for the best prediction performance within the architectural constraints imposed by the particle filter algorithm. 
We explore two  loss functions for learning: 
one minimizes the error of prediction with the mean particle and the other maximizes the log-likelihood of
the target $y_t$ given the full particle distribution.
Combining the two loss functions gives the best performance empirically.

\subsection{Particle Filter Extension to RNNs}\label{sec:extension}

Extending an RNN to the corresponding PF-RNN requires  the latent particle belief representation $\{ (h_t^i,w_t^i) \}_{i=1}^K$ and  the  associated belief update function. 
The new belief  is a posterior distribution conditioned on the previous belief  and the new input.  We model the PF-RNN's latent dynamics as a controlled system, with the uncontrolled system as a special case. The standard belief update consists of two steps:  the transition update  $\tilde{b}_t = f_\mathrm{tr}(b_{t-1}, u_t)$ for control~$u_t$ and the observation update  $b_t = f_\mathrm{obs}(\tilde{b}_t, o_t)$ for observation $o_t$. However,  $u_t$ and $o_t$ are not separated a priori in general sequence prediction problems.
PF-RNNs use input $x_t$ in both $f_\mathrm{tr}$ and $f_\mathrm{obs}$, and learn to extract latent control $u_t(x_t)$ and observation $o_t(x_t)$ from $x_t$ through task-oriented discriminative training.
This approach is more flexible and provides a richer function approximation class.  
\begin{figure}[!htb]
	\centering
	\includegraphics[width=0.5\linewidth]{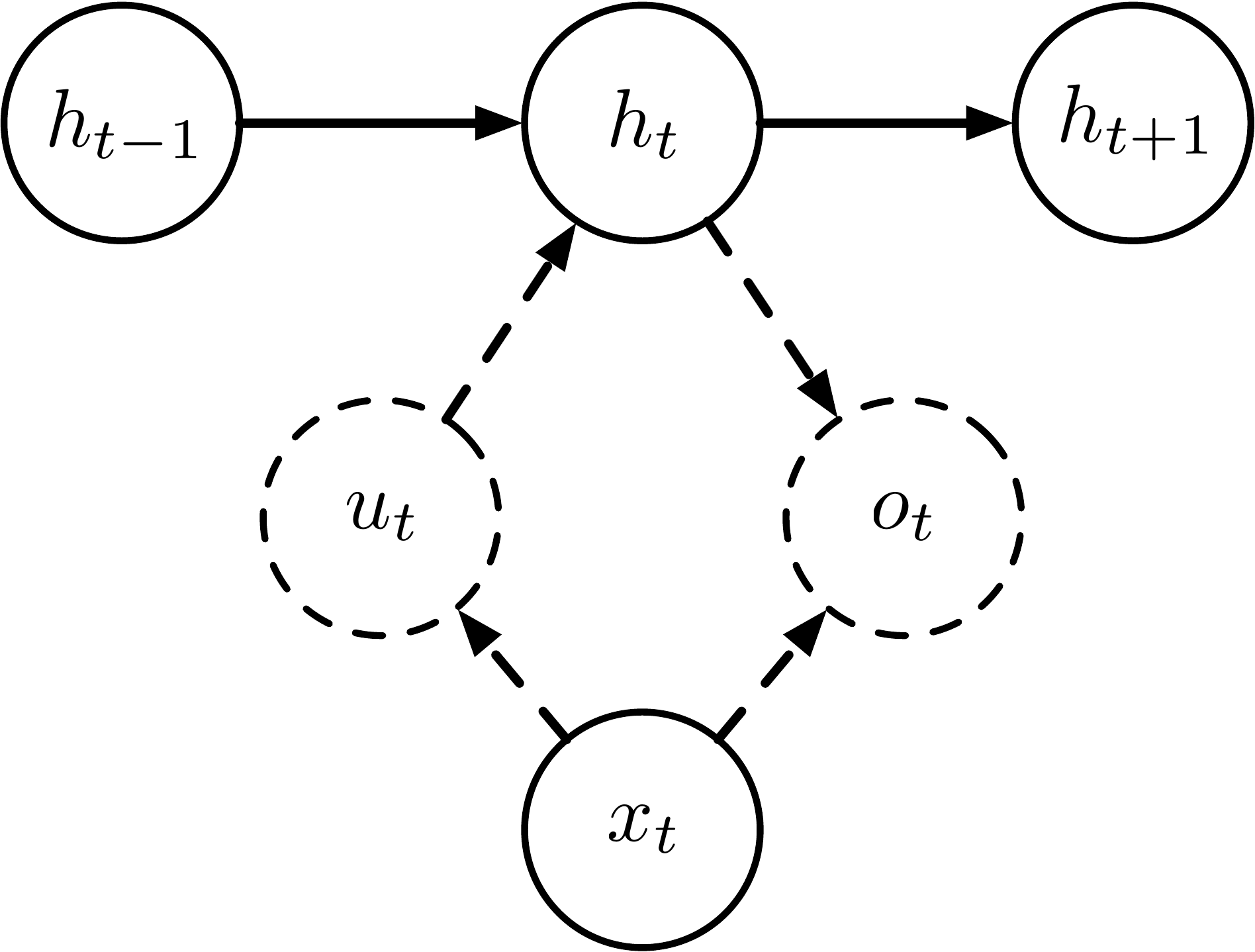}
	\centering
	\caption{%
		\textbf{The graphical model for PF-RNN belief  update}. The dashed circles indicate variables not explicitly defined in our models. 
	}
	\label{fig:diagram}
\end{figure}

\textbf{Stochastic memory update.} 
We apply a learned transition function $f_\mathrm{tr}$ to  each particle state $h_t^i$:
\begin{equation}
	h_t^i = f_\mathrm{tr}(h_{t-1}^i, u_t(x_t), \xi_t^i), \quad\xi_t^i\sim p(\xi_t^i | h_{t-1}^i, u_t(x_t))\label{eqn:1}
\end{equation}
where  $x_t$ is the input and $\xi_t^i$ is a learned noise term. We assume $p(\xi_t^i | h_{t-1}^i, u_t(x_t))$ to be a Gaussian distribution
and use the \emph{reparameterization trick}~\cite{kingma2013auto}
to the function to simulate the transition in a differentiable manner. 
From the RNN perspective,  $\xi_t^i$  captures the stochasticity in the latent dynamics. From the particle filtering perspective,  $\xi_t^i$ increases particle diversity, relieving the issue of particle depletion after resampling.  %

\textbf{Particle weight update.} We update the weight $w_t^i$ of the particle $h_t^i$   recursively in a Bayesian manner, using the likelihood $p(o_t|h_{t}^i)$. Instead of  modeling $p(o_t | h_t^i)$ as a  generative distribution, we approximate it directly as a learned function $f_\mathrm{obs}(o_t(x_t), h_t^i)$ with $o_t(x_t)$ and $h_t^i$ as inputs, and have 
\begin{equation}
	w_t^i = \eta f_\mathrm{obs}(o_t(x_t), h_t^i) w_{t-1}^i,
\end{equation}
where $\eta$ is a normalization factor.
Our observation model $f_\mathrm{obs}$, coupled with discriminative training, significantly improves the performance of PF-RNNs in practice. The generative distribution $p(o_t|h_t^i)$ is parameterized for generating $o_t$ and may contain irrelevant features for the actual objective of  predicting $y_t$. In contrast, $f_\mathrm{obs}$ skips  generative modeling 
and learns only features useful for predicting $y_t$, as PF-RNNs are trained discriminatively to optimize prediction accuracy.
\begin{figure*}[t]
	\centering
	\begin{tabular}{c c}
		\small PF-LSTM & \small PF-GRU\\
		\includegraphics[width=0.48\linewidth]{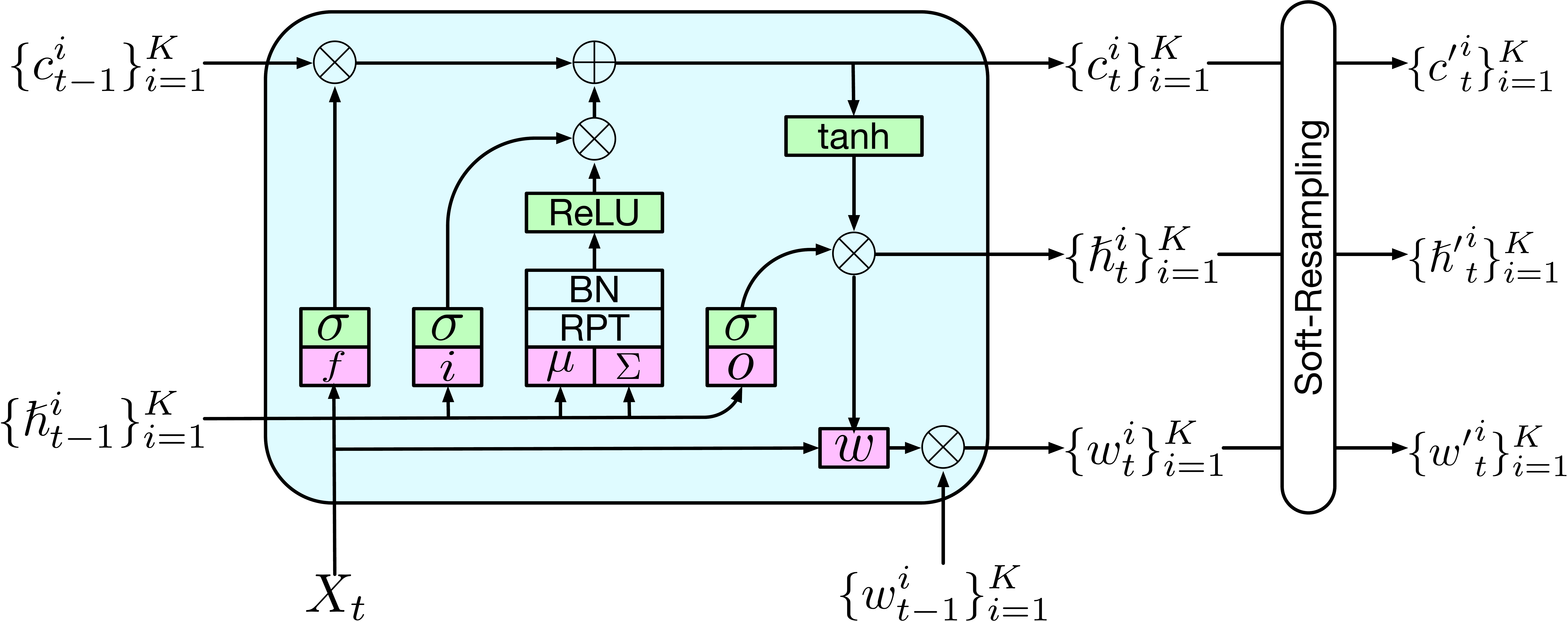} &
		\includegraphics[width=0.48\linewidth]{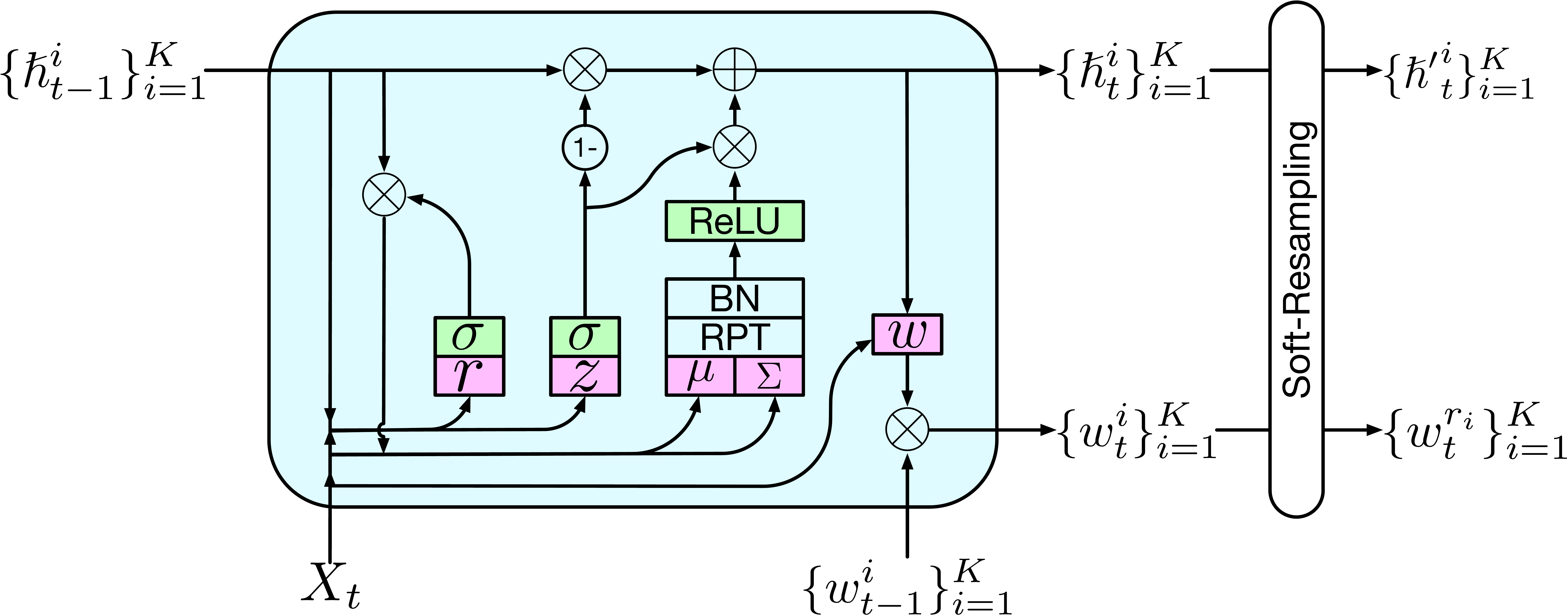}
	\end{tabular}
	\centering
	\caption{\textbf{ PF-LSTM and PF-GRU network architecture}. Notation (1)~green box: activation (2)~pink box: learned function (3)~\textit{RPT}: reparameterization trick (4)~\textit{BN}: batch normalization.}
	\label{fig:nets}
\end{figure*}

\textbf{Soft resampling.} In particle filter algorithms, resampling is required to avoid particle degeneracy, i.e., most particles having near-zero weights. 
Resampling constructs a new particle set $\{ (h_t'^j,w_t'^j) \}_{j=1}^K$ with constant weights $w_t'^j = 1/K$. Each new particle $h_{t}'^j$ takes the value of an \emph{ancestor} particle, $h_t^{a^j}$, where the ancestor index $a^j$ is sampled from a multinomial distribution defined by the original particle weights, $a^j \sim p$,
where $p$ is a multinomial distribution with $p(i) = w_t^i$.

However, resampling is not differentiable, which prevents the use of backpropagation for training PF-RNNs. 
To make our latent belief update differentiable, we use  \emph{soft resampling}~\cite{karkus2018particle}. Instead of resampling particles according to %
$p$, we resample from $q$, a mixture of $p$ and a uniform distribution:
$
q(i) = \alpha w_t^i + (1 - \alpha) (1/ K)
$, for 
$\alpha \in (0,1]$.
The new weights are computed according the importance sampling formula, which leads to an unbiased estimate of the belief:
\begin{equation}
	w_{t}'^j = \frac{p(i=a^j)}{q(i=a^j)} = \frac{w_t^{a^j}}{\alpha w_t^{a^j} + (1-\alpha) (1/K)}
\end{equation}
Soft resampling provides non-zero gradients when $\alpha > 0$. We use $\alpha=0.5$ in our experiments.

\subsection{Model Learning}
\label{sec:model_learning}

One natural objective is to minimize the total prediction loss over all training sequences. For each training sequence, the prediction loss is 
\begin{equation}
	L_\mathrm{pred}(\theta) = \sum_{t\in \cal{O}} \ell(y_t,\hat{y}_t, \theta)
\end{equation}
where $\theta$ represents the predictor parameters,
$\cal{O}$ is the set of time indices with outputs,  and $\ell$ is a task-dependent loss function measuring the difference between the target output $y_t$ and the predicted output $\hat{y}_t$. 
Predictions in PF-RNN are made using the mean particle, $\hat{y}_t = f_\mathrm{out}(\bar{h}_t)$, where $f_\mathrm{out}$ is a learned function, and $\bar{h}_t = \sum_{i=1}^K w_t^i h_t^i$ is the weighted mean of the PF-RNN particles.
In our regression experiments, $\hat{y}_t$ is a real value and $\ell$ is the squared loss. In our classification experiments, $\hat{y}_t$ is a multinomial distribution over classes and $\ell$ is the cross-entropy loss.

The PF-RNN prediction  $\hat{y}_t$ is actually randomized, 
as its value depends on random variables used for stochastic memory updates and soft-resampling;
hence, a reasonable objective would be to minimize $\mathrm{E}[L_\mathrm{pred}(\theta)]$. In our algorithm, we optimize $L_\mathrm{pred}(\theta)$ instead of $\mathrm{E}[L_\mathrm{pred}(\theta)]$. We discuss the effect of this approximation in  Appendix. 

Another possible learning objective applies stronger model assumptions. To maximize the likelihood, we optimize a sampled version of an ELBO of  $p(y_t|x_{1:t},\theta)$, 
\begin{equation}
	L_{\mathrm{ELBO}}(\theta) = - \sum_{t\in \cal{O}}\log \frac{1}{K} \sum\limits_{i=1}^K p(y_t | \tau_{1:t}^i, x_{1:t}, \theta).
\end{equation}
where $\tau^i_{1:t}$ is a history chain for particle $i$, consisting of its ancestor indices during particle resampling and the random numbers used in stochastic memory updates\footnote{We have negated the ELBO to make it consistent with loss minimization.}.  
The derivation of the ELBO 
is provided in Appendix. 

The $p(y_t | \tau_{1:t}^i, x_{1:t}, \theta)$ terms in $L_{\mathrm{ELBO}}(\theta)$ are computed using appropriate probabilistic models. We apply $f_\mathrm{out}$ to each particle at each time step, generating outputs $\hat{y}_t^i = f_\mathrm{out}(h_t^i)$. Then, for classification $p(y_t|\tau_{1:t}^i, x_{1:t}, \theta)=\mathrm{CrossEntropy}(y_t, \hat{y}_t^i)$, which follows from the multinomial model; 
and for regression $p(y_t | \tau_{1:t}^i, x_{1:t}, \theta) = \mathrm{exp}(-||y_t - \hat{y}_t^i||)$, which follows from a unit variance Gaussian model.

Intuitively, $L_\mathrm{pred}$ encourages PF-RNN to make good predictions using the mean particle,
while $L_{\mathrm{ELBO}}$
encourages the model to learn a more meaningful belief distribution. They make different structure assumptions that result in different gradient flows. Empirically, we find that combining the two learning objectives works well: $L(\theta) = L_\mathrm{pred}(\theta) + \beta L_{\mathrm{ELBO}}(\theta)$, where $\beta$ is a weight parameter. We use $\beta=1.0$ in the experiments. 

\subsection{PF-LSTMs and PF-GRUs}

We now apply the PF-RNN to the two most popular RNN architectures, LSTM and GRU. 

For standard LSTM, the memory state $h_t$ consists of cell state $c_t$ and hidden state $\hslash_t$. The  memory   update is a  deterministic mapping controlled by   input gate~$i_t$,   forget gate~$f_t$, and output gate~$o_t$:
\begin{align}
	c_t &= f_t \circ c_{t-1} + i_t\circ \textrm{tanh}(\tilde{c}_t)\\
	\tilde{c}_t &= W_c[\hslash_{t-1}, x_t] + b_c\\
	\hslash_t &= o_t\circ\textrm{tanh}(c_t)
\end{align}
where $x_t$ is the current input and $\circ$ is the element-wise product.
For PF-LSTM, the memory state consists of a set of weighted particles $\{(\hslash_t^j, c_t^j, w_t^j)\}_{j=1}^K$.  To help PF-LSTM  track the latent particle belief effectively over a long history of data, we make two changes to the memory update equations. One is to add stochasticity: 
\begin{align}
	\tilde{c}_t^j &= W_c[\hslash_{t-1}^j, x_t] + b_c + \xi_t^j\\
	\xi_t^j &\sim \mathcal{N}(0, \Sigma_t^j)\\
	\Sigma_t^j &= W_\Sigma[\hslash_{t-1}^j, x_t] + b_\Sigma
\end{align}
for  $j=1,2,\ldots, K$. The motivation for stochastic memory updates is discussed in the previous section.
The other change, inspired by  LiGRU~\cite{ravanelli2018light}, is to replace the hyperbolic tangent activation of LSTM  by  ReLU activation and batch normalization~\cite{ioffe2015batch}: 
\begin{equation}
	c_t^j = f_t^j\circ c_{t-1}^j + i_t^j\circ \textrm{ReLU}(\textrm{BN}(\tilde{c}_t^j)),
\end{equation} 
for $j=1,2,\ldots, K$.  Recurrent networks are usually trained with truncated Back-Propagation-Through-Time (BPTT). %
Truncation affects the training of PF-RNNs, more than that of RNNs, as PF-RNNs maintain the latent belief  explicitly and may need a long sequence of inputs in order to approximate the belief well.
As shown in~\cite{ravanelli2018light}, 
ReLU activation, combined with batch normalization, has better numerical properties for backpropagation through many time steps, and thus it allows us to use longer sequence length with truncated BPTT.
The update equation for $\hslash_t^j$ remains the same as that for LSTM.
After updating the particles and their weights, we perform soft resampling to get a new set of particles. The PF-LSTM architecture is shown in Fig.~\ref{fig:nets}~(left side).

The PF-GRU model can be constructed similarly (right side of Fig.~\ref{fig:nets}). The details are available in Appendix. The same idea can be easily applied to other gated recurrent neural networks as well.

We implement PF-LSTM and PF-GRU in a parallel manner for all operations. In this case, all particles are updated in parallel benefiting from GPU acceleration.

\section{Experiments}

We evaluate PF-RNNs, specifically,  PF-LSTM and PF-GRU, on a synthetic 2D robot localization task and 10 sequence prediction datasets from various domains.
We compare PF-RNNs with the corresponding RNNs.
First, we use a fixed
latent state size for all models. The sizes are chosen so that the PF-RNN and the RNN have roughly the same number of trainable parameters. Next, we search over the latent state size for all models independently and compare the best results achieved.
We also perform an ablation study to understand the effect of individual components of PF-RNNs.
Implementation details are in Appendix. 

\subsection{Robot Localization}

\begin{figure}[!htb]
	\centering
	\includegraphics[width=\linewidth]{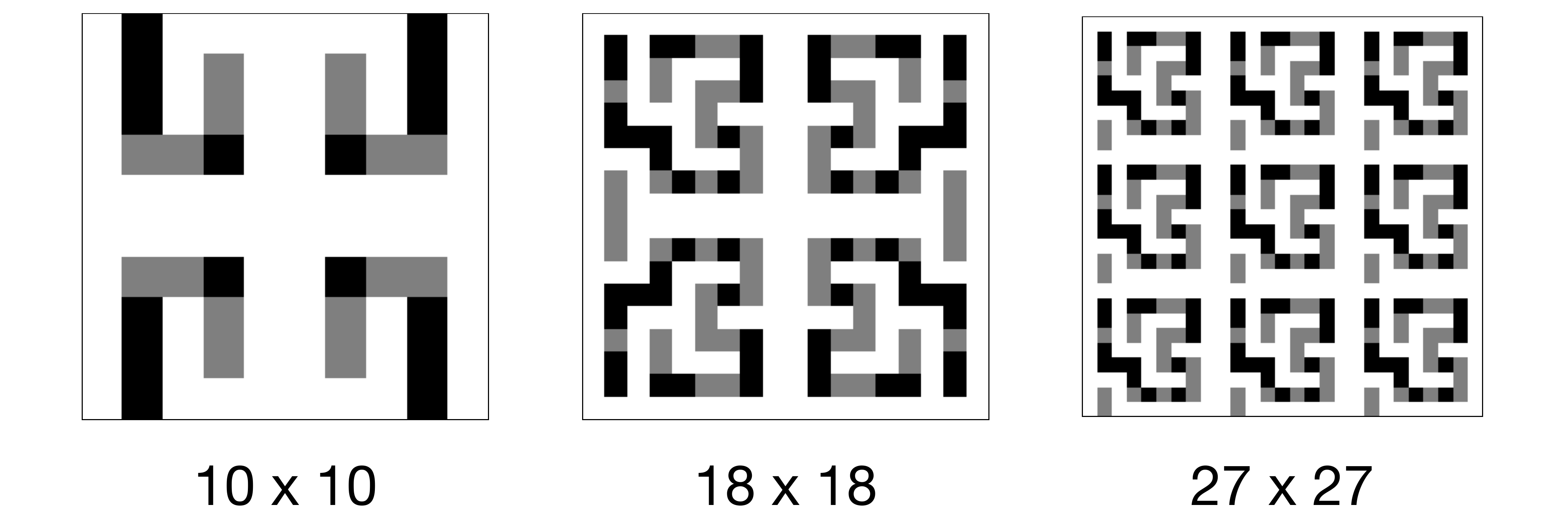}
	\centering
	\caption{\textbf{Localization in a symmetric maze}. Each maze is an $N\times N$ grid, with black and gray obstacles. Black obstacles serve also as landmarks.}
	\label{fig:maze}
\end{figure}

\begin{figure*}[t]
	\centering
	\begin{tabular}{c c c}
		\includegraphics[width=0.28\linewidth]{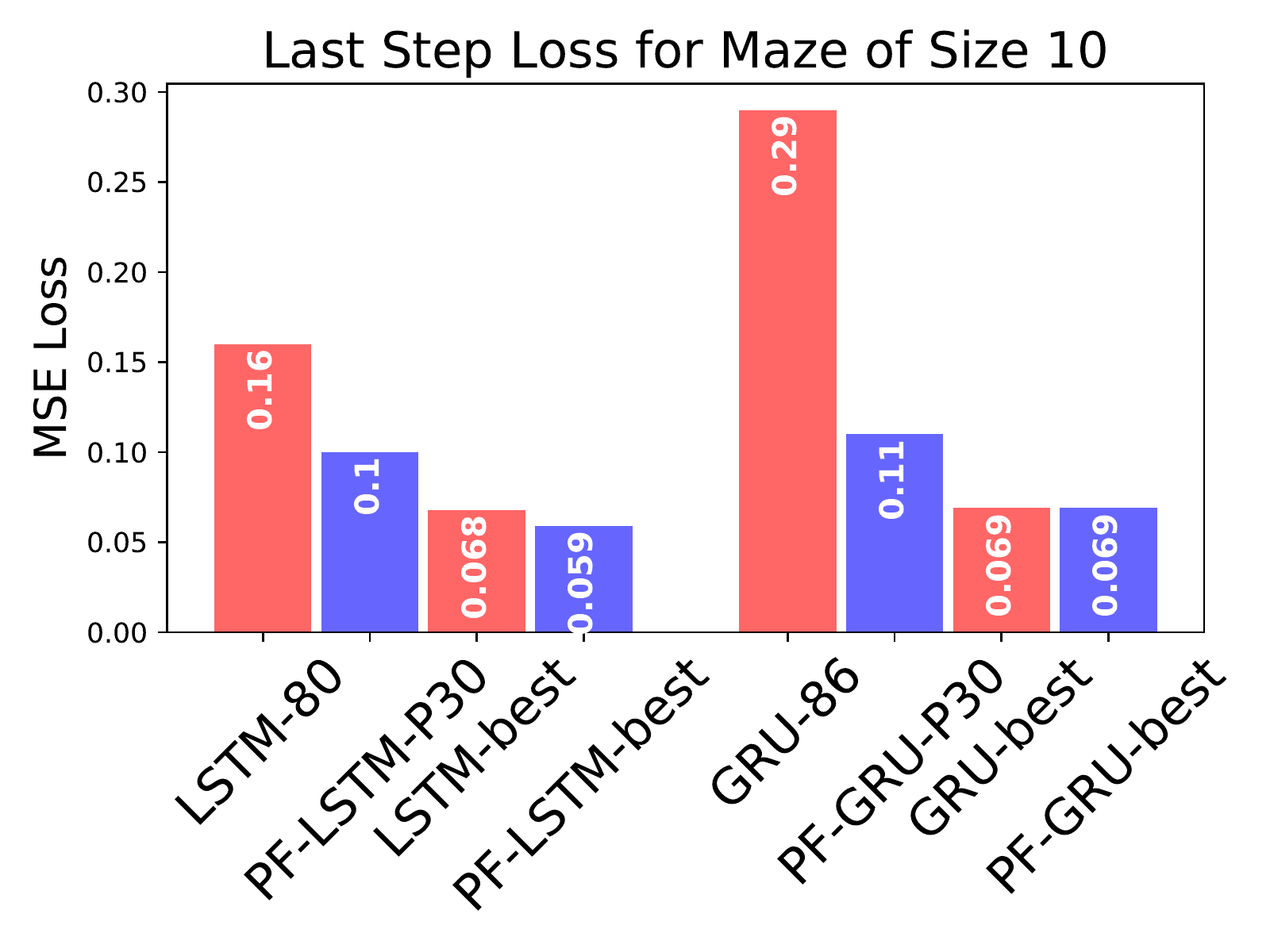} &
		\includegraphics[width=0.28\linewidth]{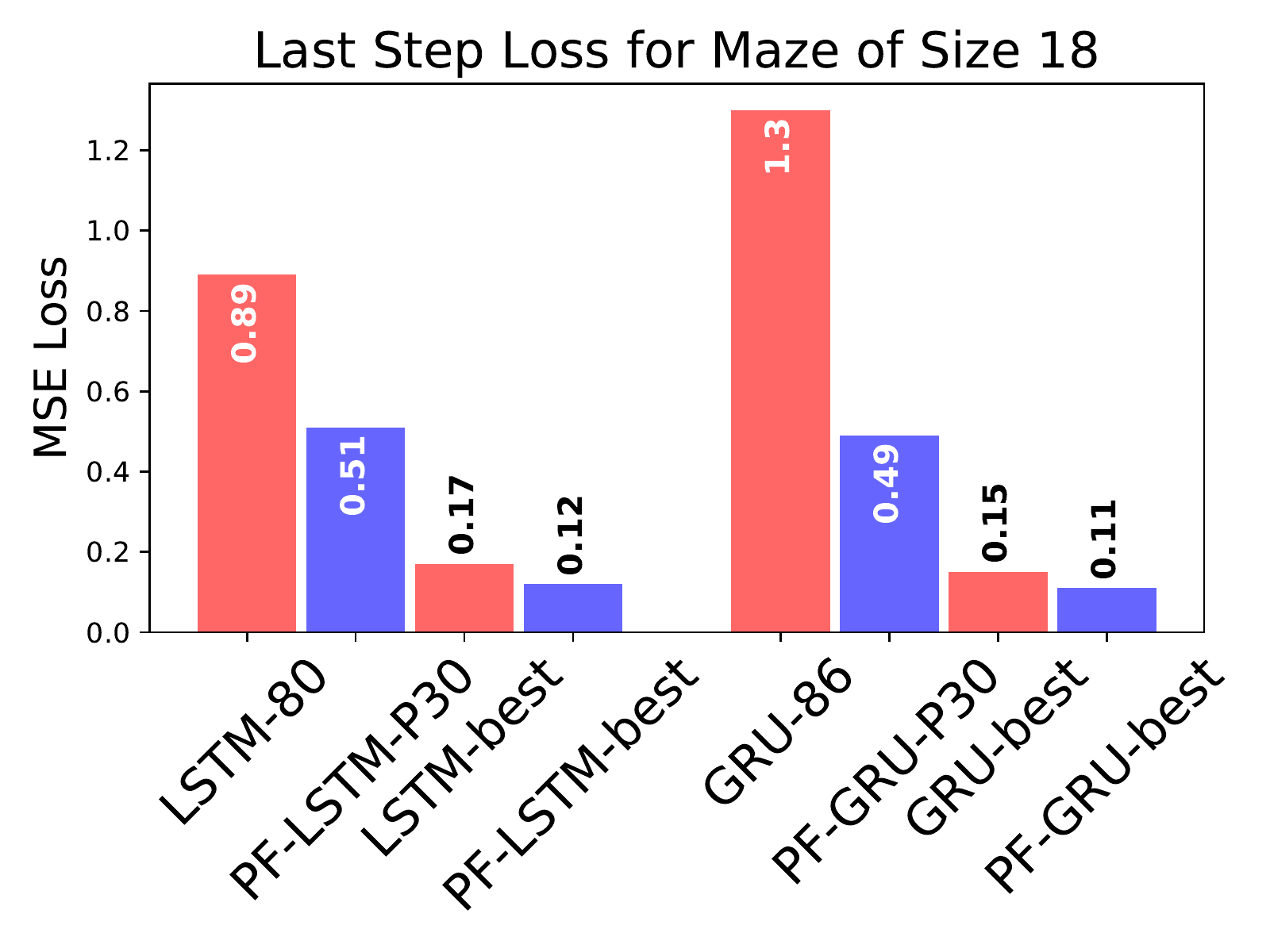} &
		\includegraphics[width=0.28\linewidth]{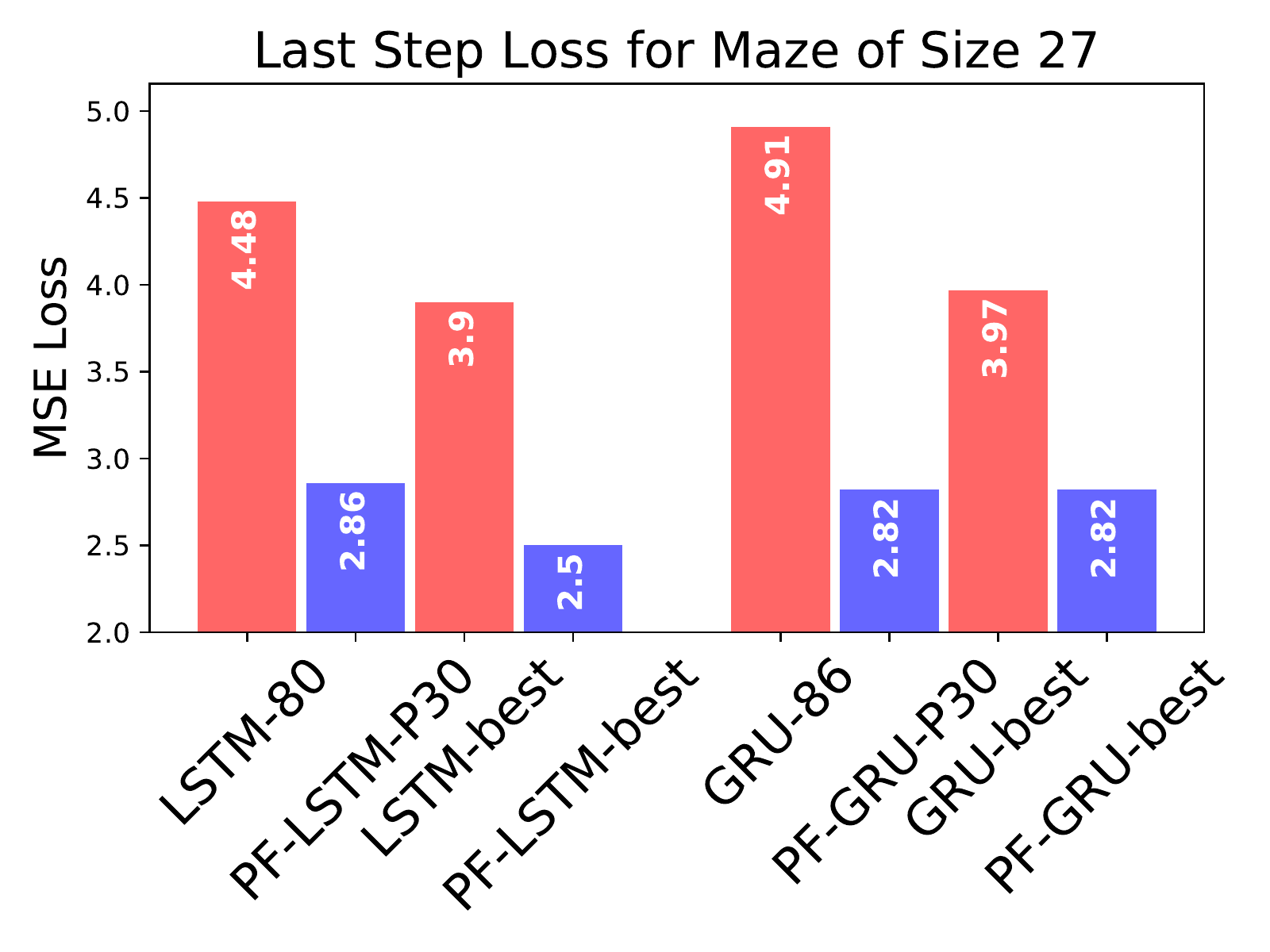}
	\end{tabular}
	\caption{\textbf{Performance comparison of the last-step prediction loss in robot localization}. 
		LSTM-80 and GRU-86 indicate LSTM and GRU  with latent state size of 80 and 86, respectively. 
		PF-LSTM-P30 and  PF-GRU-P30 indicate  PF-LSTM and PF-GRU with 30 particles and latent state size of 64. These parameters are chosen so that the RNN and the corresponding PF-RNN  have roughly the same number of trainable parameters.
		LSTM-best, GRU-best, PF-LSTM-best, and PF-GRU-best indicate the best-performing model after a search over the latent state size and the number of particles.
	}
	\label{fig:loc}
\end{figure*}

\begin{figure*}[t]
	\centering
	\begin{tabular}{ccccc}
		\includegraphics[width=0.18\linewidth]{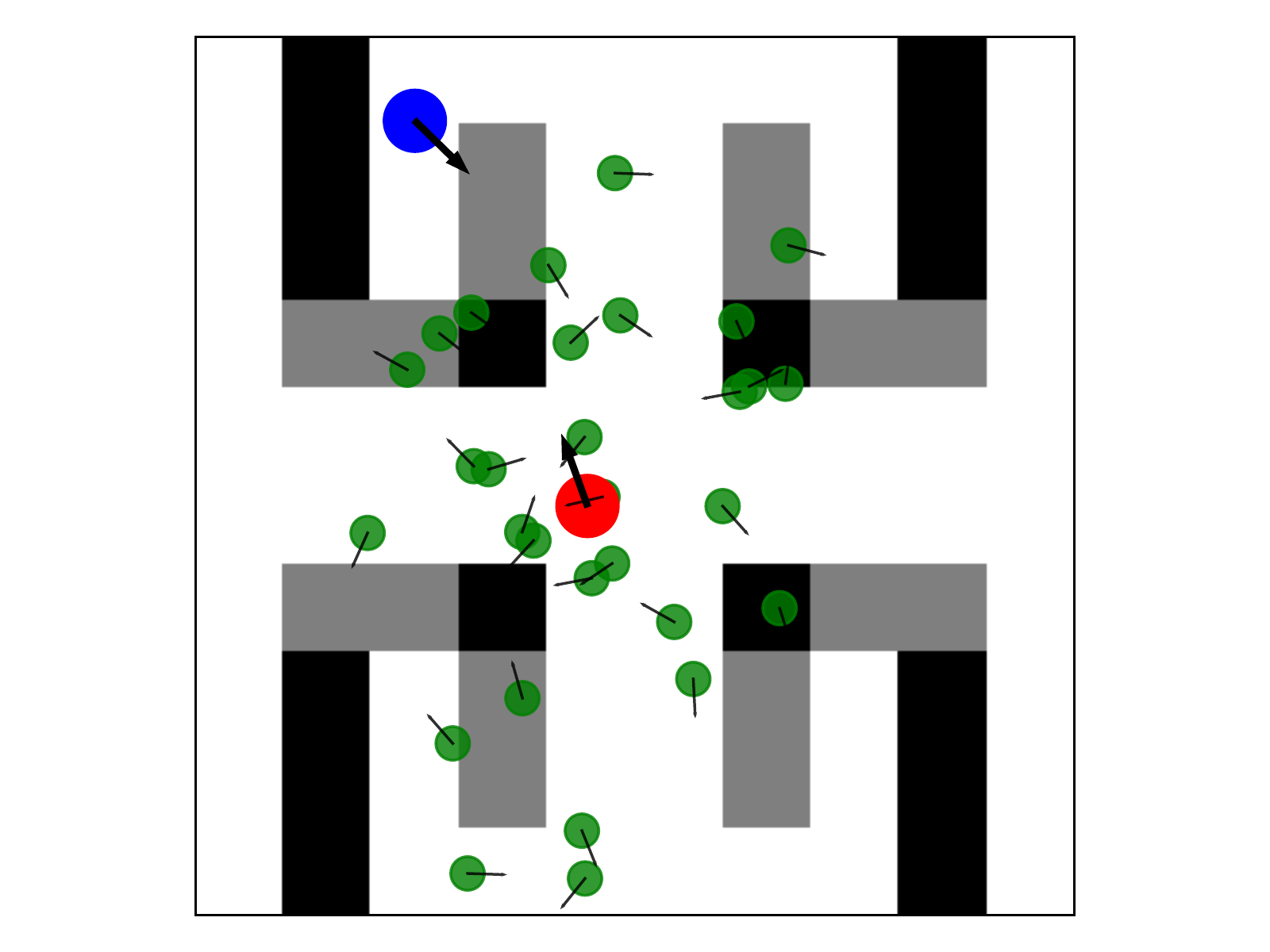} &
		\includegraphics[width=0.18\linewidth]{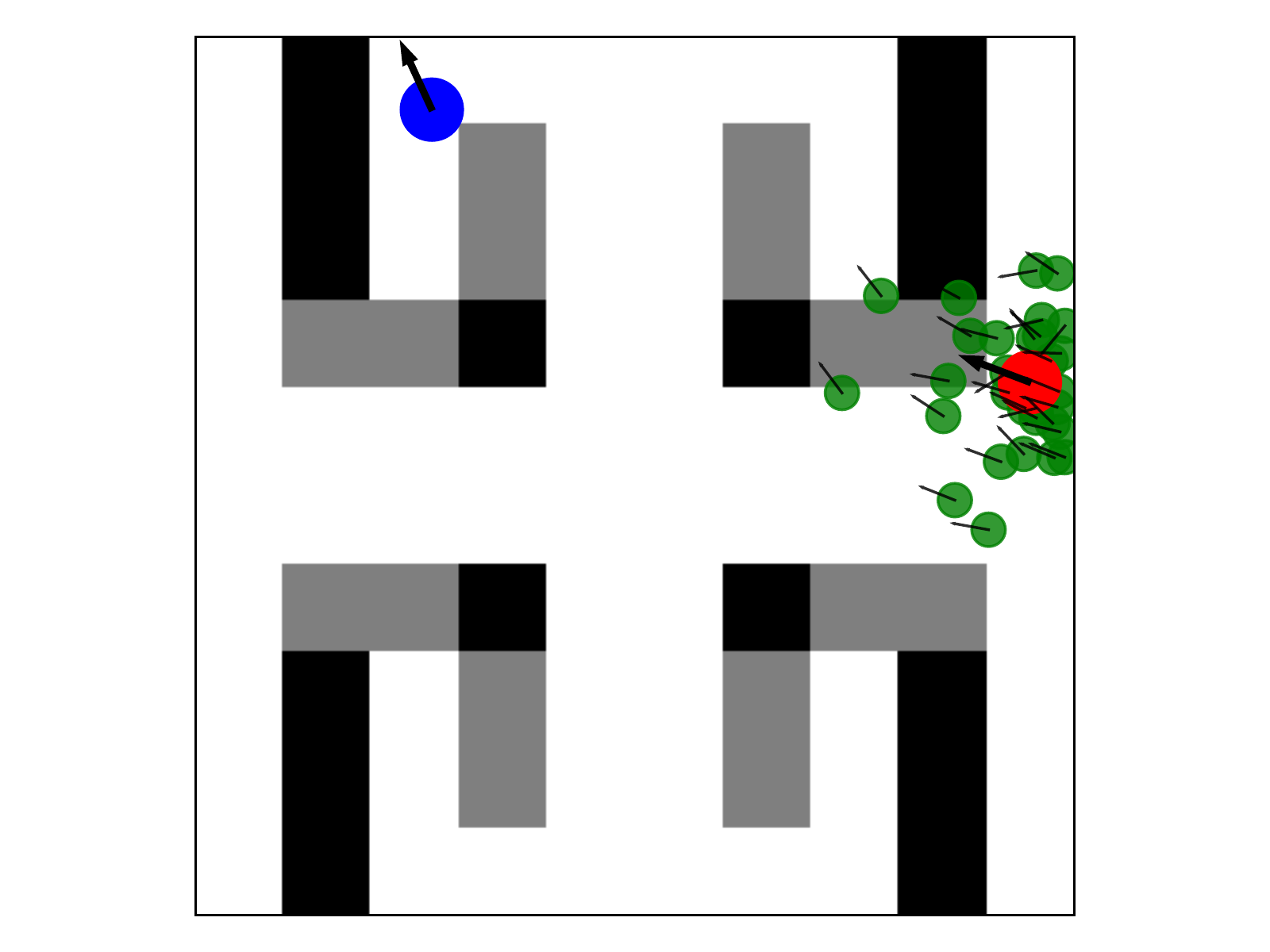} &
		\includegraphics[width=0.18\linewidth]{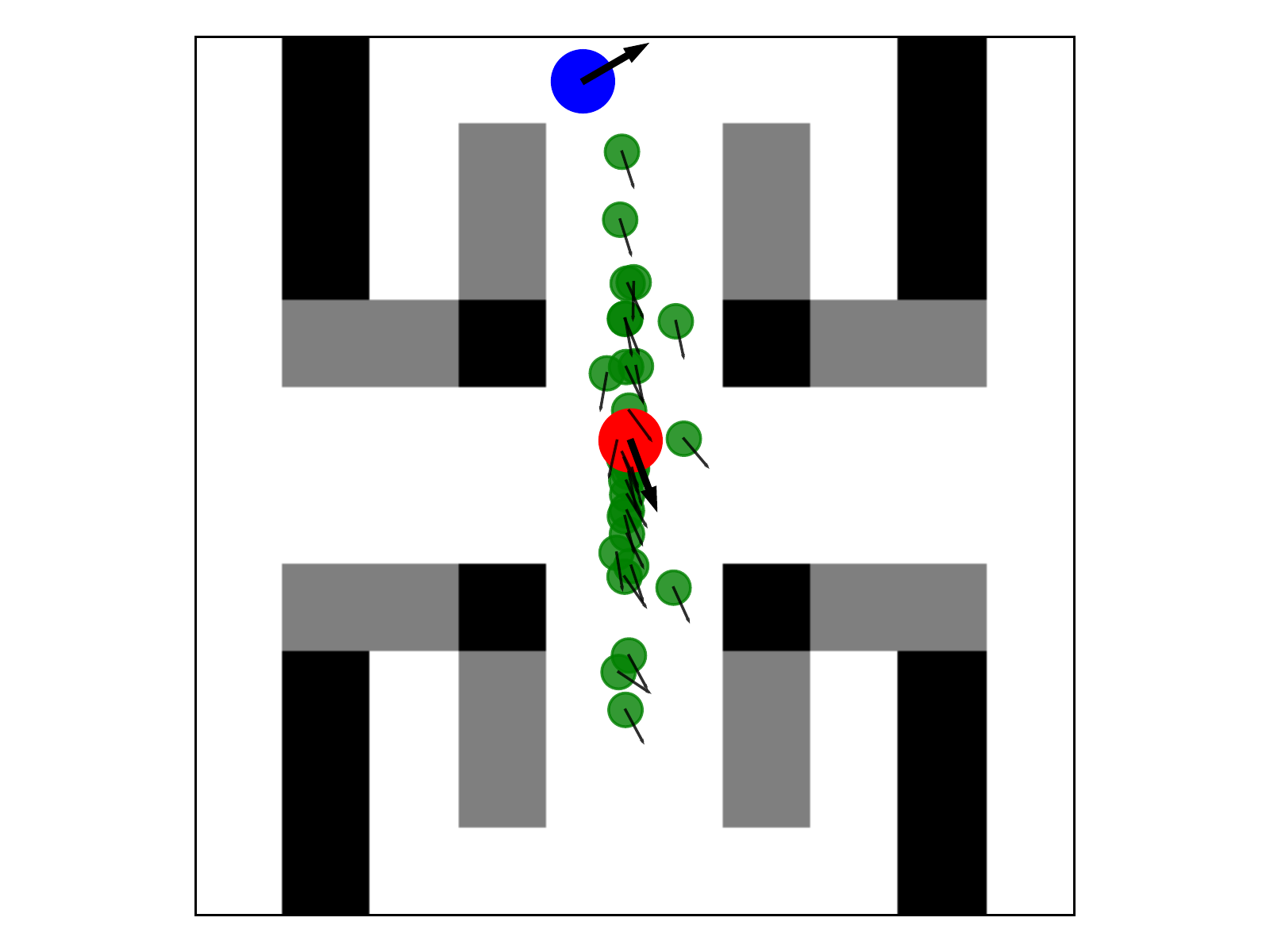} &
		\includegraphics[width=0.18\linewidth]{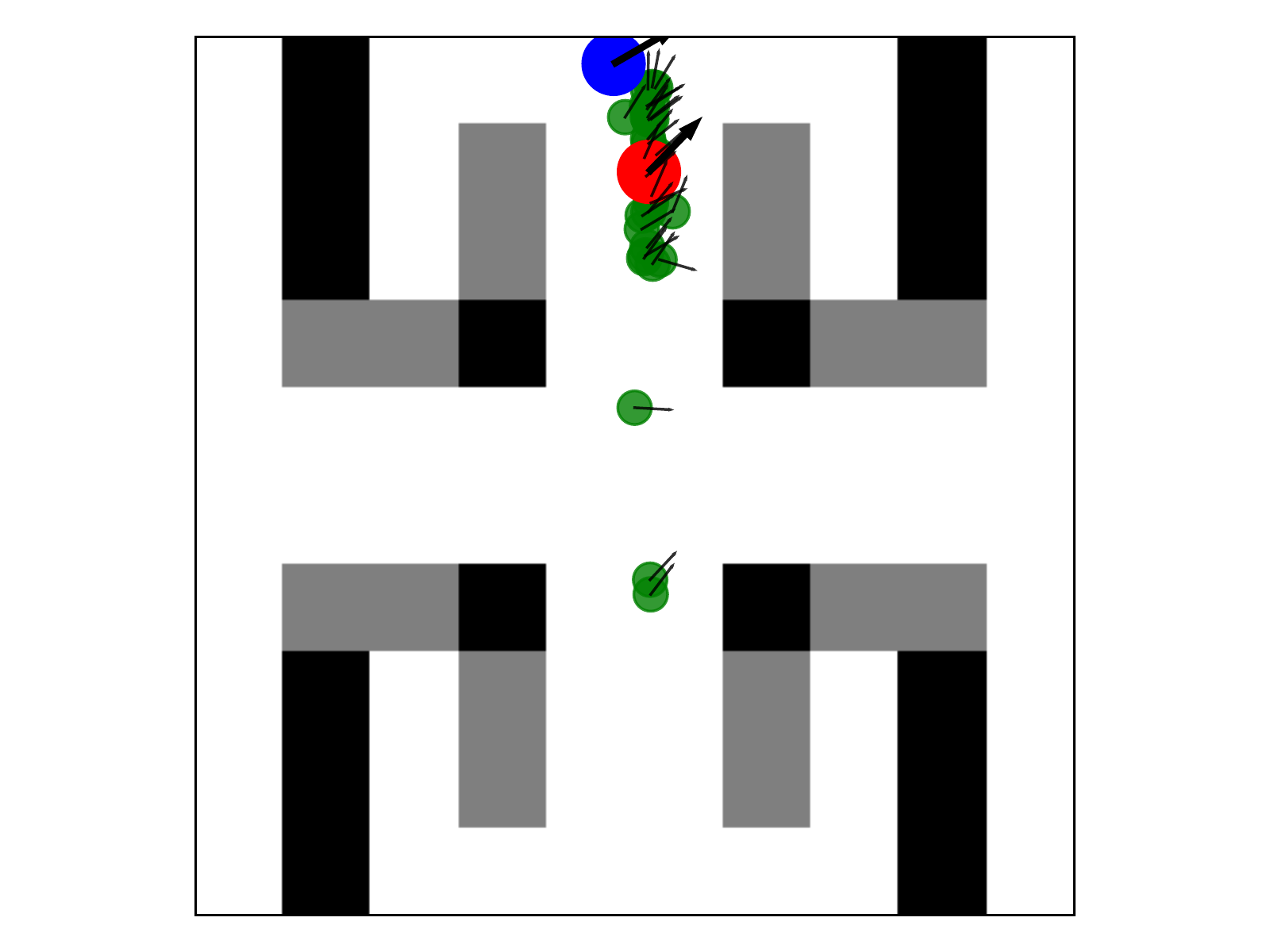} &
		\includegraphics[width=0.18\linewidth]{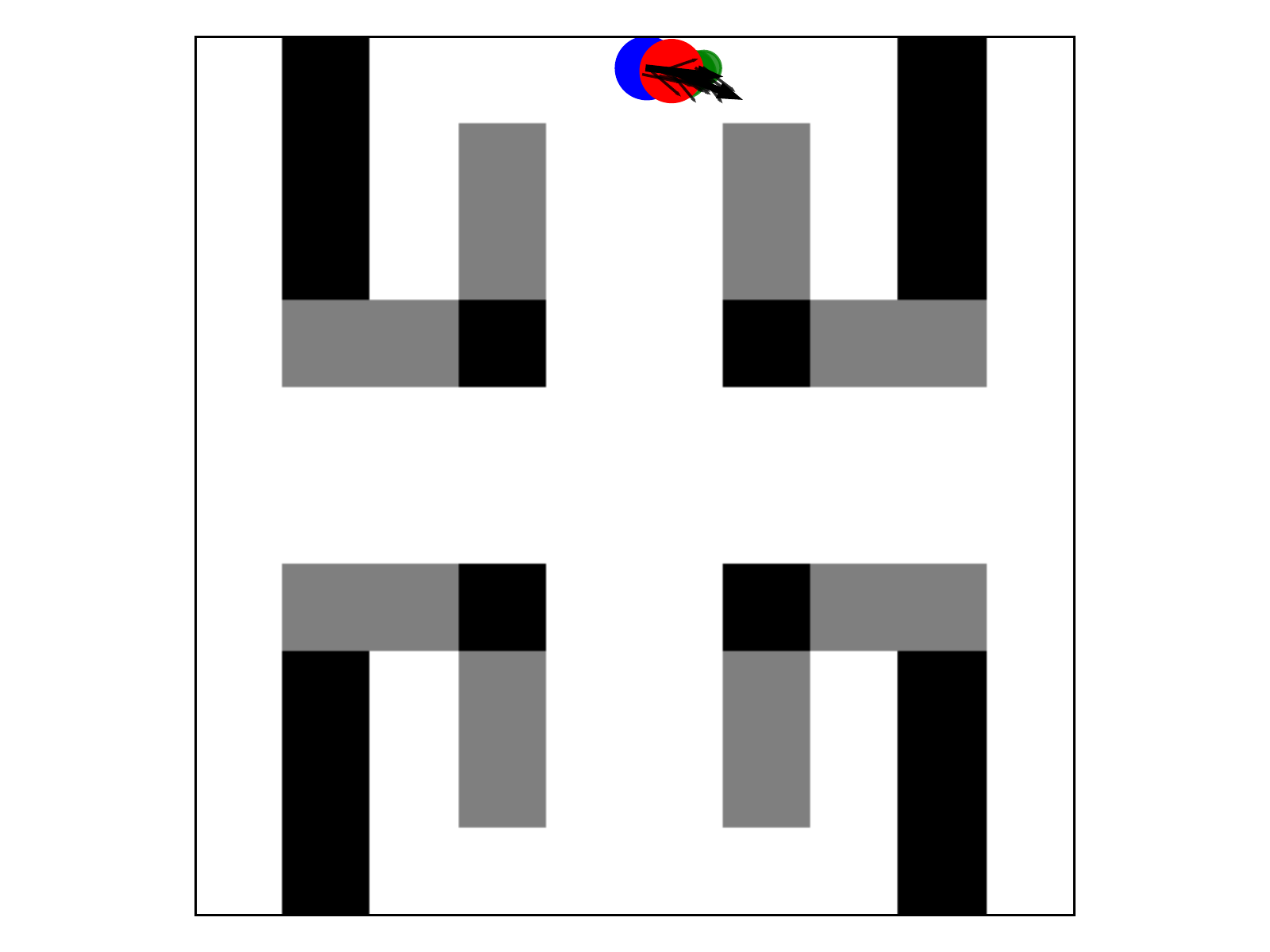} \\
		$t=0$ & $t=6$ & $t=22$ & $t=24$ & $t=26$
	\end{tabular}
	\caption{\textbf{Visualization of PF-LSTM latent particles}. Each figure shows the true robot pose (blue), predicted pose (red), and 30 predicted poses according to the latent particles (green).}
	\label{fig:vis}
\end{figure*}

We first evaluate PF-RNNs in a simple synthetic  domain 
designed to demonstrate the strengths of our approach.
The task is to localize a robot on a 2D  map (Fig.~\ref{fig:maze}),  given distance measurements to a set of landmarks.
In each environment, half of the obstructions have landmarks placed on their corners. The maps are designed to be highly symmetric, thus maintaining a belief is essential for localization.
We evaluate the models' sensitivity to the amount of uncertainty in the task by increasing the maze size.

The robot starts at a random location. In each step, it moves forward a distance $d = 0.2 + \epsilon_d, \epsilon_d\sim \mathcal{U}[-0.02, 0.02]$ where $\mathcal{U}$ is the uniform distribution, or chooses a new heading direction when about to hit a wall. The input $x_t$ consists of the last action $u_{t-1}$ and current observation $o_t$, which are noisy distance measurements to the 5 closest landmarks. The observation noise is $\mathcal{U}[-0.1, 0.1]$ for all mazes. The model takes $x_t$ as input and has no prior information on which components represent $u_t$ and $o_t$.  This information must be extracted from $x_t$ through learning. The task is to predict the robot pose at each time $t$, 2D coordinates and heading, given the history of inputs $\{x_i\}_{i=1}^t$.

We train models on a set of $10,000$ trajectories. We evaluate and test on another $1,000$ and $2,000$ trajectories, respectively.
The network architecture is the same for all models. Map features are extracted from the map using convolutional layers, two for the smaller maps and 5 for the largest map. Control and observation features are extracted from the inputs by two fully connected layers. The features are concatenated and input to the recurrent cell, PF-LSTM/PF-GRU or LSTM/GRU. The output of the recurrent cell is mapped to the pose prediction by a fully connected layer.
The training loss is the Mean Square Error~(MSE) between the predicted and ground truth pose, summed along the trajectories. The last step MSE is used as the evaluation metric. 

We make two sets of comparisons. First, we fix the latent state size and number of particles, such that PF-RNNs and the corresponding RNNs have a similar number of parameters. Specifically, we use a latent state size of 64 and 30 particles for PF-LSTM and PF-GRU, and a latent state size of 80 for LSTM and 86 for GRU. Second, we perform a search over these hyper-parameters independently for all models and datasets, including learning rate, dropout rate, batch size, and gradient clipping value, and report the best-achieved result. %

We compare PF-LSTM and PF-GRU with LSTM and GRU (Fig.~\ref{fig:loc}). 
Results show that PF-RNNs consistently outperform the corresponding RNNs. The performance benefits become more pronounced, as the size of the maze grows, resulting in increased difficulty of belief tracking.

Fig.~\ref{fig:vis}  visualizes the particle belief progression in a trained PF-LSTM. 
Additional examples are in Appendix.
PF-LSTM works similarly to a standard particle filtering algorithm and demonstrates a reasonable belief distribution. The particle predictions of the robot pose are initially spread out ($t=0$). As inputs accumulate, they begin to converge on some features, \textit{e.g.}, horizontal position ($t=22$); and eventually they converge to the true pose ($t=26$). The depicted example also demonstrates an interesting property of PF-LSTM. Initially, particle predictions converge towards a wrong, distant pose ($t=6$), because of the ambiguous observations in the symmetric environment. However, particle predictions spread out again ($t=22$), and converge to the true pose ($t=26$). 
This would be unusual for a standard particle filtering algorithm under the true robot dynamics: once particles converge to a wrong pose, it would be difficult to recover~\cite{thrun2001robust}. 
PF-LSTM succeeds here, because its latent representation and dynamics are learned from data.  Each particle in a PF-LSTM  independently aggregates information from the input history, and its dynamics is not constrained explicitly by the robot dynamics.
\begin{table*}[!htb]
\fontsize{8}{9}\selectfont
    \parbox{.5\linewidth}{
    \caption{Regression Loss}
    \label{tab:reg}
    \centering
    \begin{tabular}{lcccc}
    \toprule
             & NASDAQ        & AEP           & AIR            & PM             \\
             \midrule
LSTM         & 37.33         & 6.53          & 18.34          & 26.14          \\
PF-LSTM      & 4.65          & 4.57          & \textbf{13.12} & 21.23          \\
LSTM-best    & 2.53          & 6.53          & 17.69          & 26.14          \\
PF-LSTM-best & \textbf{1.82} & \textbf{3.72} & \textbf{13.12} & \textbf{19.04} \\
\midrule
GRU          & 4.93          & 6.57          & 17.7           & 23.59          \\
PF-GRU       & 1.33          & 5.33          & 19.32          & 20.77          \\
GRU-best     & 4.93          & 5.61          & \textbf{14.78} & 23.59          \\
PF-GRU-best  & \textbf{1.33} & \textbf{3.73} & 18.18          & \textbf{20.77} \\
SOTA         & \textbf{0.33} & -             & -              & -  \\
\bottomrule
\end{tabular}
    }
    \parbox{.5\linewidth}{
    \caption{Prediction Accuracy (\%)}
    \label{tab:pred}
    \centering
    \begin{tabular}{lcccccc}
    \toprule
             & LPA           & AREM         & GAS           & MR            & R52           & UID           \\
             \midrule
LSTM         & 91.7          & 99.1         & 76.6          & 73.1          & 81.1          & 93.2          \\
PF-LSTM      & \textbf{100}  & 99.1         & 89.9          & 78.3          & 89.1          & 95.3          \\
LSTM-best    & 98.3          & 100          & 81.2          & 73.1          & 81.1          & 99.1          \\
PF-LSTM-best & \textbf{100}  & \textbf{100} & \textbf{94.1} & \textbf{82.2} & \textbf{91.3} & \textbf{99.6} \\
\midrule
GRU          & 97.8          & 98.4         & 76.8          & 75.1          & 84.2          & 96.1          \\
PF-GRU       & 98.1          & 99.1         & \textbf{83.3} & 76.2          & 87.2          & 94.1          \\
GRU-best     & 97.8          & \textbf{100} & 80            & 75.2          & 84.2          & 99            \\
PF-GRU-best  & \textbf{99.2} & \textbf{100} & \textbf{83.3} & \textbf{79.6} & \textbf{89.1} & \textbf{99.5} \\
SOTA         & 91.3          & 94.4         & 80.9          & \textbf{83.1} & \textbf{93.8} & -         \\
\bottomrule
\end{tabular}
    }
\end{table*}

\subsection{General Sequence Prediction Tasks}

\textbf{Comparison with RNNs.} We evaluate PF-RNNs on various real-world sequence prediction datasets across multiple domains for both regression and classification tasks. Regression tasks include stock index prediction (NASDAQ~\cite{qin2017dual}), appliances energy prediction (AEP~\cite{candanedo2017data}), air quality prediction (AIR~\cite{de2008field} and PM~\cite{liang2015assessing}). Classification tasks include activity recognition (UID~\cite{casale2012personalization}, LPA~\cite{kaluvza2010agent}, AREM~\cite{palumbo2016human} and GAS~\cite{huerta2016online}) and text classification (R52~\cite{2007:phd-Ana-Cardoso-Cachopo} and MR~\cite{maas-EtAl:2011:ACL-HLT2011}). For regression tasks, a prediction is made after each time step, while for classification tasks prediction is made at the end of each sequence using the belief at the last time step. We demonstrate PF-RNN is sample efficient on relatively small datasets in terms of the sample efficiency (e.g., 9358 instances for AIR dataset) but also scales to large datasets for the generalization to complex tasks (e.g., around 1 million instances for GAS dataset).

We compare PF-LSTM and PF-GRU with the corresponding LSTM and GRU models. As PF-RNNs possess a similar structure with gated RNNs, we can directly use them as drop-in replacements.
We use a similar network architecture for all models and datasets. We follow the same experiment setup with the localization experiment and use 20 particles for PF-LSTM and PF-GRU. The input is passed through two fully-connected layers and input to the recurrent cell. The output of the recurrent cell is processed by another two fully connected layers to predict an output value. 
We perform grid search over standard training parameters (learning rate, batch size, gradient clipping value, regularization weight) for each model and dataset.

Results are shown in Table~\ref{tab:reg} and Table~\ref{tab:pred}. We observe that PF-LSTM and PF-GRU are generally better than the LSTM and GRU, both when the number of parameters is comparable, as well as with the best hyper-parameters within the model class.
PF-RNNs outperform the corresponding RNNs on 90\% of the datasets across different domains, including classification and regression tasks.

\textbf{Comparison with the State-of-the-Art.} Achieving state-of-the-art (SOTA) performance is not the focus of this paper. Nevertheless, we include the SOTA results for reference, when available, in the last row of Table~\ref{tab:reg} and Table~\ref{tab:pred}. We use the same training, validation and test sets as the SOTA methods. 
\begin{figure*}[!htb]
	\centering
	\begin{tabular}{c c c c} 
		\includegraphics[width=0.2\linewidth]{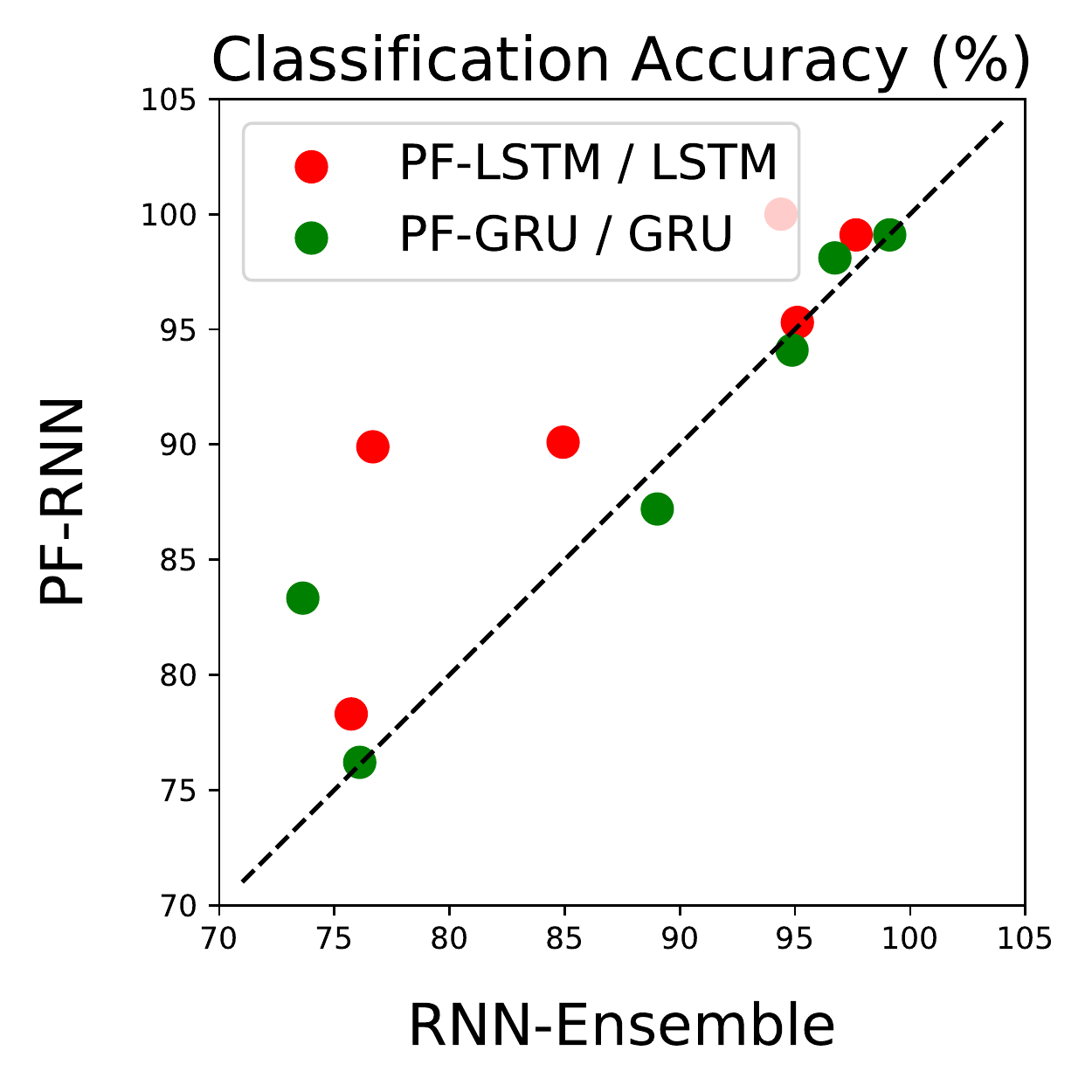} &
		\includegraphics[width=0.2\linewidth]{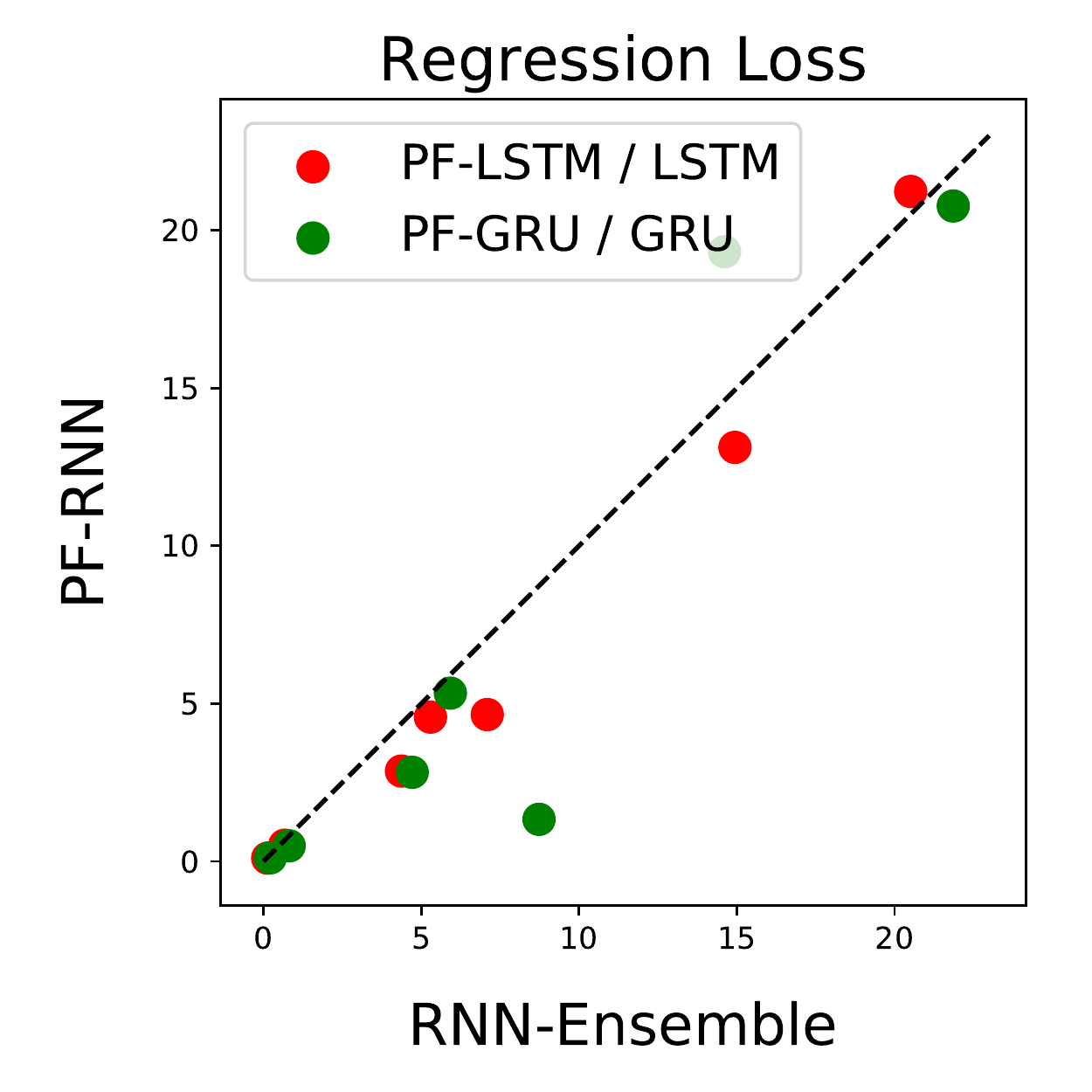}&
		\includegraphics[width=0.2\linewidth]{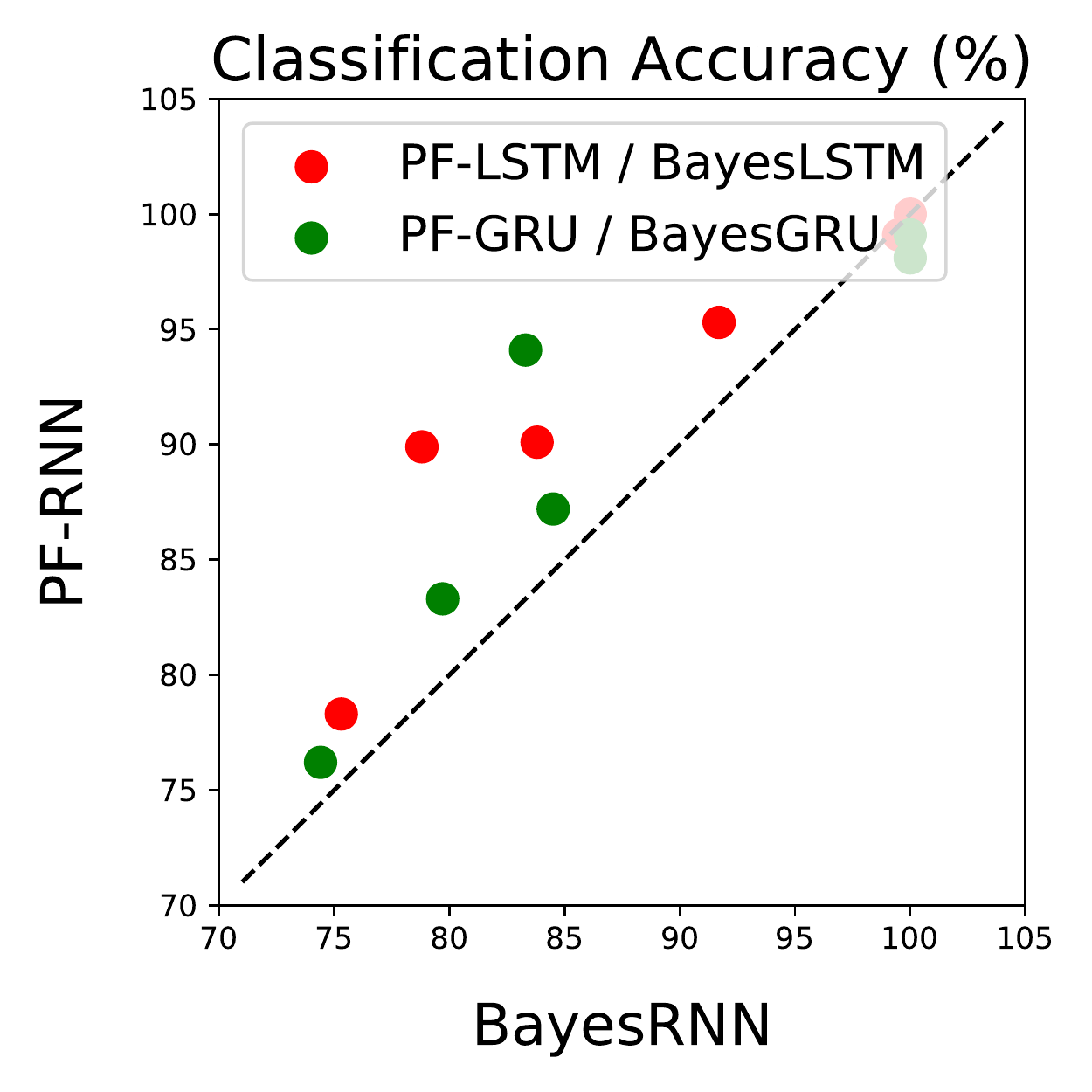} &
		\includegraphics[width=0.2\linewidth]{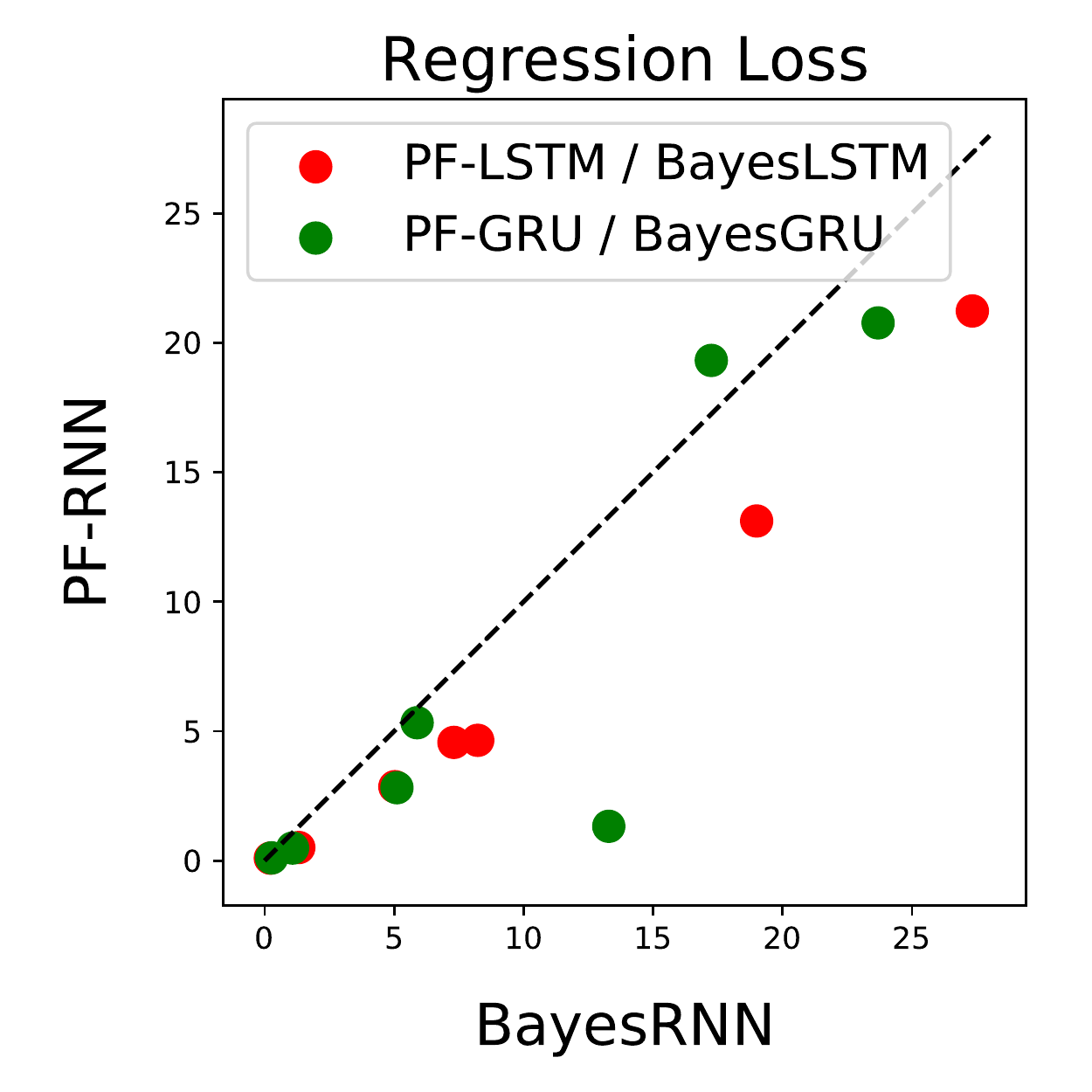}
	\end{tabular}
	\caption{\textbf{Comparison with ensembles and Bayesian RNNs}. Scatter plot of classification accuracy~(left) and regression loss~(right) over all datasets. The $x$-axis is the performance of an RNN ensemble or a Bayesian RNN, the $y$-axis is the performance of a PF-RNN. The dashed line corresponds to equal performance. Note that PF-RNN is better for points above the dashed line for classification, and for points below the dashed line for regression. Red points compare PF-LSTM with LSTM ensemble / Bayesian LSTM, blue points compare PF-GRU with GRU ensemble / Bayesian GRU. 
	}
	\label{fig:ensemble}
\end{figure*}
PF-RNNs, despite having only simple vanilla network components, perform better than the SOTA on LPA, AREM, and GAS. For NASDAQ, R52 and MR, SOTAs~\cite{qin2017dual,zhou2016text,2007:phd-Ana-Cardoso-Cachopo} use complex network structure designed specifically for the task, thus they work better than PF-RNNs. Future work may investigate PF-RNNs with larger, more complex network components, which would provide a more fair %
comparison with the SOTA.
\begin{table*}[!htb]
	\centering
	\caption{Ablation study.}
	\label{tab:abl}
	\fontsize{9}{9}\selectfont
	\begin{tabular}{lcccccccccc}
		\toprule
		\multirow{2}{*}{} & \multicolumn{9}{c}{PF-LSTM}                                                                & LSTM   \\
		\cmidrule(lr){2-10}\cmidrule(lr){11-11}
		& P1    & P5    & P10   & P20   & P30  & NoResample & NoBNReLU & NoELBO &ELBOonly & BNReLU \\
		\midrule
		Regression        & 0.81  & 0.66  & 0.59  & 0.55  & \textbf{0.51}  & 0.76  & 0.87         & 0.68     & 0.73   & 0.74   \\
		Classification    & 85.88 & 87.55 & 91.25 & 91.28 & \textbf{93.28} & 89.92 & 88.55        & 90.70    & 88.60  & 86.28 \\
		\bottomrule
	\end{tabular}
\end{table*}

\textbf{Comparison with Bayesian RNNs.}
As an alternative method for handling highly variable and noisy data, Bayesian RNNs~\cite{fortunato2017bayesian} maintain a distribution over learned parameters. We compare PF-RNNs with Bayesian RNNs on all the datasets. In particular, we use the same latent vector length with the LSTMs / GRUs for Bayesian RNNs following our setting in previous experiments. A summary of the results is given by Fig.~\ref{fig:ensemble} and detailed results are in the appendix. 
Bayesian RNNs can improve over RNNs, but PF-RNNs outperform Bayesian RNNs in 16 out of 20 cases, considering both PF-LSTM and PF-GRU.    

\textbf{Comparison with Ensembles.}
PF-RNN can be considered as a method for improving an RNN, by maintaining multiple copies and using particle filter updates.
Another common technique for improving the performance of a predictor using multiple copies is ensembling.
We compare the performance of PF-RNN with bagging \cite{breiman1996bagging}, an effective and commonly used ensembling method. Bagging trains each predictor in an ensemble by constructing a different training set of the same size through randomly sampling the original training set. We compare PF-RNNs using $K$ particles, with ensembles of $K$ RNNs. The results are summarized in Fig.~\ref{fig:ensemble}. Detailed results are in the appendix. 

Bagging reduces the variance of the predictors, but does not address belief tracking. 
Indeed, ensembles of RNNs generally improve over a single RNN, but the performance is substantially poorer than PF-RNNs for most datasets.

\subsection{Ablation Study}

We conduct a thorough ablation study on all datasets for PF-LSTMs to better understand their behavior. 
A summary of the results, averaged over all datasets, is shown in Table~\ref{tab:abl}. Detailed results are in the appendix. To compute a meaningful average for regression tasks, we need values with the same order of magnitude. We rescale all regression errors across the ablation methods into a range between 0 and 1.

We evaluate the influence of the following components: 1)~number of particles (P1, P5, P10, P20, P30); 
2)~soft-resampling (NoResample); 3)~replacing the hyperbolic tangent with ReLU activation and batch normalization (NoBNReLU); 4)~combining prediction loss and ELBO for training (NoELBO and ELBOonly). We conduct the same, independent hyper-parameter search as in previous experiments.

We observe that: 1) Using more particles makes a better prediction in general. The general performance increases from PF-LSTM P1 to P30. 
2) Soft-resampling improves the performance of PF-LSTM. 3) The combination of prediction loss and ELBO loss leads to better performance. 4) The ReLU activation with batch normalization helps to train PF-LSTM. Simply using ReLU with batch normalization does not make LSTM better than PF-LSTM.

\section{Conclusion}

PF-RNNs combine the strengths of RNNs and particle filtering by learning a latent particle belief representation and updating the belief with a particle filter. %
We apply the idea specifically to LSTM and GRU and construct PF-LSTM and PF-GRU, respectively. Our results show that PF-RNNs outperform the corresponding RNNs  on various sequence prediction tasks, including text classification, activity recognition, stock price prediction, robot localization, etc. %

Various aspects of the PF-RNN approach provide opportunities for further investigation. Currently, PF-RNNs make a prediction with the mean of the particle belief. An alternative is to  aggregate particles, e.g., by estimating higher-order moments, estimating
the entropy, attaching an additional RNN~\cite{igl2018deep}.

PF-RNNs can serve as drop-in replacements for  RNNs.
While this work focuses on sequence prediction and classification, the idea can be applied equally well to sequence generation and sequence-to-sequence prediction, with application to, e.g., image captioning and machine translation.

\section*{Acknowledgments}
This research is supported in part by the NUS AcRF Tier 1 grant R-252-000-A87-114 and the National Research Foundation Singapore under its AI Singapore Programme grant AISG-RP-2018-006.

\bibliographystyle{aaai}

\appendix

\section{Background}\label{sec:background}
\subsection{Bayes Filter}
In Bayes filters, we consider the problem of estimating the state $s_t$, given the history of controls $u_{1:t}$ and observations $o_{1:t}$. A Bayes filter aims to track the belief $b(s_t)$ of $s_t$, i.e., a posterior distribution $p(s_t\mid u_{1:t}, o_{1:t})$ with two steps: transition update with control $u_t$ and measurement update with observation $o_t$. 

\textbf{Transition update.} In transition update, we apply the motion model $p(s_t\mid s_{t-1}, u_t)$ to update $b(s_{t-1})$ into a predicted belief $\bar{b}(s_t)$. This predicts the state $s_t$ based only on the control $u_t$. 
\begin{equation}
    \bar{b}(s_t) = \int p(s_t\mid s_{t-1}, u_t)b(s_{t-1})d s_{t-1}\label{eqn:bayes_trans}
\end{equation}

\textbf{Measurement update.} In measurement update, we further use the observation model $p(o_t\mid s_t)$ to to update $\bar{b}(s_t)$ with the received observation $o_t$. 
\begin{equation}
    b(s_t) = \eta p(o_t\mid s_t)\bar{b}(s_t)\label{eqn:bayes_obs}
\end{equation}
where $\eta$ is the normalization factor. 

Bayes filter has many approximate implementations, e.g., Kalman filter with Gaussian posterior assumption and histogram filter with discrete approximation.

\subsection{Particle Filter}
Particle filter approximate the belief $b(s_t)$ with an empirical distribution, i.e., a set of $K$ particles $\{s_t^i\}_{i=1}^K$ with weights $\{w_t^i\}_{i=1}^K$. It approximates the Bayes filter algorithm by moving particles and updating their weights.

\textbf{Transition update.} Particle filter approximates function \ref{eqn:bayes_trans} by sampling $s_{t}^i$ from the motion model $p(s_t\mid s_{t-1}^i, u_t)$, for all particles $\{s_{t-1}^i\}_{i=1}^K$. 
\begin{equation}
    s_t^i\sim p(s_t\mid s_{t-1}^i, u_t)
\end{equation}

\textbf{Measurement update.} In particle filter, the measurement update is implemented by updating the weights of particles $\{w_{t-1}^i\}_{i=1}^K$ with observation model $p(o_t\mid s_t)$. 
\begin{equation}
    w_t^i = \eta p(o_t\mid s_t^i) w_{t-1}^i
\end{equation}
where $\eta$ is the normalization factor. Normally in particle filter, the observation model $p(o_t\mid s_t^i)$ is implemented as a generative observation model, e.g., a Gaussian. The observation likelihood is computed by evaluating $o_t$ in the parameterized distribution $p(o_t\mid s_t^i)$.

\textbf{Resampling.} One common issue of particle filter algorithm is particle degeneracy, i.e., most of the particles have near-zero weights. In this case, particle filter would fail to approximate the distribution because most of the particles are ineffective. One solution to this is to resample particles. After measurement update, we sample new particles from the belief $b(s_t)$ with probability proportionate to their weights $\{w_t^i\}_{i=1}^K$, i.e., we sample from a categorical distribution $p$ with probability for the $i$-th category $p(i)=w_t^i$. The particle weights are then updated to $1/K$. The new particle distribution approximates the same distribution, but with most of the particles distribute in the important region of the belief space.

\section{Derivation of Gradient Estimation}\label{sec:grad}

The PF-RNN predictor $\hat{y}_t$ is a randomized predictor, i.e.,
the value of $\hat{y}$ depends on random variables used for stochastic memory updates and soft-resampling. 
A reasonable objective function to minimize is $\mathrm{E}[L_\mathrm{pred}(\theta)] = \int_{h_p} p(h_p|\theta)\sum_{t\in \cal{O}}\ell(y_t,\hat{y}_t(h_p,\theta)),$ where $h_p$ denotes the set of random variables used in constructing the predictor. We have used the notation $\hat{y}_t(h_p,\theta)$ to make the dependence of $\hat{y}_t$ on $h_p$ and the predictor parameter $\theta$ explicit; however, for notational convenience, we have left out the dependence on input $x$. After fixing $h_p$, $\hat{y}_t$ becomes a fixed function of $\theta$ and the input $x$ which we use as our predictor; hence, we are optimizing the value of $\theta$ to generate good values of $h_p$ as well as a good prediction once $h_p$ has been generated.  

For simplicity of notation, we consider only one term in the summation. The gradient is:

\begin{align*}
\nabla_\theta \int_{h_p} p(h_p|\theta)& \ell(y_t,\hat{y}_t(h_p, \theta)) \\
&= \int_{h_p} \nabla_\theta (p(h_p|\theta) \ell(y_t,\hat{y}_t(h_p, \theta)))\\
&= \int_{h_p} (\nabla_\theta p(h_p|\theta)) \ell(y_t,\hat{y}_t(h_p, \theta)) \\
&\qquad+ \int_{h_p} p(h_p|\theta) \nabla_\theta \ell(y_t,\hat{y}_t(h_p, \theta)) \\
&= \int_{h_p} p(h_p|\theta) (\nabla_\theta \log p(h_p|\theta)) \ell(y_t,\hat{y}_t(h_p, \theta)) \\
&\qquad + \int_h p(h_p|\theta) \nabla_\theta \ell(y_t,\hat{y}_t(h_p, \theta))  
\end{align*}

The product rule is used in the second line of the derivation; and identity $\nabla_\theta \log p(h_p|\theta) = (\nabla_\theta \log p(h_p|\theta))/p(h_p|\theta) $ is used in the last line. Sampling $h_p$ and then computing $\nabla_\theta \log p(h_p|\theta) \ell(y_t,\hat{y}_t(h_p, \theta)) + \nabla_\theta \ell(y_t,\hat{y}_t(h_p, \theta))$ would give an unbiased estimator for the gradient of the objective function, as often required for stochastic gradient optimizers. However, the first term $\nabla_\theta \log p(h_p|\theta) \ell(y_t,\hat{y}_t(h_p, \theta))$ tends to have high variance~\cite{le2017auto}. As in other works, e.g. \cite{le2017auto}, we use only the second term $\nabla_\theta \ell(y_t,\hat{y}_t(h_p, \theta))$ to reduce the variance while introducing a small bias. 

We now consider the variables in $h_p$. In the PF-RNN we
obtain the value of the particle $h_t^i$ at time $t$ by sampling in the transition, conditioned on the parent $h_{t-1}^i$. 
We then do soft resampling which may replace the particle at position $i$ with another particle which has a different parent. 

For the transition, we use the \emph{reparameterization trick} \cite{kingma2013auto}, which reparameterizes the problem so that the subset of randomization variables $h_p$ that is involved in the transition becomes independent of $\theta$. In our problem, we would like to learn the covariance of $\xi_t^i$. It has a zero mean Gaussian distribution with diagonal covariance, where the diagonal covariance is a learned function of $h_{t-1}^i$ and $x_t$. As the covariance matrix is diagonal, we can consider each component separately. In the reparameterization trick, to draw a value $v$ from a Normal distribution $\mathcal{N}(0,\sigma(h_{t-1}^i,x_t))$ with mean 0 and standard deviation $\sigma(h_{t-1}^i,x_t)$, we instead draw $\varepsilon$ from $\mathcal{N}(0,1)$ and output $v=\varepsilon\sigma(h_{t-1}^i,x_t)$. The value $\varepsilon$, which forms part of $h_p$ can be treated as part of the input to the RNN, making it independent of $\theta$.

For soft resampling, the position of the replacement particle (and correspondingly its parent in the chain of ancestors) forms part of $h_p$. This depends on the parameter $\theta$ in $p(h_p|\theta)$ hence is affected by our approximation in the gradient estimate. We have not observed any problem attributable to the approximation in our experiments.

Finally, our predictor has the form $\hat{y}_t = f_\mathrm{out}(\bar{h}_t)$ where $\bar{h}_t = \sum_{i=1}^K w_t^i h_t^i$ and $f_\mathrm{out}$ if a learned function. The particles at time $t$ are averaged, each particle at each time step (except the first step) has a parent that is constructed by the soft resampling operations, and the transitions have sampled inputs from the reparameterization operation. Our approximate gradient is computed by running the particle filter, forming the predictor, then computing the gradient of the loss of the predictor formed from the filter.  

\section{Derivation of ELBO}\label{sec:elbo}

As discussed in the Model Learning section, we jointly optimize the prediction loss and an ELBO of $p(y_t|x_{1:t},\theta)$. In standard maximum likelihood estimation, we aim to find the parameter $\theta$ that maximizes the log likelihood $\log p(y_t|x_{1:t},\theta)$. This is often intractable. Instead, the evidence lower bound (ELBO) is often used, which is a variational lower bound of the log likelihood. We derive an ELBO using a similar technique as the one used in \cite{burda2015importance}, but heavily modified for the sequence prediction task.

We give a formal definition to the particle chain $\tau_t^i$. We define $\tau_t^i$ in a recursive manner: $\tau_t^i = \tau_{t-1}^i \bigcup \{\epsilon_t^i, a_t^i\}$, where $\epsilon_t^i\sim \mathcal{N}(0, I)$ is the reparameterized random number used in the stochastic transition and $a_t^i$ is the parent index chosen during soft-resampling. Given the input $x_{1:t}$, the particle chain determinizes the stochastic processes in PF-RNNs and produces a fixed particle $h_t^i$. In this case, sampling the particle chains $\tau_{1:t}^{1:K}$ conceptually gives $K$ RNNs.

We first consider the simpler case of the particle chains being drawn independently. Assume that $p(y_t|x_{1:t},\theta)=\int_{\tau_{1:t}}p(y_t,\tau_{1:t}|x_{1:t},\theta)$. Then

\begin{align*}
	\log p(y_t|x_{1:t},\theta) &= \log \int_{\tau_{1:t}} p(y_t,\tau_{1:t}|x_{1:t},\theta)\\
&= \log \int_{\tau_{1:t}^1,\ldots,\tau_{1:t}^K}q(\tau_{1:t}^1|x_{1:t})\ldots q(\tau_{1:t}^K|x_{1:t})\\
&\qquad\frac{1}{K}\sum_{i=1}^K  \frac{p(y_t,\tau_{1:t}^i|x_{1:t})}{q(\tau_{1:t}^i|x_{1:t})}\\
&\geq \int_{\tau_{1:t}^i,\ldots,\tau_{1:t}^K}q(\tau_{1:t}^1|x_{1:t})\ldots q(\tau_{1:t}^K|x_{1:t})\\
&\qquad\log \frac{1}{K}\sum_{i=1}^K  \frac{p(y_t,\tau_{1:t}^i|x_{1:t},\theta)}{q(\tau_{1:t}^i|x_{1:t})}
\end{align*}

where we have used Jensen's inequality and assumed a variational distribution $q(\tau_{1:t}^1|x_{1:t})\ldots q(\tau_{1:t}^K|x_{1:t})$ where $\tau_{1:t}^i$ are independently drawn from the same distribution.

We would like to use a particle filter to sample the particle chains for approximating the ELBO. The operations of our particle filter are as follows: at step $t$, we sample the transitions to generate new particles, we do resampling after the weights are updated by the observations, and we predict the target $y_t$. Due to the resampling operations, the particle chains in the particle filter are not independent. However, the particles generated at time $t$ by the resampling operation are conditionally independent given the history of what has been generated before then; it is determined just by the ancestor indices $a_t^i$ which are independently sampled given the particle chain distribution at that time by the resampling operation. Let $\Upsilon_{t}$ denote all the history before soft-resampling.
Assuming $p(y_t|x_{1:t},\theta)=\int_{\Upsilon_{t},a_t}p(y_t,\Upsilon_{t},a_t|x_{1:t},\theta)$, $a_t\sim p$, where $p$ is a multinomial distribution with $p(i) = w_t^i$,

\begin{align*}
\log \int_{\Upsilon_{t},a_t}& p(y_t,a_t,\Upsilon_{t}|x_{1:t},\theta)\\
&= \log \int_{\Upsilon_{t},a_t^1,\ldots,a_t^K}q(\Upsilon_{t}|x_{1:t},\theta)\\
&\qquad q(a_t^1|\Upsilon_{t},x_{1:t},\theta)\ldots q(a_t^K|\Upsilon_{t},x_{1:t},\theta)\cdot\\
&\qquad\frac{1}{K}\sum_{i=1}^K  \frac{p(y_t,a_t^i,\Upsilon_{t}|x_{1:t},\theta)}{q(\Upsilon_{t}|x_{1:t},\theta)q(a_t^i|\Upsilon_{t},x_{1:t},\theta)}\\
&\geq \int_{\Upsilon_{t},a_t^1,\ldots,a_t^K}q(\Upsilon_{t}|x_{1:t},\theta)\\
&\qquad q(a_t^1|\Upsilon_{t},x_{1:t},\theta)\ldots q(a_t^K|\Upsilon_{t},x_{1:t},\theta)\cdot\\ 
&\qquad \log \frac{1}{K}\sum_{i=1}^K  \frac{p(y_t,a_t^i,\Upsilon_{t}|x_{1:t},\theta)}{q(\Upsilon_{t}|x_{1:t},\theta)q(a_t^i|\Upsilon_{t},x_{1:t},\theta)}
\end{align*}

We now use the variational distributions $q(a_t^i|\Upsilon_{t}, x_{1:t},\theta)=p(a_t^i|\Upsilon_{t},x_{1:t},\theta)$ and $q(\Upsilon_{t}|x_{1:t},\theta) = p(\Upsilon_{t}|x_{1:t},\theta)$. This allows the simplification $\frac{p(y_t,a_t^i,\Upsilon_{t}|x_{1:t},\theta)}{p(a_t^i|\Upsilon_{t},x_{1:t},\theta)p(\Upsilon_{t}|x_{1:t},\theta)}=p(y_t|a_t^i,\Upsilon_{t},x_{1:t},\theta)$. We opt for simplicity in the ELBO; the use of other variational distributions may allow the weights $w_t^i$ to be optimized, as well as to incorporate information about $y_t$ in the variational distribution, but at the expense of optimizing a more complex objective. We rely on $L_\mathrm{pred}$ to optimize $w_t^i$, which is fully task-oriented. 

After negation (to be consistent with minimizing loss), we get our objective function

\begin{align*}
-\int_{\Upsilon_{t},a_t^i,\ldots,a_t^K}&p(\Upsilon_{t}|x_{1:t},\theta)p(a_t^1|\Upsilon_{t},x_{1:t},\theta)\ldots p(a_t^K|\Upsilon_{t},x_{1:t},\theta)\\
&\log \frac{1}{K}\sum_{i=1}^K  p(y_t|a_t^i,\Upsilon_{t},x_{1:t},\theta)
\end{align*}

We then sample $\Upsilon_{t},a_t^1,\ldots,a_t^K$ from $p(\Upsilon_{t}|x_{1:t},\theta)p(a_t^1|x_{1:t},\theta)\ldots p(a_t^K|x_{1:t},\theta)$ to get $\log \frac{1}{K}\sum_{i=1}^K  p(y_t|a_t^i,\Upsilon_{t},x_{1:t},\theta)$ and differentiate to get an estimator for the gradient.

Computing the derivative:

\begin{align*}
	&\nabla_{\theta}\log \frac{1}{K}\sum_{i=1}^K  p(y_t|a_t^i,\Upsilon_{t},x_{1:t},\theta)\\
	&=\frac{1}{\frac{1}{K}\sum_{i=1}^K  p(y_t|a_t^i,\Upsilon_{t},x_{1:t},\theta)}\nabla_{\theta}(\frac{1}{K}\sum_{i=1}^K  p(y_t|a_t^i,\Upsilon_{t},x_{1:t},\theta))\\
&=\sum_{i=1}^K\frac{p(y_t|a_t^i,\Upsilon_{t},x_{1:t},\theta)}{\sum_{i=1}^K  p(y_t|a_t^i,\Upsilon_{t},x_{1:t},\theta)}\nabla_{\theta}\log p(y_t|a_t^i,\Upsilon_{t},x_{1:t},\theta).
\end{align*}

The parameter $\theta$ also affects the sampling distribution $p(\Upsilon_{t}|x_{1:t},\theta)$ of the ELBO. To get an unbiased estimate of the gradient, we will need to compute 
\begin{align*}
(\nabla_\theta &p(\Upsilon_{t}|x_{1:t},\theta)p(a_t^1|x_{1:t},\theta)\ldots p(a_t^K|x_{1:t},\theta)) \\
&\log \frac{1}{K}\sum_{i=1}^K  p(y_t|a_t^i,\Upsilon_{t}x_{1:t},\theta) \\
&+ \nabla_{\theta}\log \frac{1}{K}\sum_{i=1}^K  p(y_t|a_t^i,\Upsilon_{t},x_{1:t},\theta)
\end{align*}
after sampling $\Upsilon_{t},a_t^1,\ldots,a_t^K$. As in the previous section, we only compute $\nabla_{\theta}\log \frac{1}{K}\sum_{i=1}^K  p(y_t|a_t^i,\Upsilon_{t},x_{1:t},\theta)$, obtaining a lower variance but biased estimator of the gradient.

Finally we have $p(y_t|a_t^i,\Upsilon_{t},x_{1:t},\theta)=p(y_t|\tau_{1:t}^i,x_{1:t},\theta)$ according to our definition, since we are able to do the prediction with only the knowledge of the single RNN. As mentioned in the Model Learning section, we compute $p(y_t|\tau_{1:t}^i,x_{1:t},\theta)$ by applying $f_\mathrm{out}$ on each particle $h_t^i$, assuming cross-entropy for classification tasks and unit variance Gaussian model for regression, gradients are carried through the chain of particle states through time by BPTT.

\section{PF-GRU Network Architecture}\label{sec:pfgru}
The standard GRU maintains the latent state $\hslash_t$, and updates it with two gates: reset gate $r_t$ and update gate $z_t$. The memory is updated according to the following equations:
\begin{align*}
\hslash_t &= (1 - z_t)\circ \mathrm{tanh}(n_t) + z_t\circ \hslash_{t-1}\\
n_t &= W_n[r_t\circ \hslash_{t-1}, x_t] + b_n
\end{align*}
where $W_n$ and $b_n$ are the corresponding weights and biases.

The state of the PF-GRU is a set of weighted particles $\{\hslash_t^i, w_t^i\}_{i=1}^K$. Similarly to PF-LSTM, we perform a stochastic cell update, where the update to the cell, $n_t^i$, is sampled from a Gaussian distribution:
\begin{align*}
n_t^i &= W_n[r_t^i\circ \hslash_{t-1}^i, x_t] + b_n + \epsilon_t^i\\
\epsilon_t^i&\sim \mathcal{N}(0, \Sigma_t^i)\\
\Sigma_t^i &= W_\Sigma[\hslash_{t-1}^i, x_t] + b_\Sigma
\end{align*}

Besides, to allow training with longer truncated BPTT, similarly to PF-LSTM, the hyperbolic tangent function is replaced by ReLU with batch normalization:
\begin{equation}
\hslash_t^i = (1-z_t^i)\circ \mathrm{ReLU}(\mathrm{BN}(n_t^i)) + z_t\circ \hslash_{t-1}^i
\end{equation}
\begin{table*}[t]
	\centering
	\fontsize{8}{9}\selectfont
	\caption{Comparison with Bagging and Bayesian RNNs}
	\label{tab:ens_bayes}
	\begin{tabular}{cccccccccccccc}
		\toprule
		& \multicolumn{7}{c}{Regression Loss}                                                                             & \multicolumn{6}{c}{Prediction Accuracy (\%)}\\
		& maze10        & maze18        & maze27        & NASDAQ        & AEP           & AIR            & PM             & LPA          & AREM          & GAS           & MR            & R52            & UID            \\
						\cmidrule(lr){0-7}\cmidrule(lr){9-14}
		BayesLSTM & 0.22          & 1.32          & 5.03          & 8.23          & 7.31          & 19.00          & 27.33          & \textbf{100} & \textbf{99.5} & 78.8          & 75.3          & 83.8           & 91.7           \\
		LSTM-E    & 0.13          & 0.67          & 4.37          & 5.29          & 7.09          & 14.94          & \textbf{20.51} & 94.39        & 97.65         & 76.67         & 75.73         & 84.93          & 95.1           \\
		PF-LSTM   & \textbf{0.10} & \textbf{0.51} & \textbf{2.86} & \textbf{4.65} & \textbf{4.57} & \textbf{13.12} & 21.23          & \textbf{100} & 99.1          & \textbf{89.9} & \textbf{78.3} & \textbf{89.1}  & \textbf{95.3}  \\
		\midrule
		BayesGRU  & 0.27          & 1.08          & 5.11          & 13.29         & 5.89          & 17.25          & 23.69          & \textbf{100} & \textbf{100}  & 79.7          & 74.4          & 84.5           & 83.3           \\
		GRU-E     & 0.21          & 0.81          & 4.71          & 5.92          & 8.73          & \textbf{14.61} & 21.86          & 96.73        & 99.11         & 73.63         & \textbf{76.1} & \textbf{89.02} & \textbf{94.87} \\
		PF-GRU    & \textbf{0.11} & \textbf{0.49} & \textbf{2.82} & \textbf{1.33} & \textbf{5.33} & 19.32          & \textbf{20.77} & 98.1         & 99.1          & \textbf{83.3} & \textbf{76.2} & 87.2           & 94.1          \\
		\bottomrule
	\end{tabular}
\end{table*}
\begin{table*}[t]
		\centering
	\fontsize{8}{9}\selectfont
	\caption{Ablation Studies on PF-LSTM}
	\label{tab:ablation_details}
	\begin{tabular}{cccccccccccccc}
		\toprule
		& \multicolumn{7}{c}{Regression Loss}                                                                             & \multicolumn{6}{c}{Prediction Accuracy (\%)}\\
		& Maze 10        & Maze 18       & Maze 27        & AEP           & AIR            & PM             & NASDAQ        & UID           & R52           & LPA          & AREM          & GAS           & MR            \\
						\cmidrule(lr){0-7}\cmidrule(lr){9-14}
		PF-LSTM-P1         & 0.41           & 1.1           & 3.84           & 6.24          & 15.41          & 22.12          & 13.21         & 89.7          & 86.4          & 97.1         & 98.1          & 71.3          & 72.7          \\
		PF-LSTM-P5         & 0.17           & 0.65          & 3.81           & 5.73          & 13.89          & 21.98          & 10.97         & 94.9          & 88.1          & 96.9         & 98.4          & 71.8          & 75.2          \\
		PF-LSTM-P10        & 0.12           & 0.6           & 3.11           & 4.88          & 14.23          & 21.33          & 9.25          & 94.7          & 88.7          & 97.7         & 98.4          & 89.2          & 78.8          \\
		PF-LSTM-P20        & 0.12           & 0.55          & 3.03           & 4.57          & \textbf{13.12} & 21.23          & 4.64          & 95.3          & \textbf{89.1} & \textbf{100} & \textbf{99.1} & 89.9          & 78.3          \\
		PF-LSTM-P30        & \textbf{0.103} & \textbf{0.51} & \textbf{2.862} & \textbf{3.24} & 14.44          & \textbf{20.05} & \textbf{4.55} & \textbf{98.4} & 88.5          & \textbf{100} & 98.1          & \textbf{94.1} & \textbf{80.6} \\
		PF-LSTM-NoResample & 0.14           & 1.42          & 4.03           & 5.27          & 18.47          & 20.24          & 21.23         & 95            & 85.2          & 95.3         & 99            & 88.7          & 76.3          \\
		PF-LSTM-NoBNReLU   & 0.19           & 1.66          & 4.01           & 5.03          & 16.73          & 21.01          & 51.32         & 95.7          & 84.8          & 96.2         & 98.6          & 78.9          & 77.1          \\
		PF-LSTM-NoELBO     & 0.18           & 1.23          & 3.21           & 4.88          & 16.89          & 21.79          & 6.59          & 94.2          & 88.1          & 97.7         & 97.2          & 90.8          & 76.2          \\
		PF-LSTM-ELBOonly   & 0.34           & 1.12          & 3.01           & 4.94          & 17.15          & 21.11          & 9.87          & 91.9          & 87.8          & 97.3         & 97            & 81.4          & 76.2          \\
		LSTM-BNReLU        & 0.21           & 1.05          & 4.15           & 5.35          & 16.29          & 21.38          & 15.72         & 91.2          & 82.3          & 97.7         & 98.9          & 75.7          & 71.9         \\
		\bottomrule
	\end{tabular}
\end{table*}

\section{Experiment Details}\label{sec:setup}
\textbf{Experiment platform.} We implement all models in PyTorch and train using NVidia GTX1080Ti GPUs. \\
\textbf{Hyper-parameter search. } For all models, we perform a grid search over standard training parameters (learning rate, batch size, gradient clipping value, regularization weight). Specifically, we perform a standard grid search over the learning rate $\{1^{-4}, 3^{-4}, 5^{-4}\}$ with RMSProp optimizer, batch size $\{32, 64, 128\}$, gradient clipping value $\{3, 5\}$ and L2 regularization weight of $\{0.001, 0.0001\}$. When model hyper-parameter search is performed,  for PF-LSTM/PF-GRU it is a simple grid search over $\{64, 128, 256\}$ latent state sizes and $\{20, 30\}$ particles. For LSTM/GRU we do a more careful search, and increase the latent state size, $\{80/86, 64, 128, 256, 512, 1024\}$, until the performance stops improving.\\
\textbf{General sequence prediction tasks details.} For all the general sequence prediction tasks (except for text classification), we encode the input $x_{1:T}$ with a encoding network, process the encoded sequences with the RNN, and make predictions with a prediction network. The encoding network is a simple fully connected layer with a output dimension of 128 for the classification tasks and 64 for regression tasks; the prediction network is also a fully connected layer whose output dimension is dependent on the specific tasks. For all fully connected layers except for those inside RNNs, we use the ReLU as activation function.\\
\textbf{Text classification details.} For the text classification tasks, we train a word embedding size of 100 from scratch for simplicity. Note that a more powerful pretrained word embedding could be used to improve the performance. We leave it for future study. The embedding is directly input to the RNNs, and the output is predicted using a fully connected layer with shape \{hidden\_dim, num\_labels\}.\\

\section{Additional Results}\label{sec:aexp}
\subsection{Comparison with RNN Ensemble Models and Bayesian RNNs}
The detailed results for the comparison with RNN ensemble models and Baeysian RNNs are given in Table~\ref{tab:ens_bayes}. LSTM-E/GRU-E denote the ensemble models of $K$ LSTMs/GRUs; BayesLSTM/BayesGRU denote the Bayesian LSTM and GRU. We compare them with PF-LSTM/PF-GRU with $K$ particles. For the localization experiments, i.e., Maze 10, Maze 18 and Maze 27 rows, we use $K=30$ ensemble models/particles. For the rest, we use $K=20$. 

\subsection{Ablation Study}
We present the detailed results for our ablation study over all datasets in Table~\ref{tab:ablation_details}.

\subsection{Visualization of Additional Examples for Localization}
We provide extra examples for the visualization of PF-LSTM particles below.
\label{sec:additional_localization}
\begin{figure*}[!htb]
	\centering
	\begin{tabular}{ccccc}
		\includegraphics[width=0.18\linewidth]{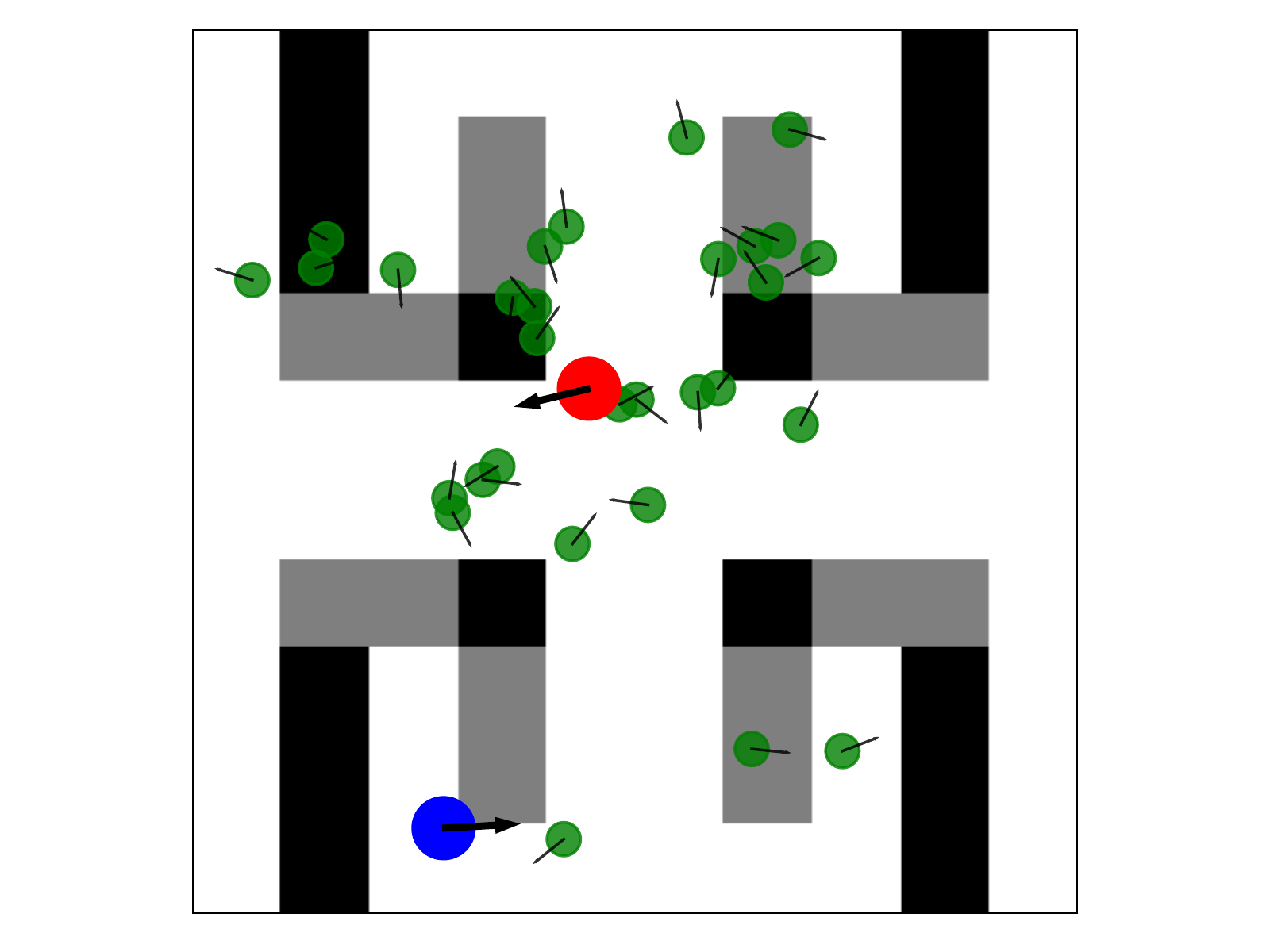} &
		\includegraphics[width=0.18\linewidth]{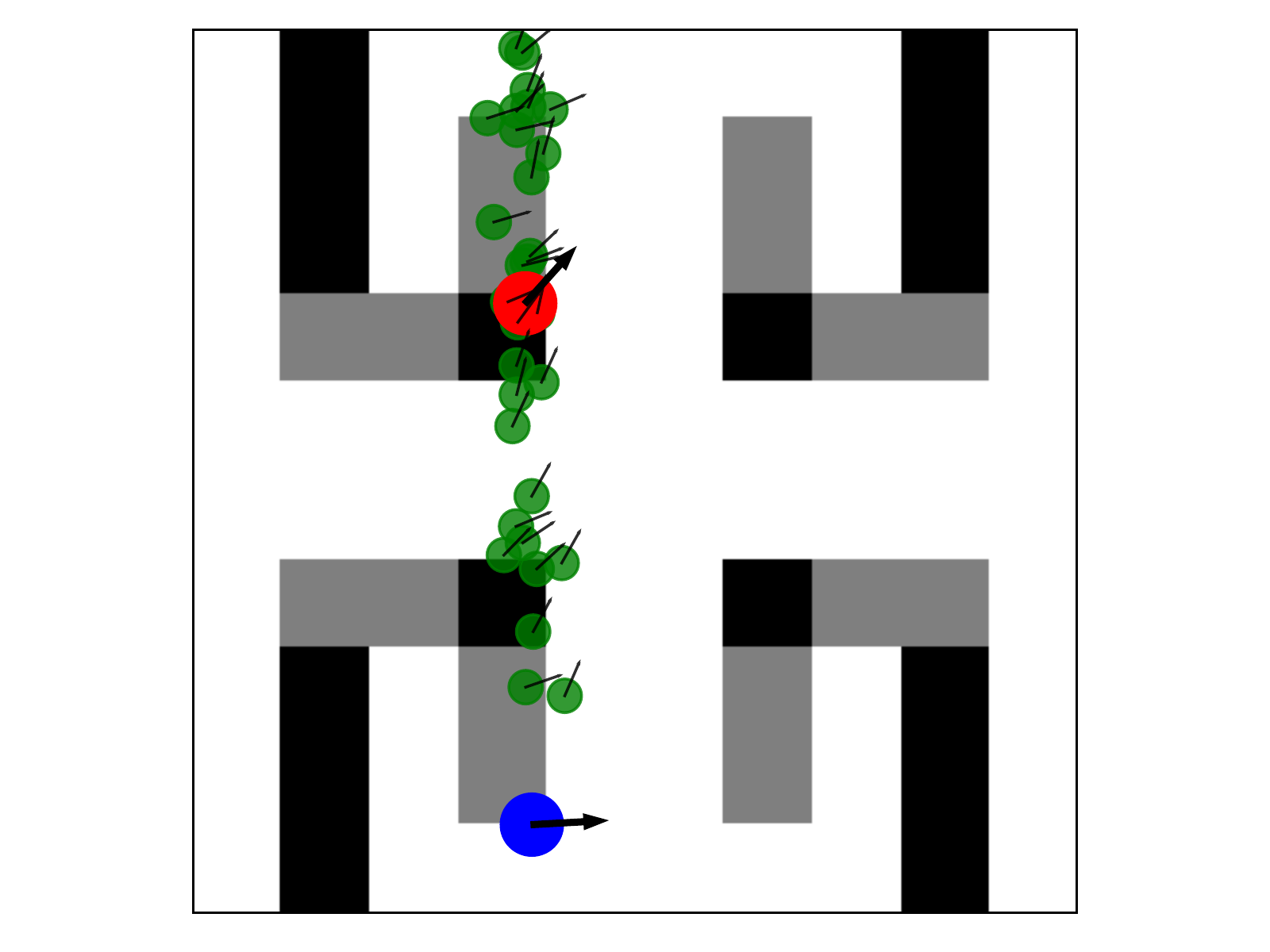} &
		\includegraphics[width=0.18\linewidth]{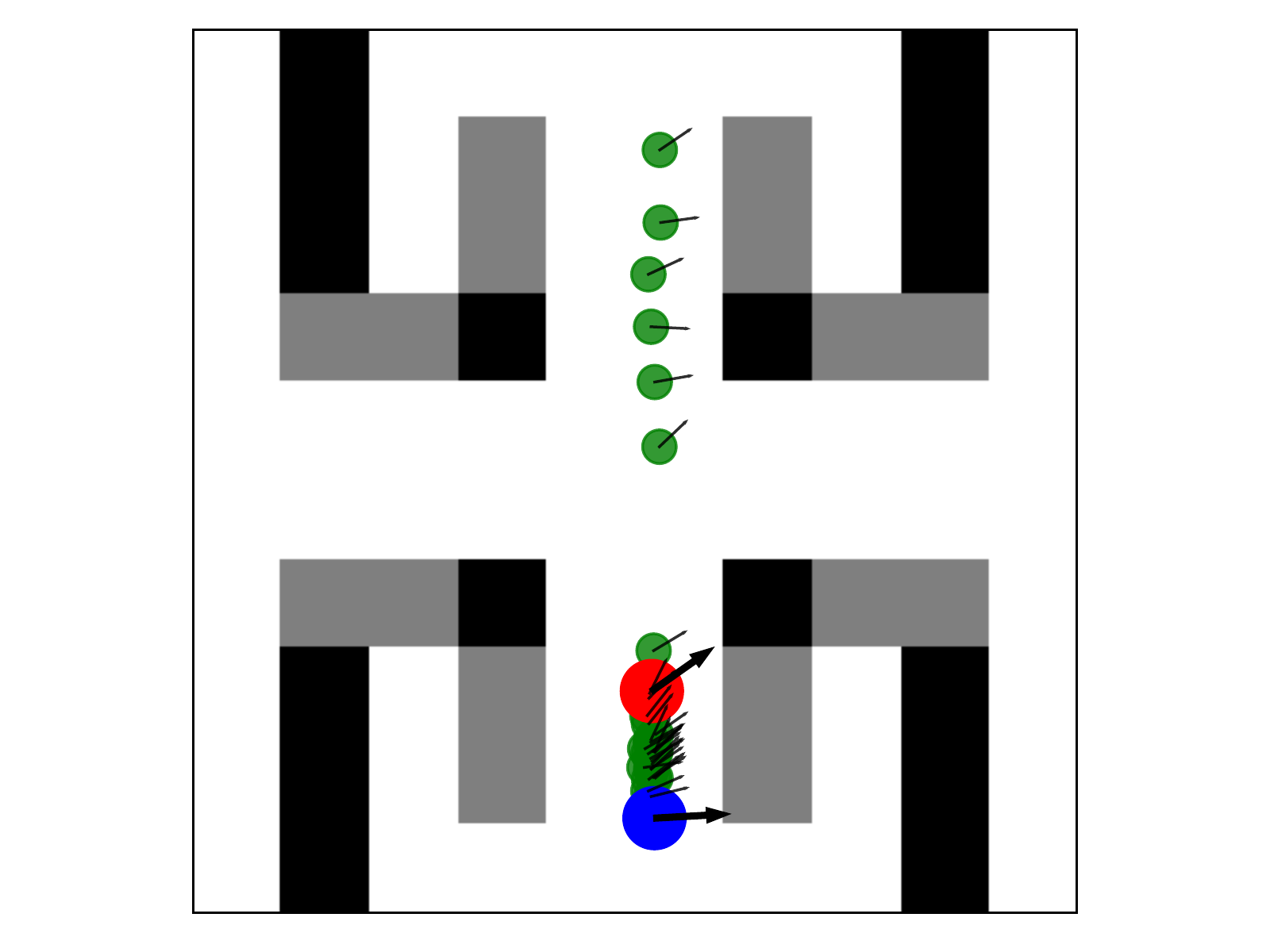} &
		\includegraphics[width=0.18\linewidth]{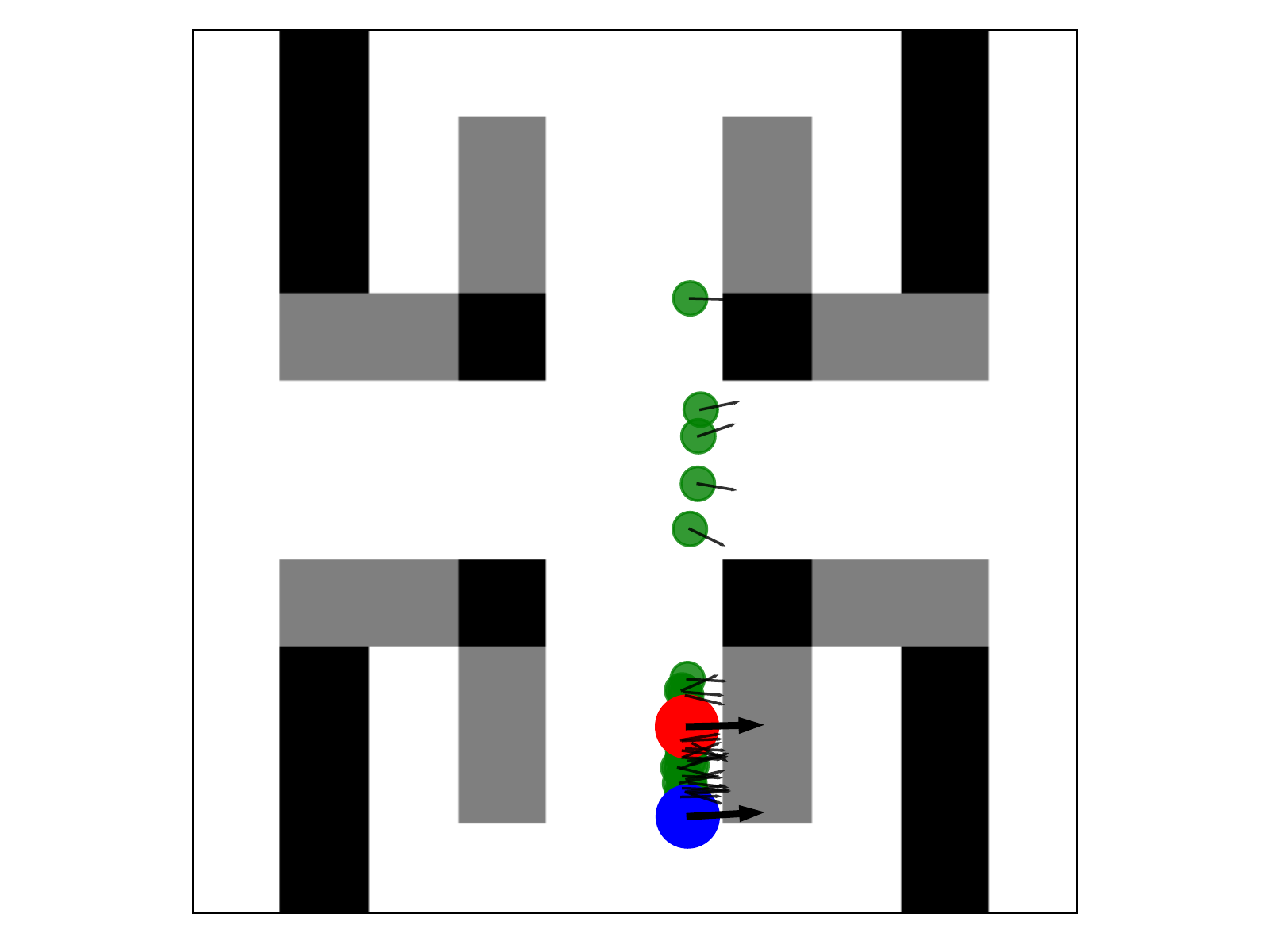} &
		\includegraphics[width=0.18\linewidth]{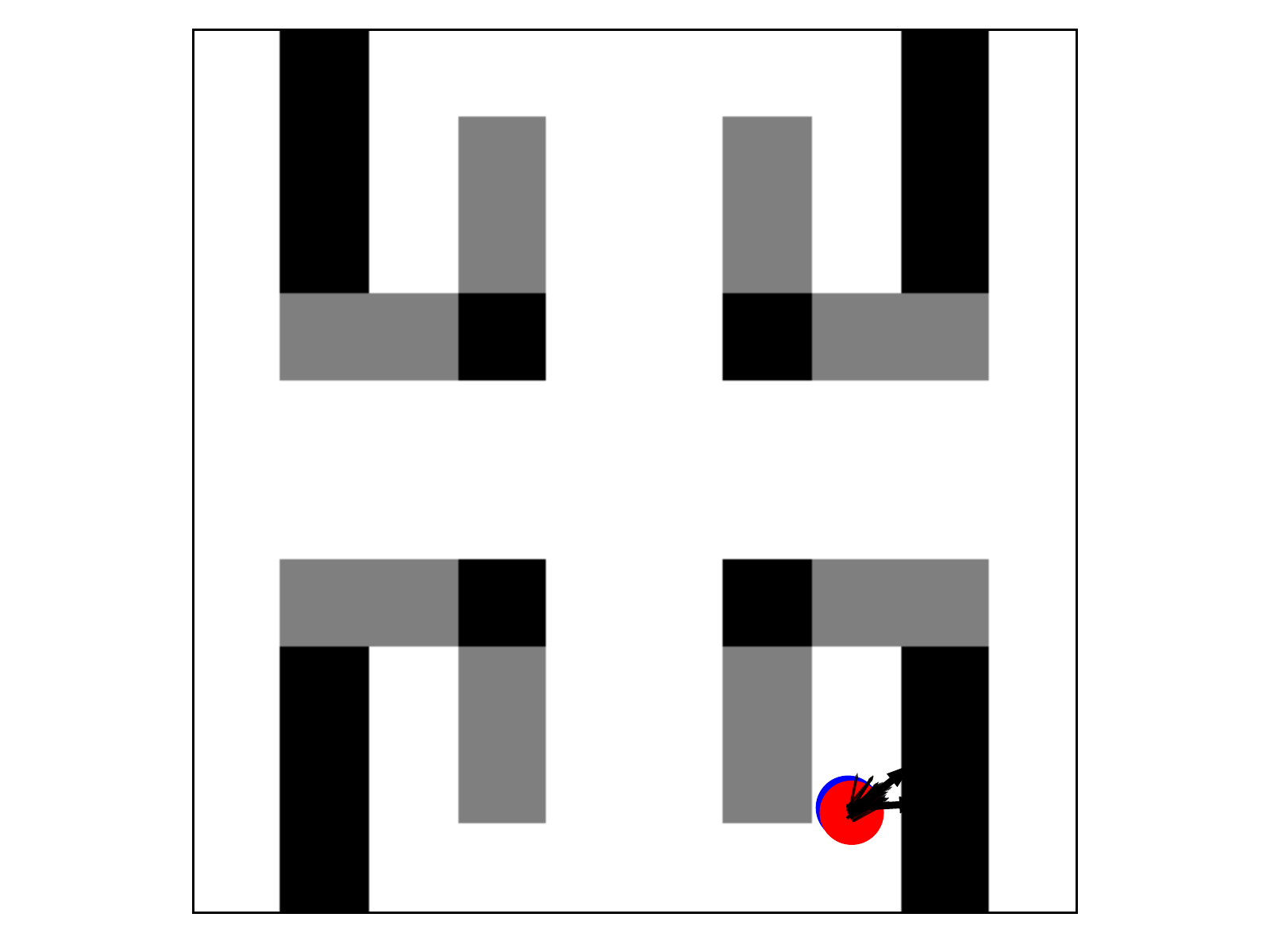} \\
		$t=0$ & $t=5$ & $t=10$ & $t=15$ & $t=22$
	\end{tabular}
\end{figure*}
\begin{figure*}[!htb]
	\centering
	\begin{tabular}{ccccc}
		\includegraphics[width=0.18\linewidth]{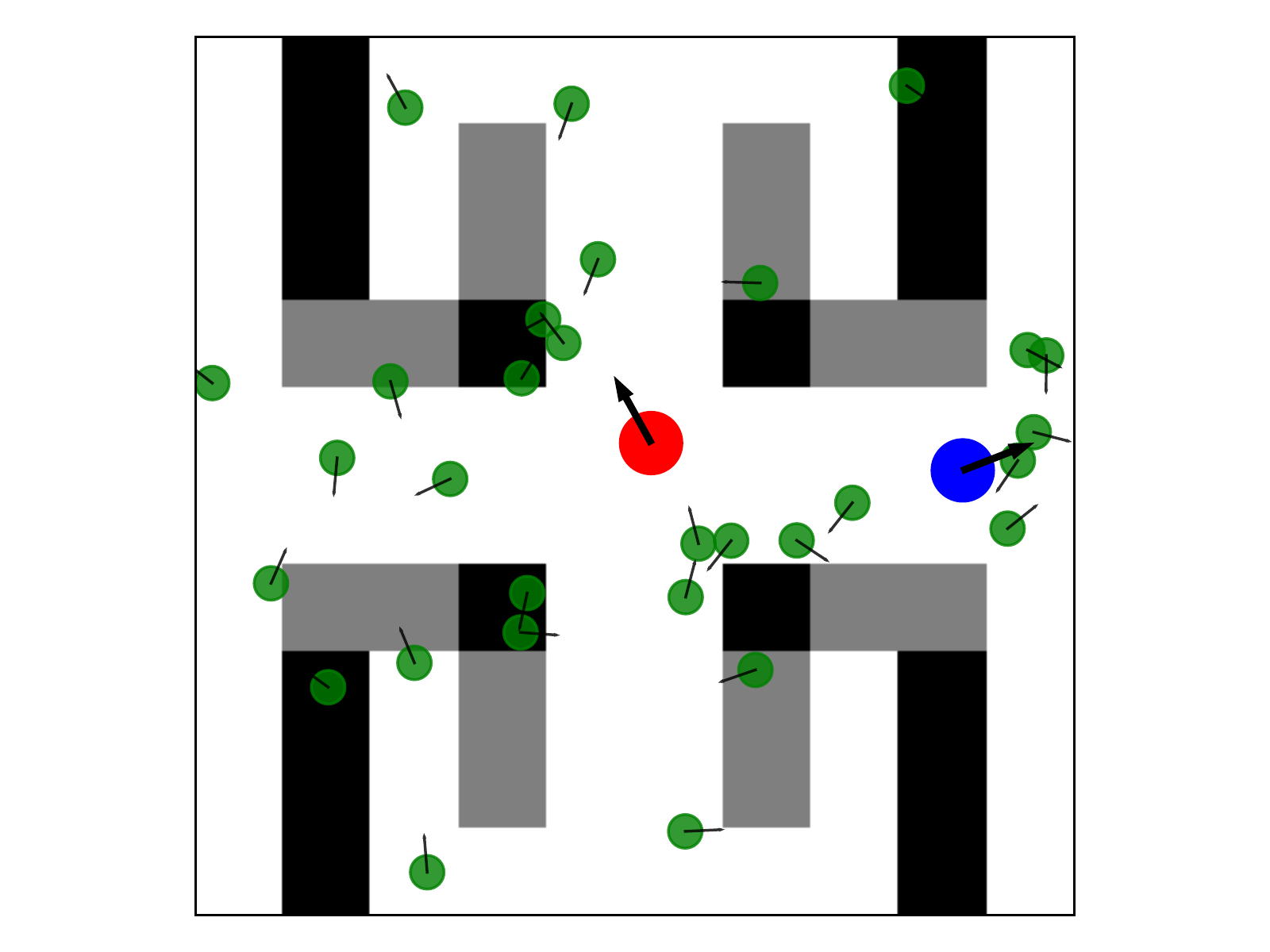} &
		\includegraphics[width=0.18\linewidth]{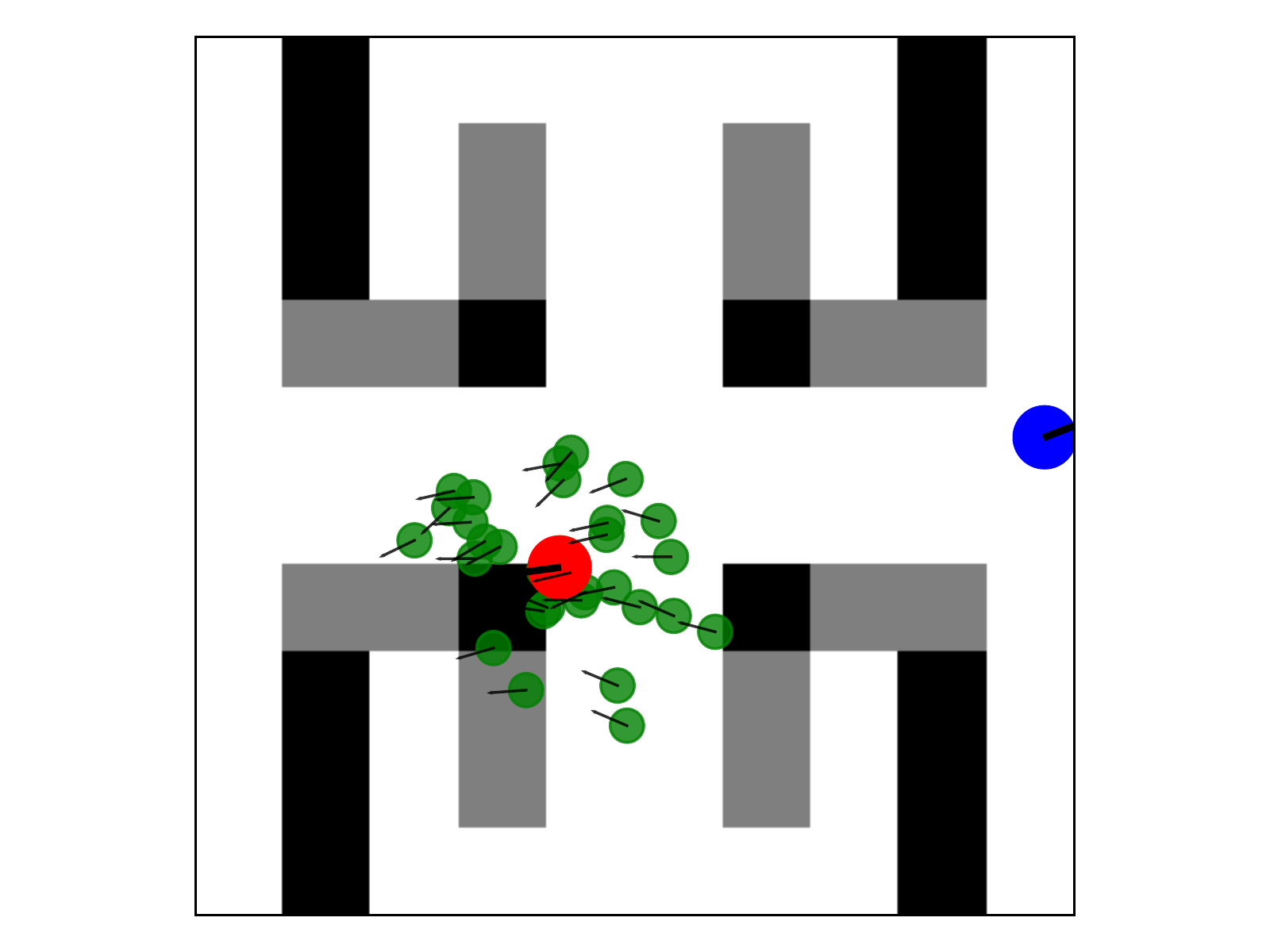} &
		\includegraphics[width=0.18\linewidth]{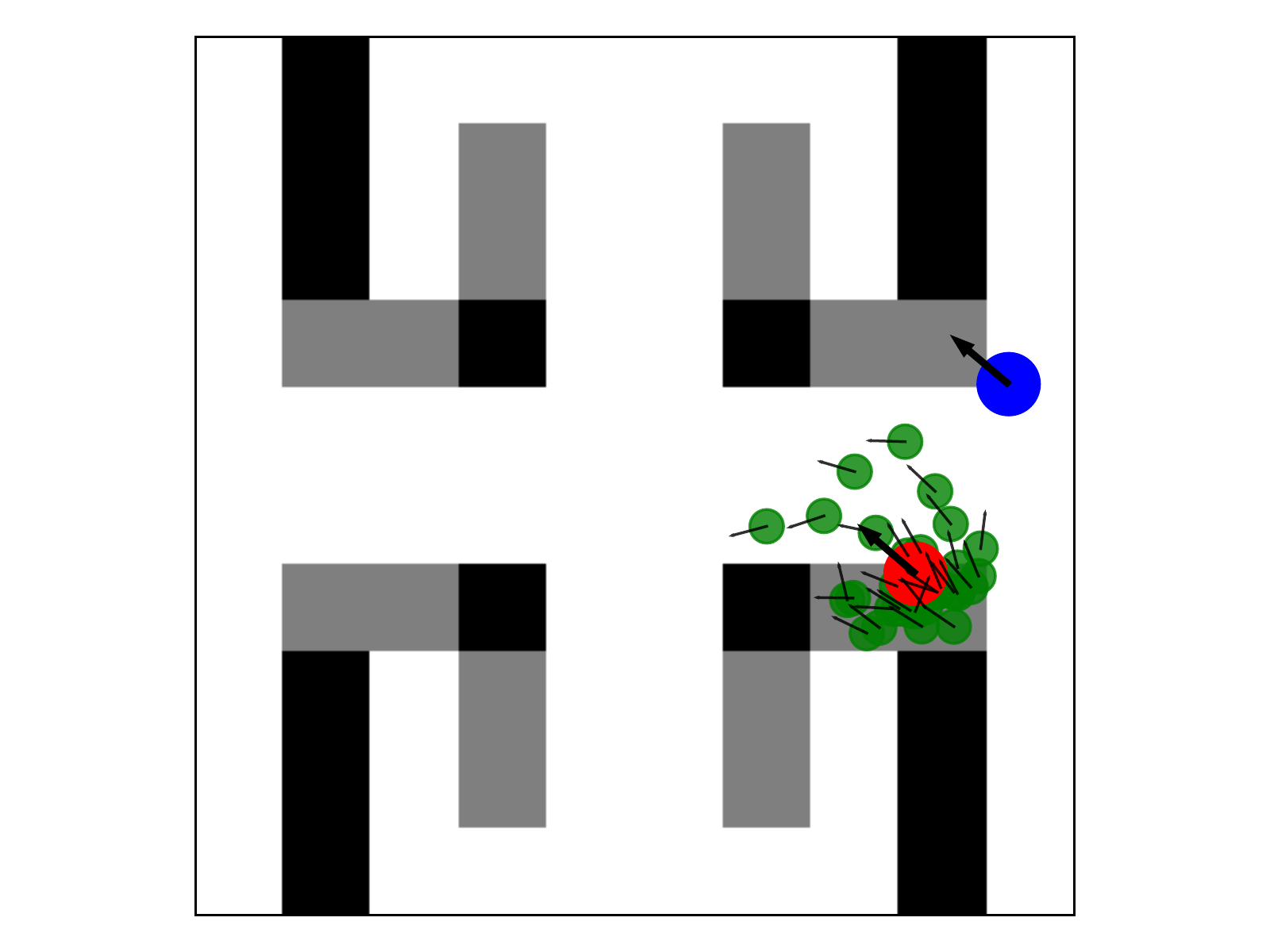} &
		\includegraphics[width=0.18\linewidth]{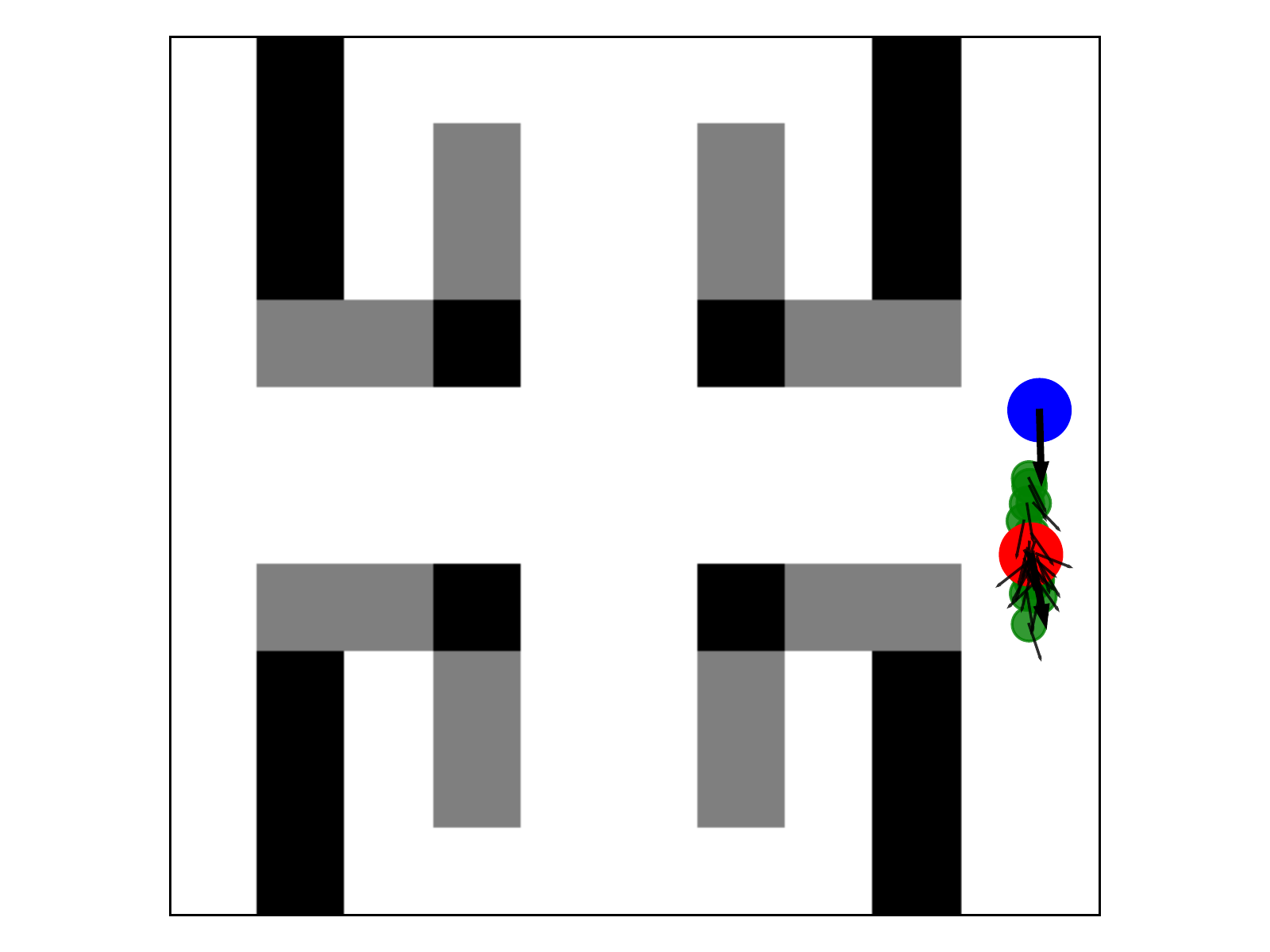} &
		\includegraphics[width=0.18\linewidth]{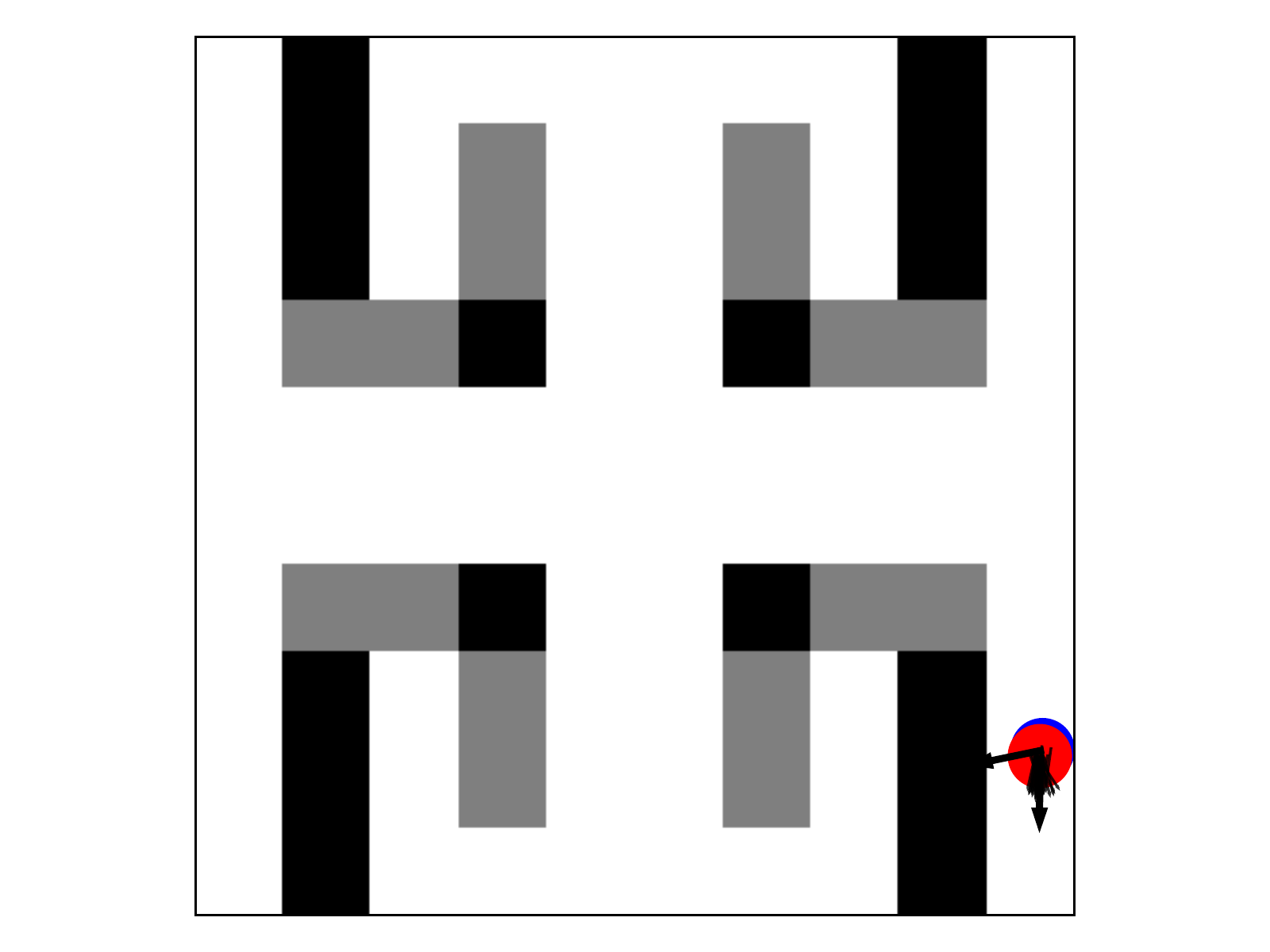} \\
		$t=0$ & $t=5$ & $t=10$ & $t=16$ & $t=25$
	\end{tabular}
\end{figure*}
\begin{figure*}[!htb]
	\centering
	\begin{tabular}{ccccc}
		\includegraphics[width=0.18\linewidth]{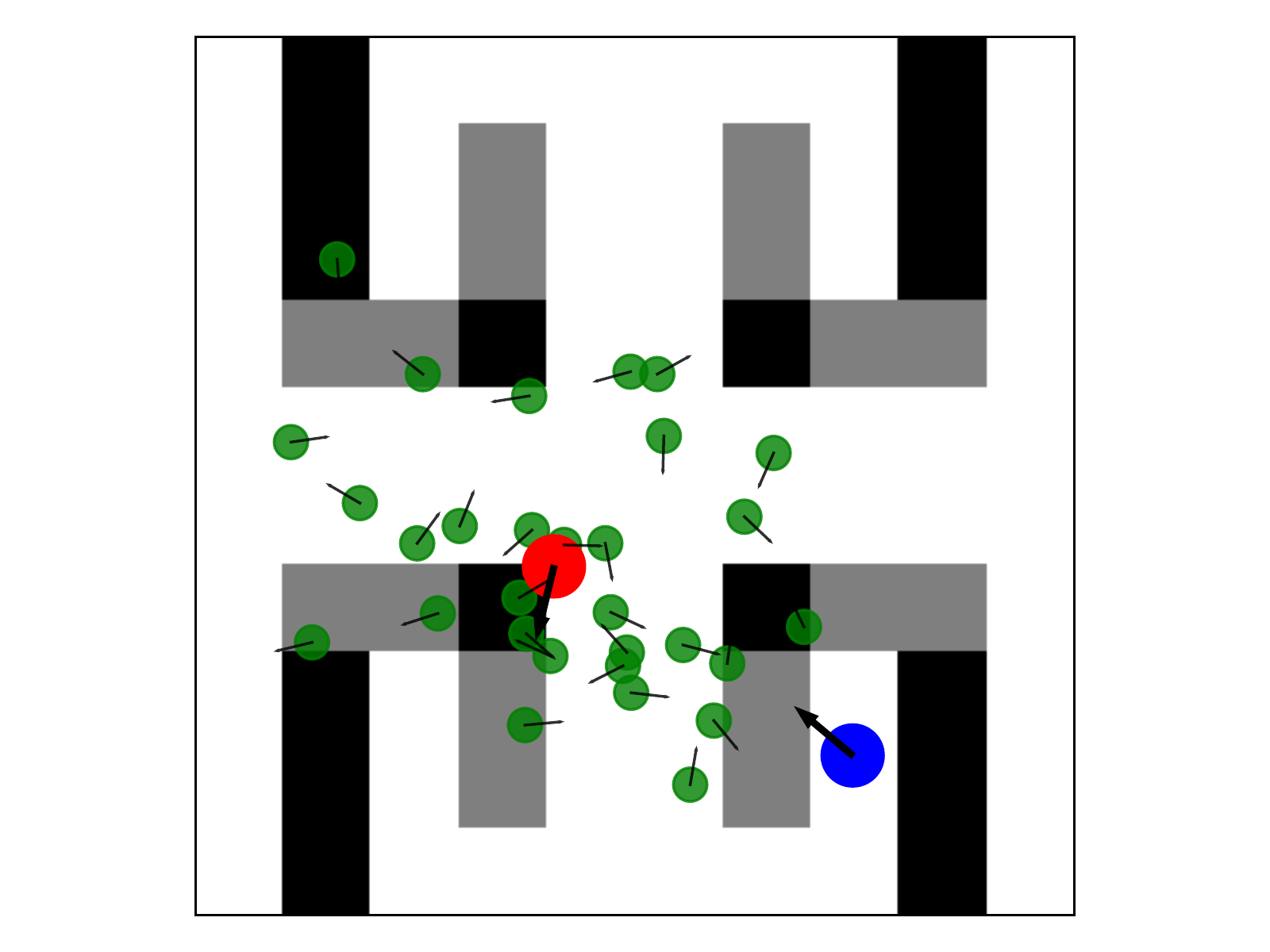} &
		\includegraphics[width=0.18\linewidth]{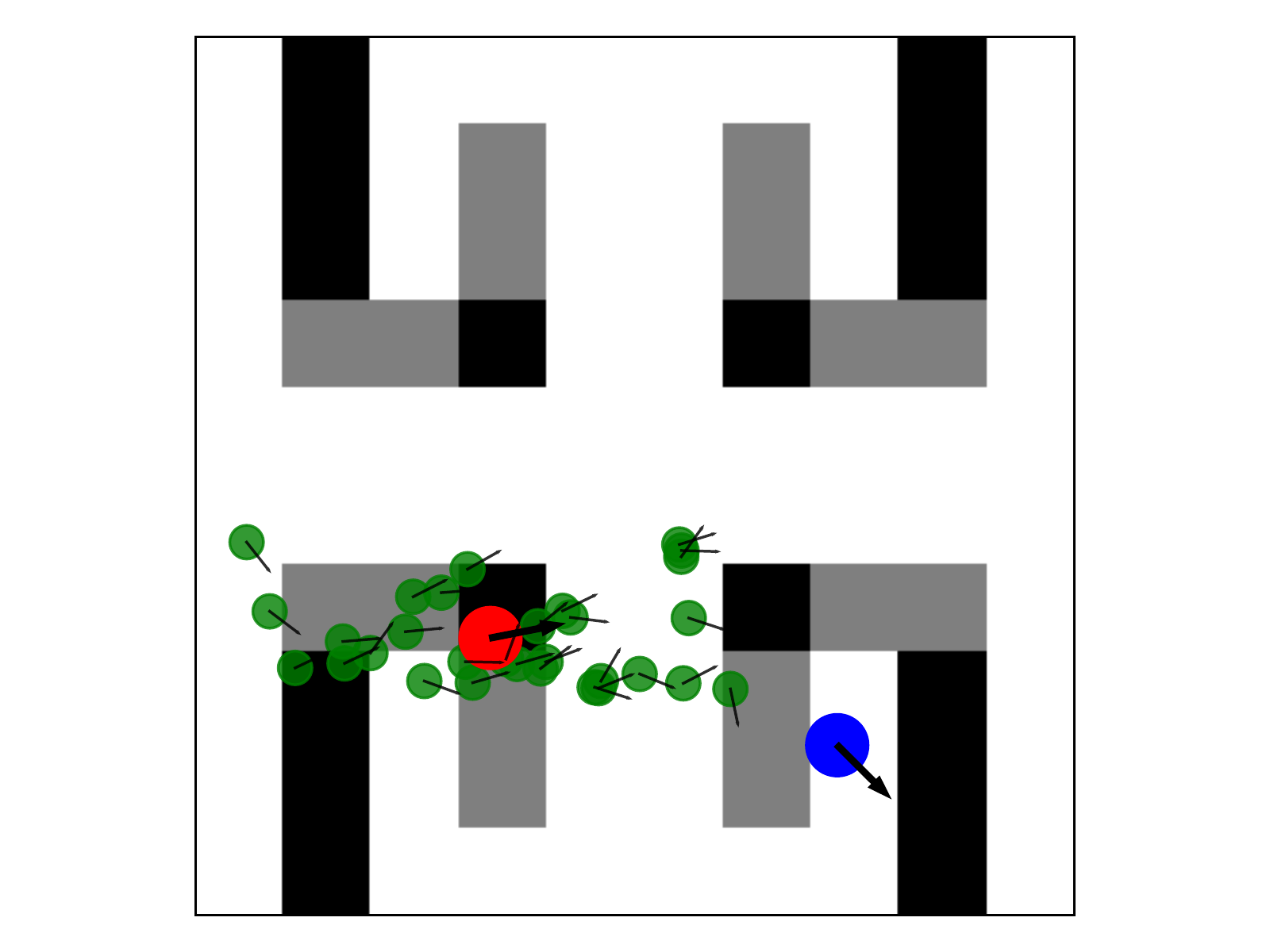} &
		\includegraphics[width=0.18\linewidth]{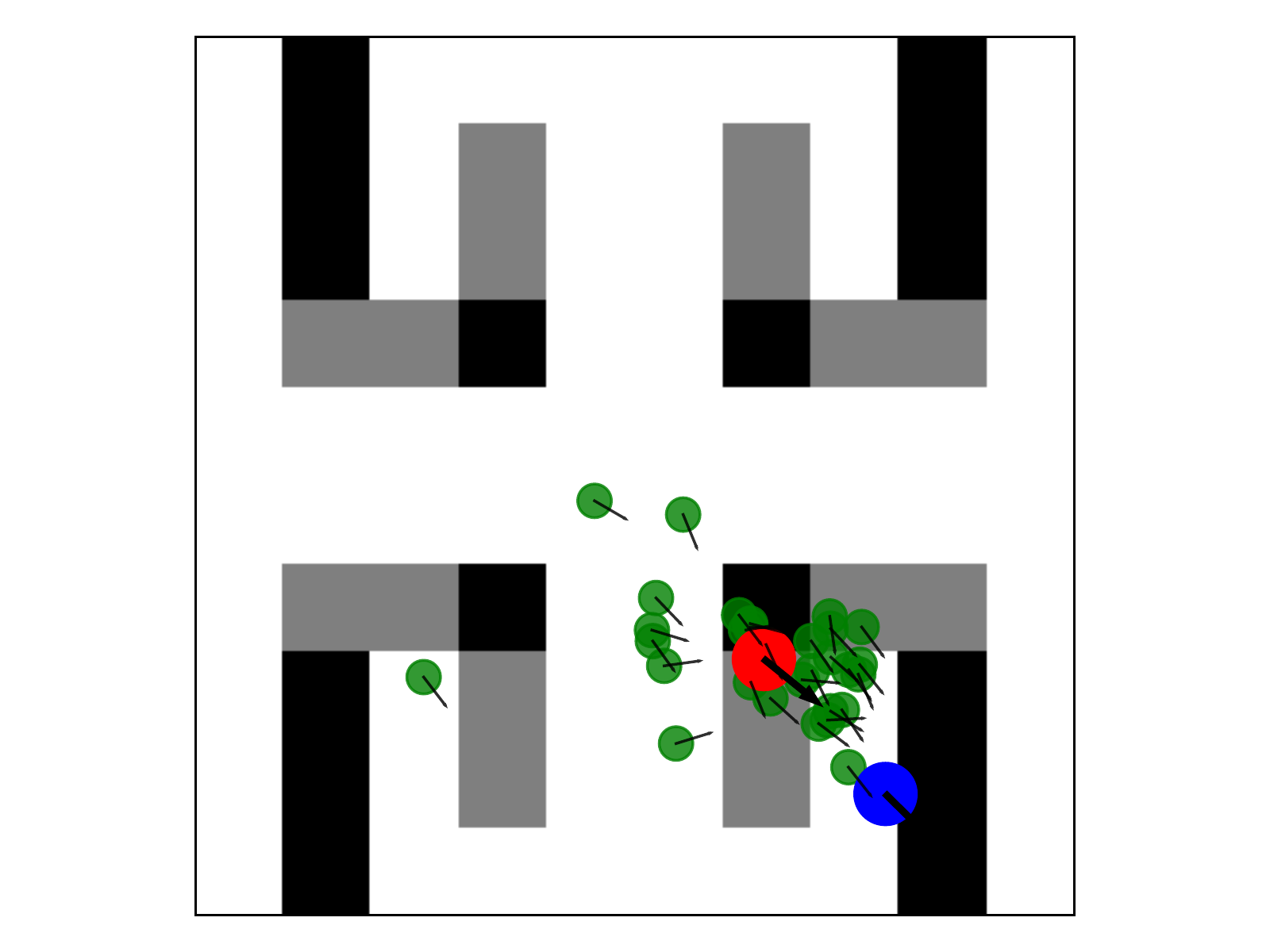} &
		\includegraphics[width=0.18\linewidth]{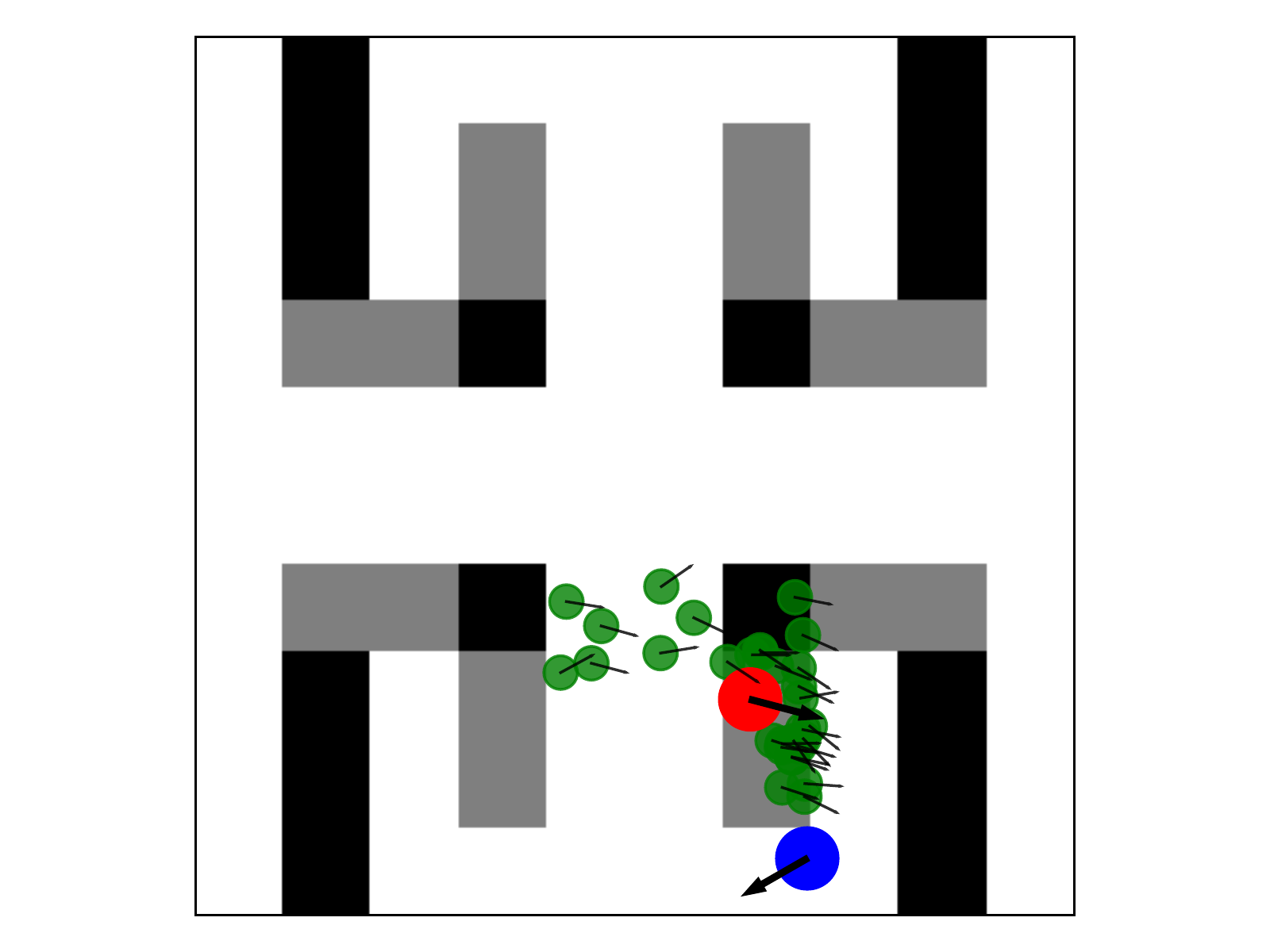} &
		\includegraphics[width=0.18\linewidth]{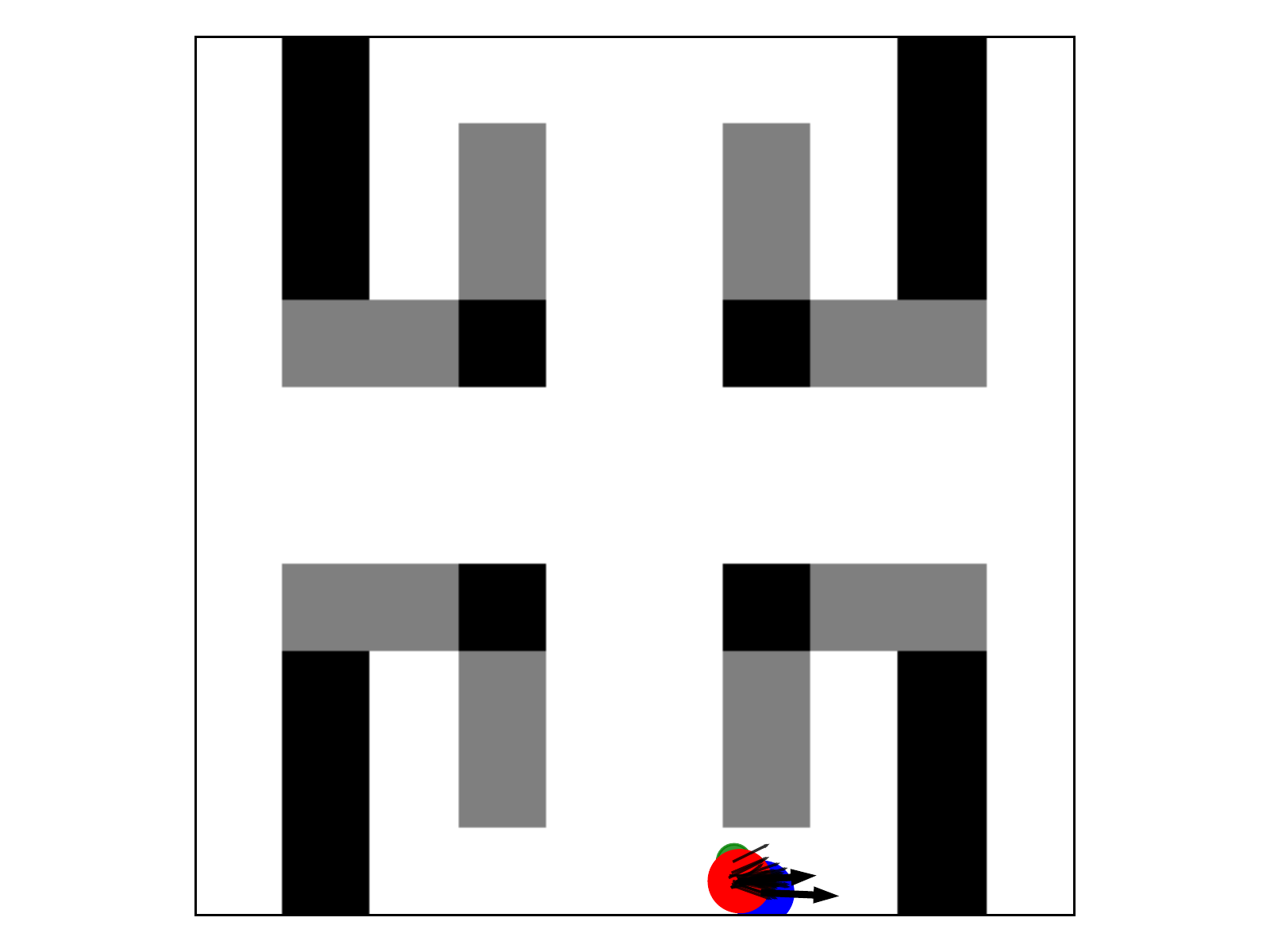} \\
		$t=0$ & $t=5$ & $t=9$ & $t=16$ & $t=21$
	\end{tabular}
\end{figure*}
\begin{figure*}[!htb]
	\centering
	\begin{tabular}{ccccc}
		\includegraphics[width=0.18\linewidth]{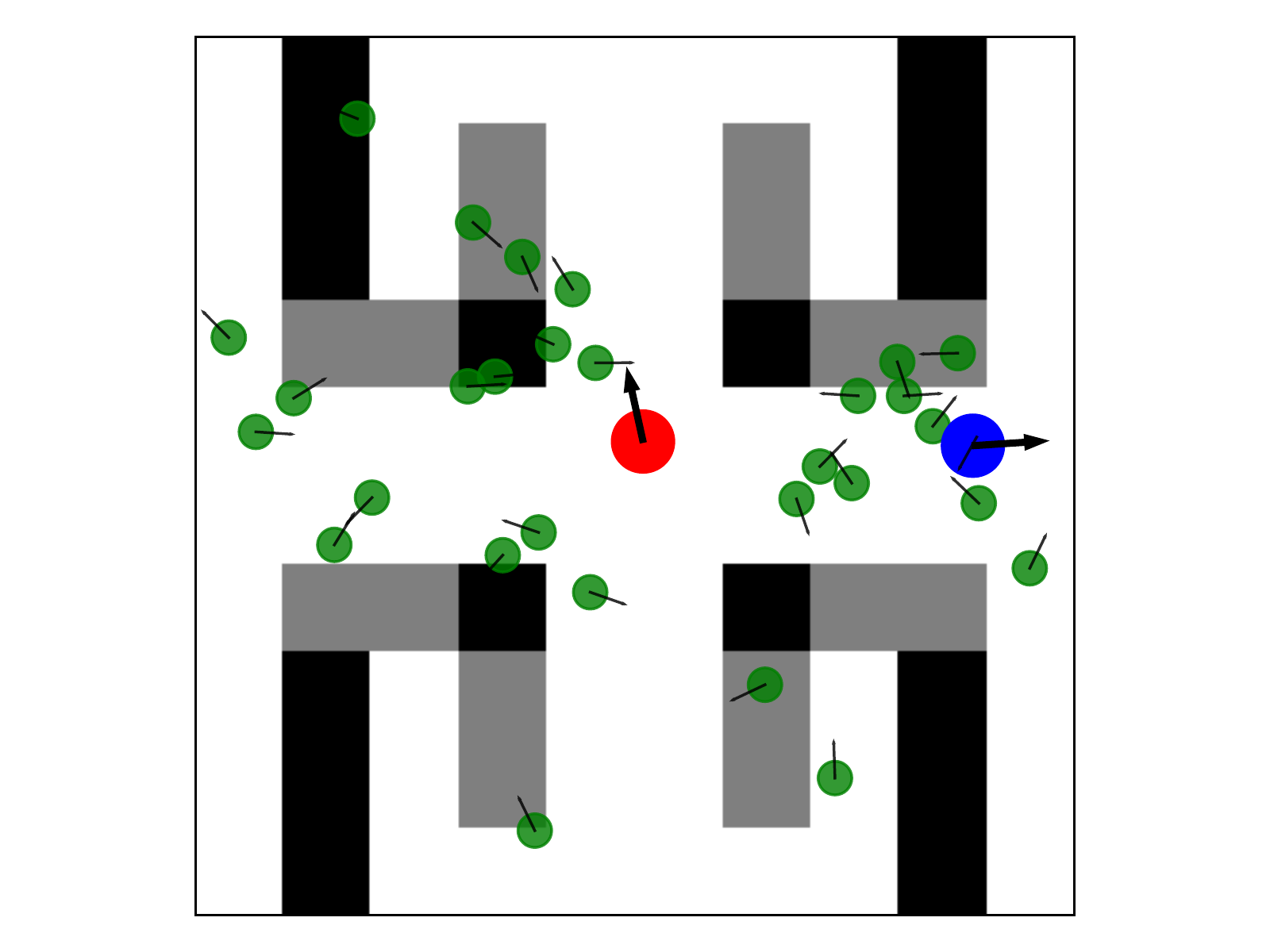} &
		\includegraphics[width=0.18\linewidth]{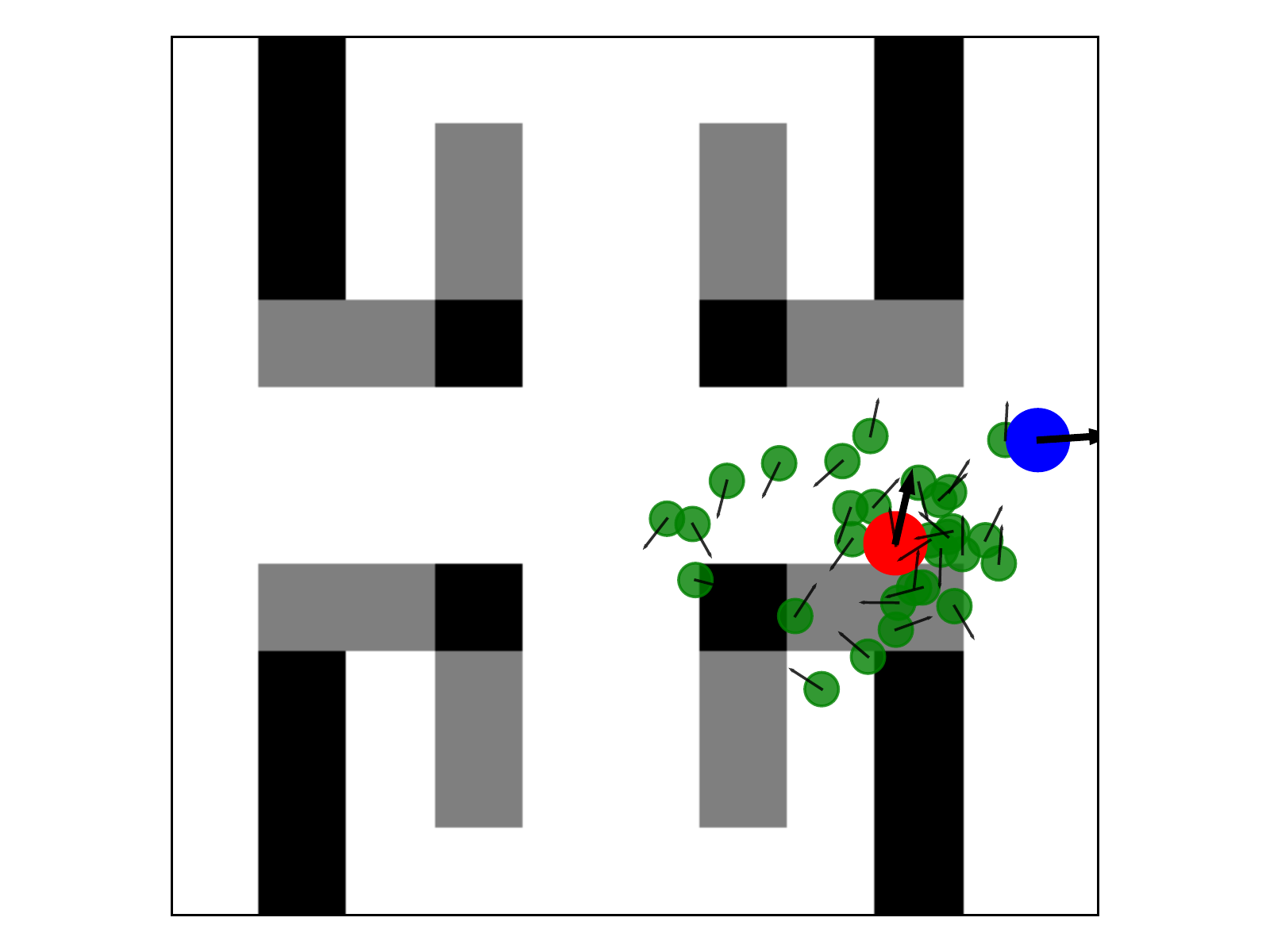} &
		\includegraphics[width=0.18\linewidth]{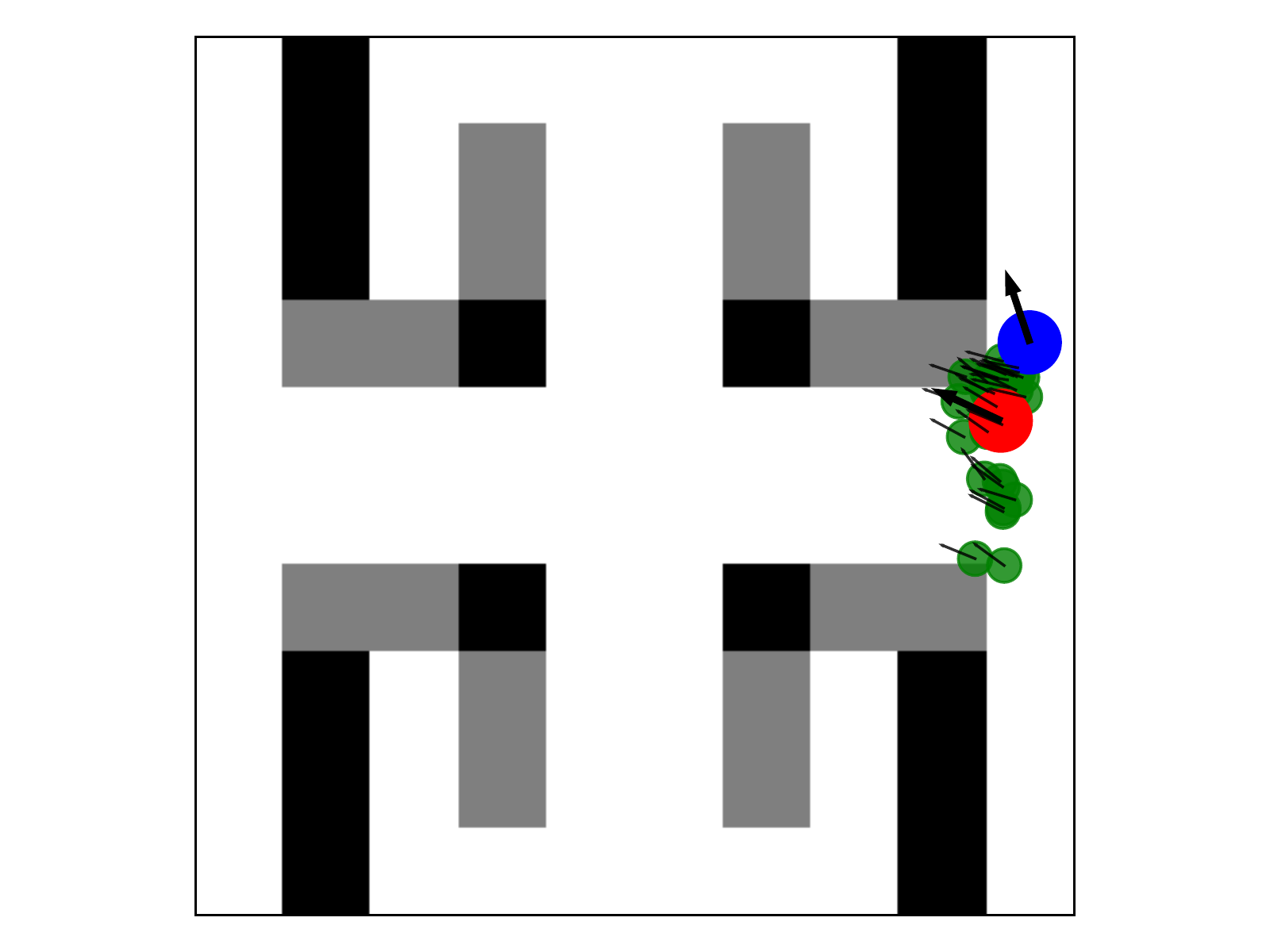} &
		\includegraphics[width=0.18\linewidth]{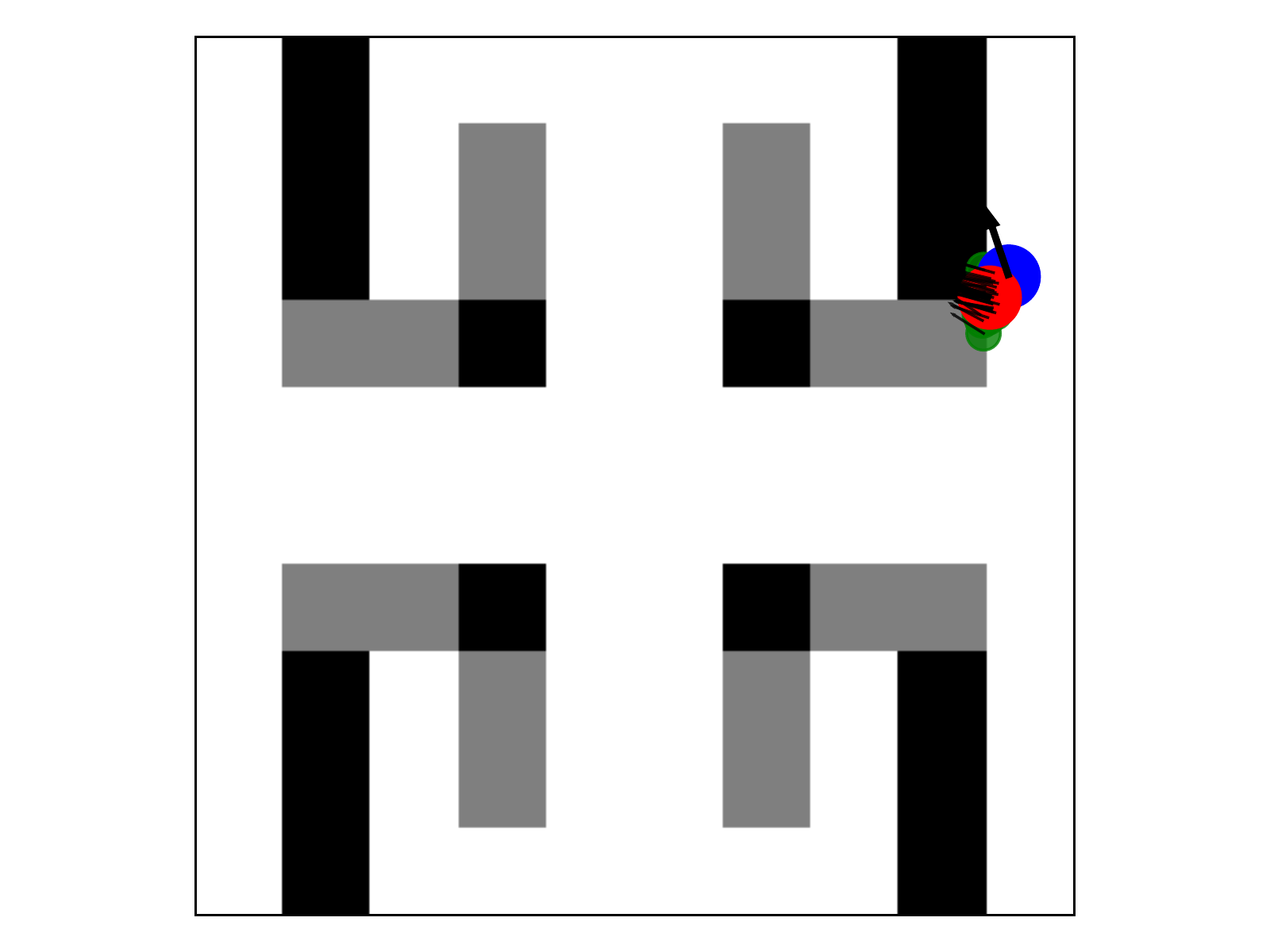} &
		\includegraphics[width=0.18\linewidth]{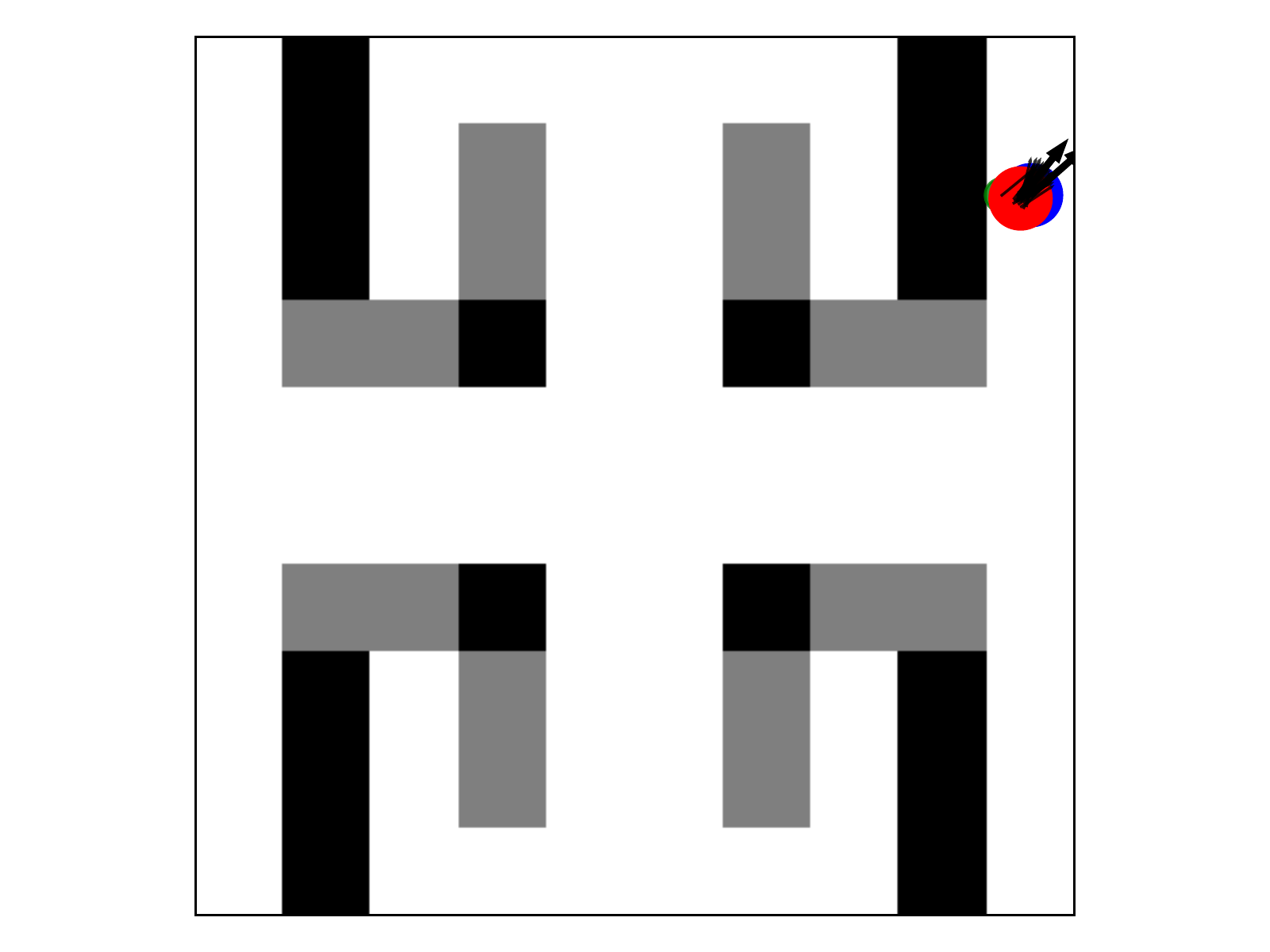} \\
		$t=0$ & $t=5$ & $t=11$ & $t=15$ & $t=21$
	\end{tabular}
\end{figure*}
\begin{figure*}[!htb]
	\centering
	\begin{tabular}{ccccc}
		\includegraphics[width=0.18\linewidth]{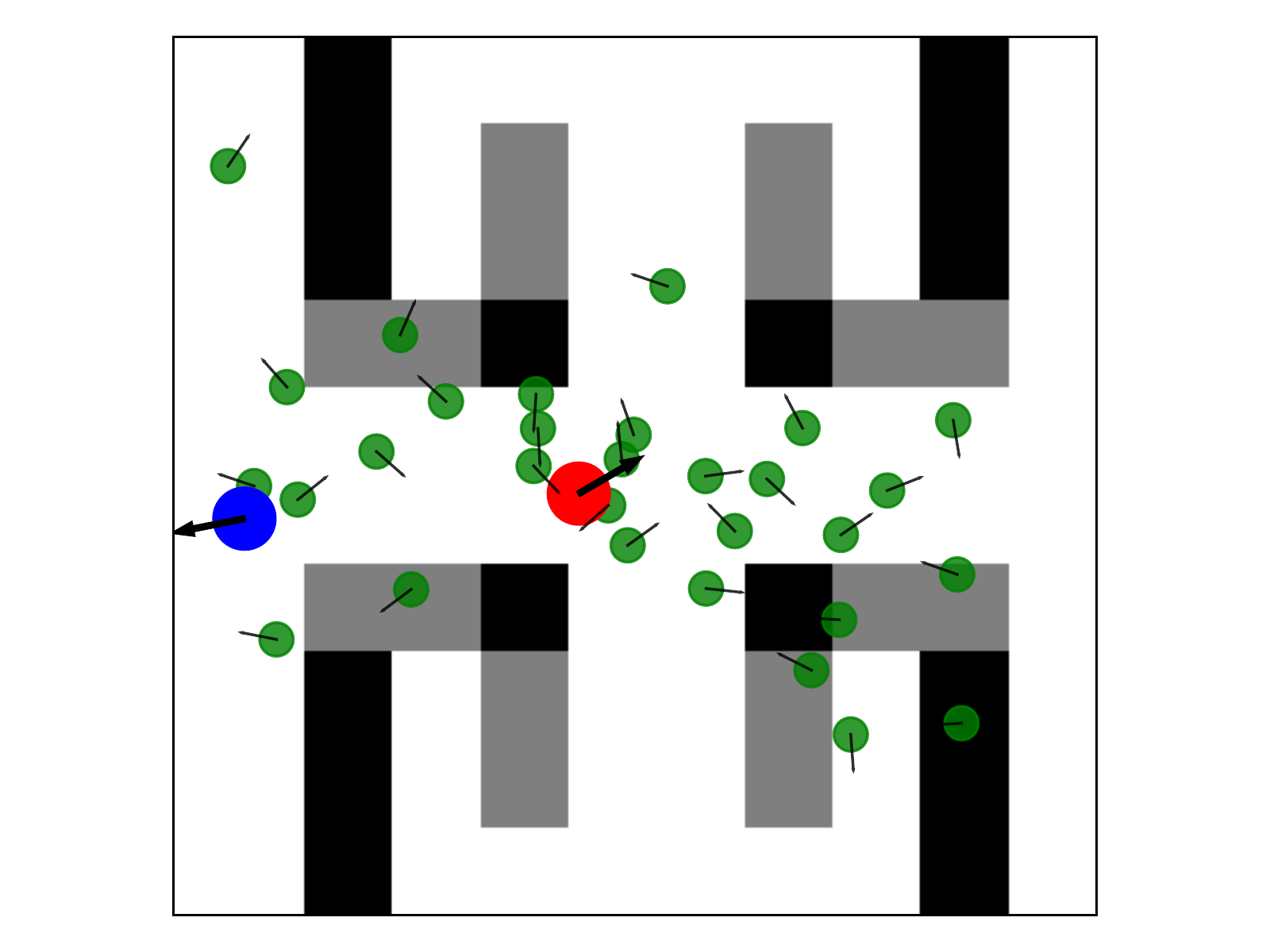} &
		\includegraphics[width=0.18\linewidth]{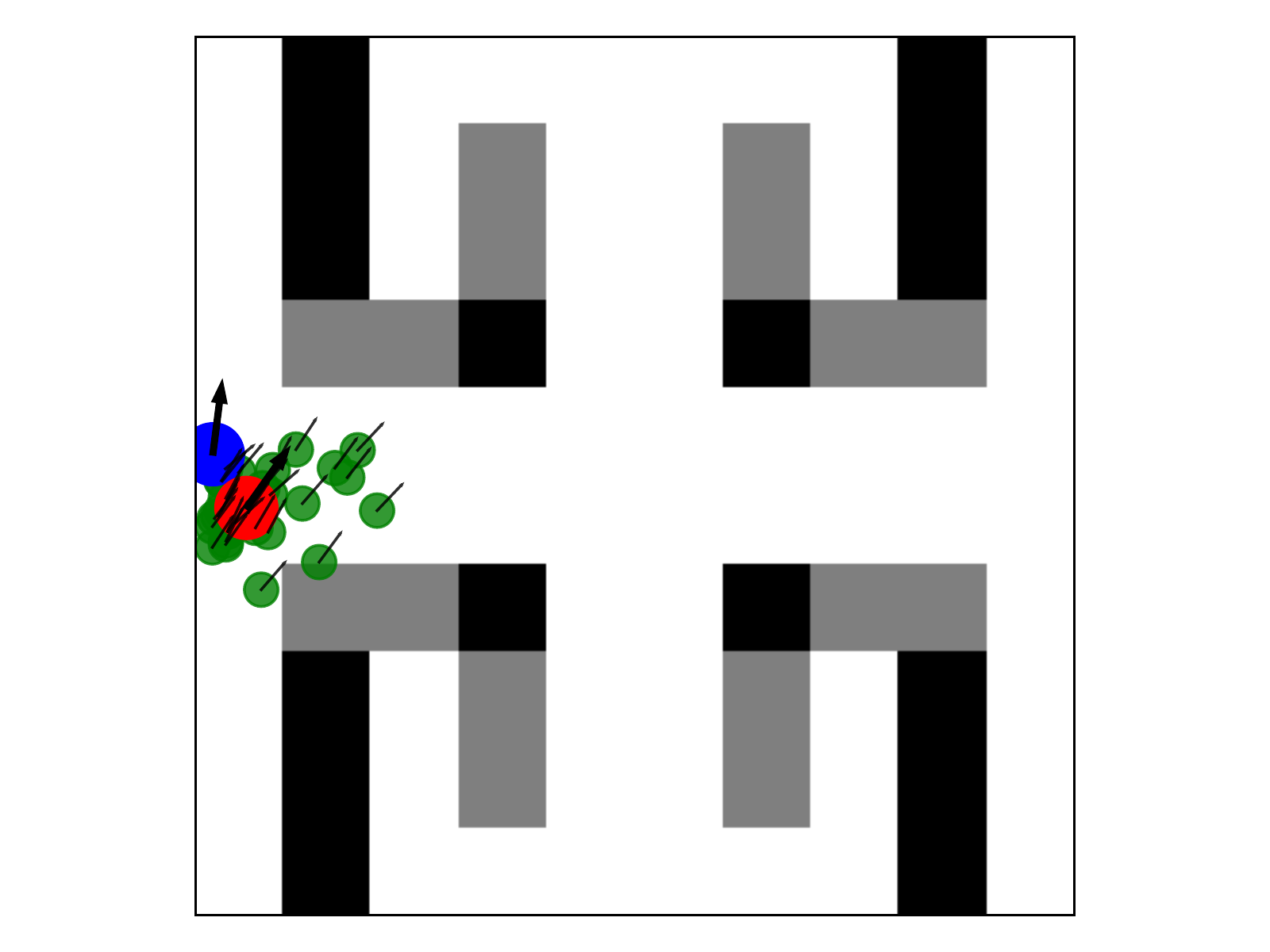} &
		\includegraphics[width=0.18\linewidth]{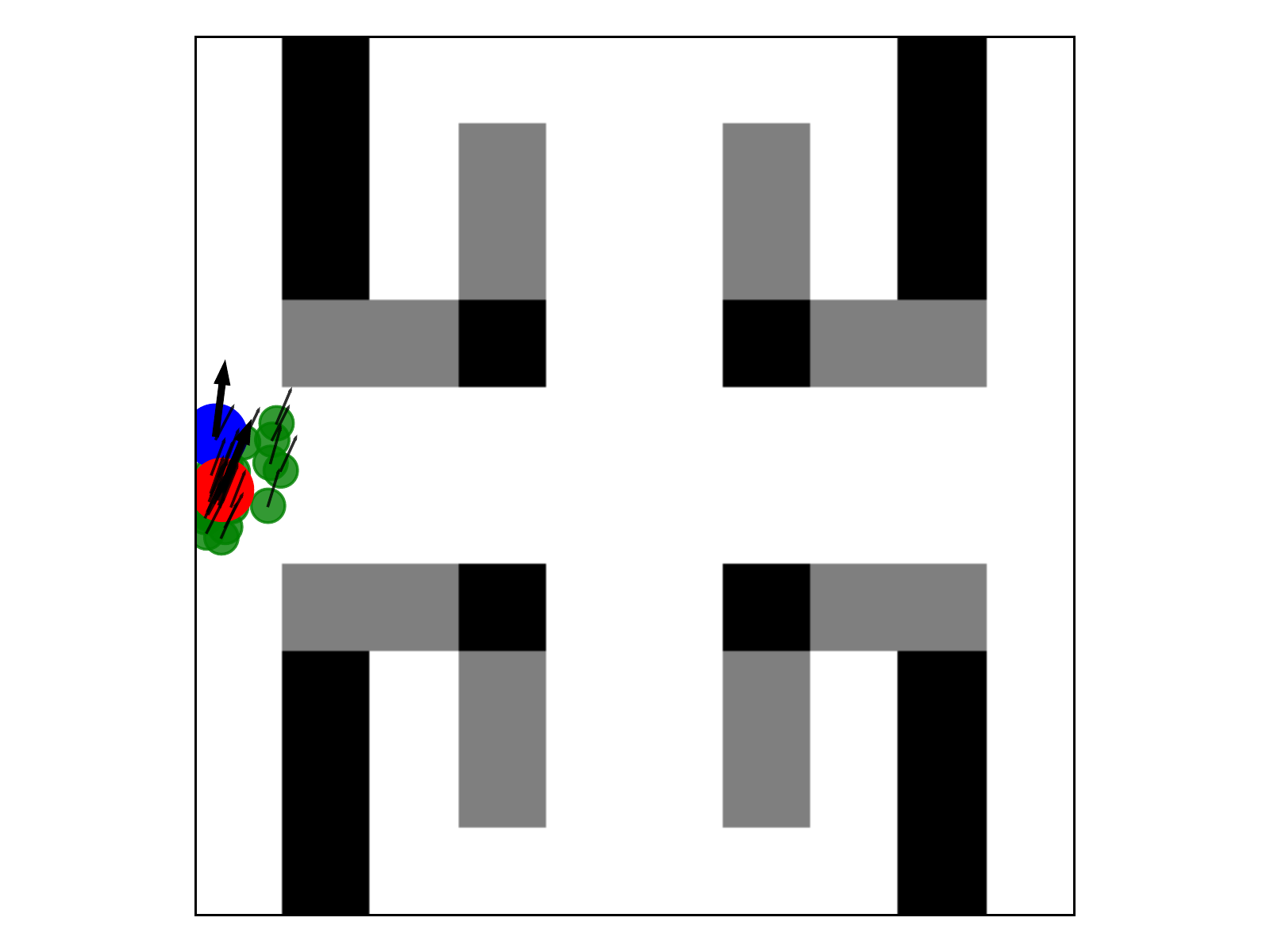} &
		\includegraphics[width=0.18\linewidth]{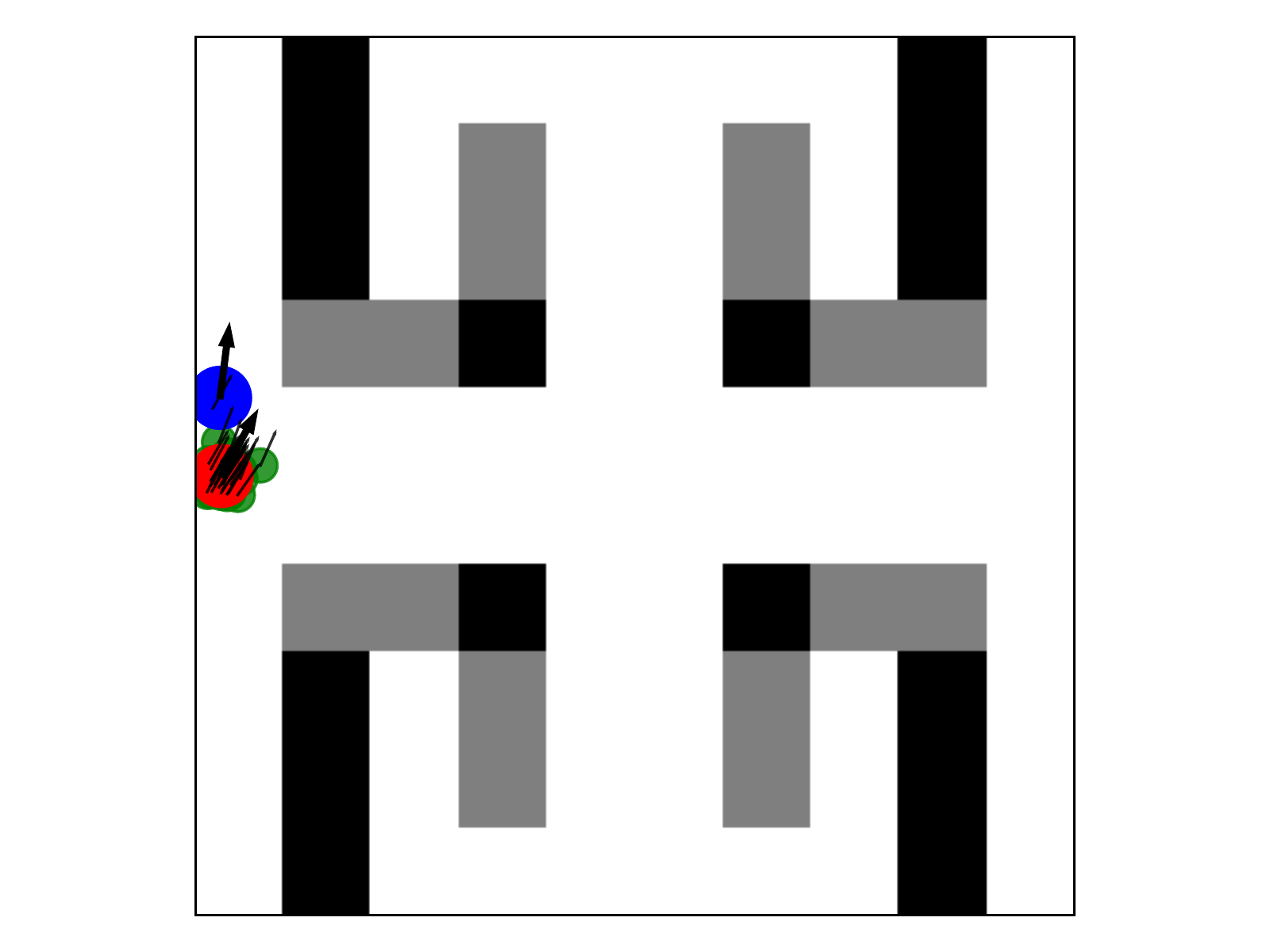} &
		\includegraphics[width=0.18\linewidth]{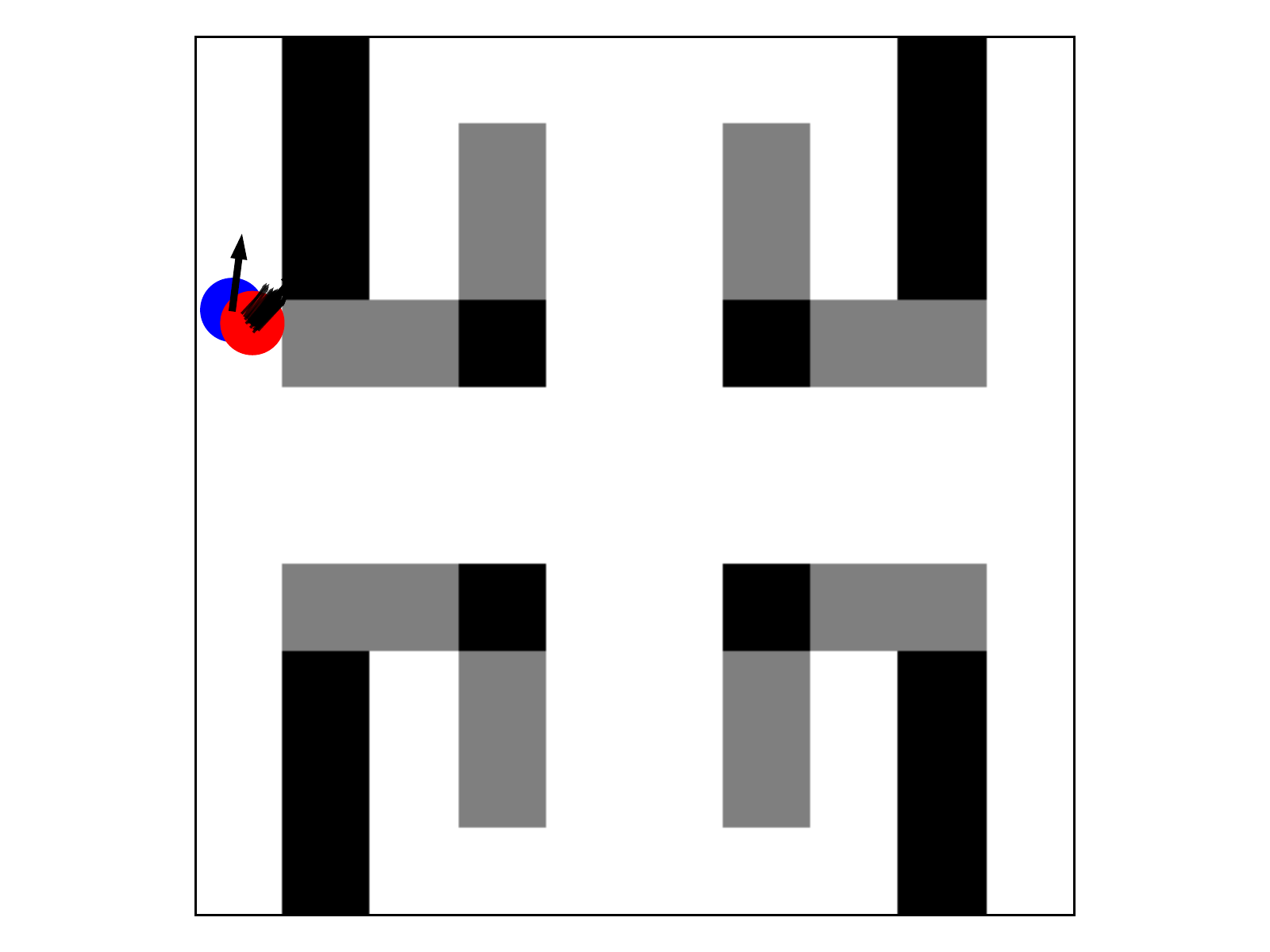} \\
		$t=0$ & $t=5$ & $t=6$ & $t=8$ & $t=13$
	\end{tabular}
\end{figure*}

\begin{figure*}[t]
	\centering
	\begin{tabular}{ccccc}
		\includegraphics[width=0.18\linewidth]{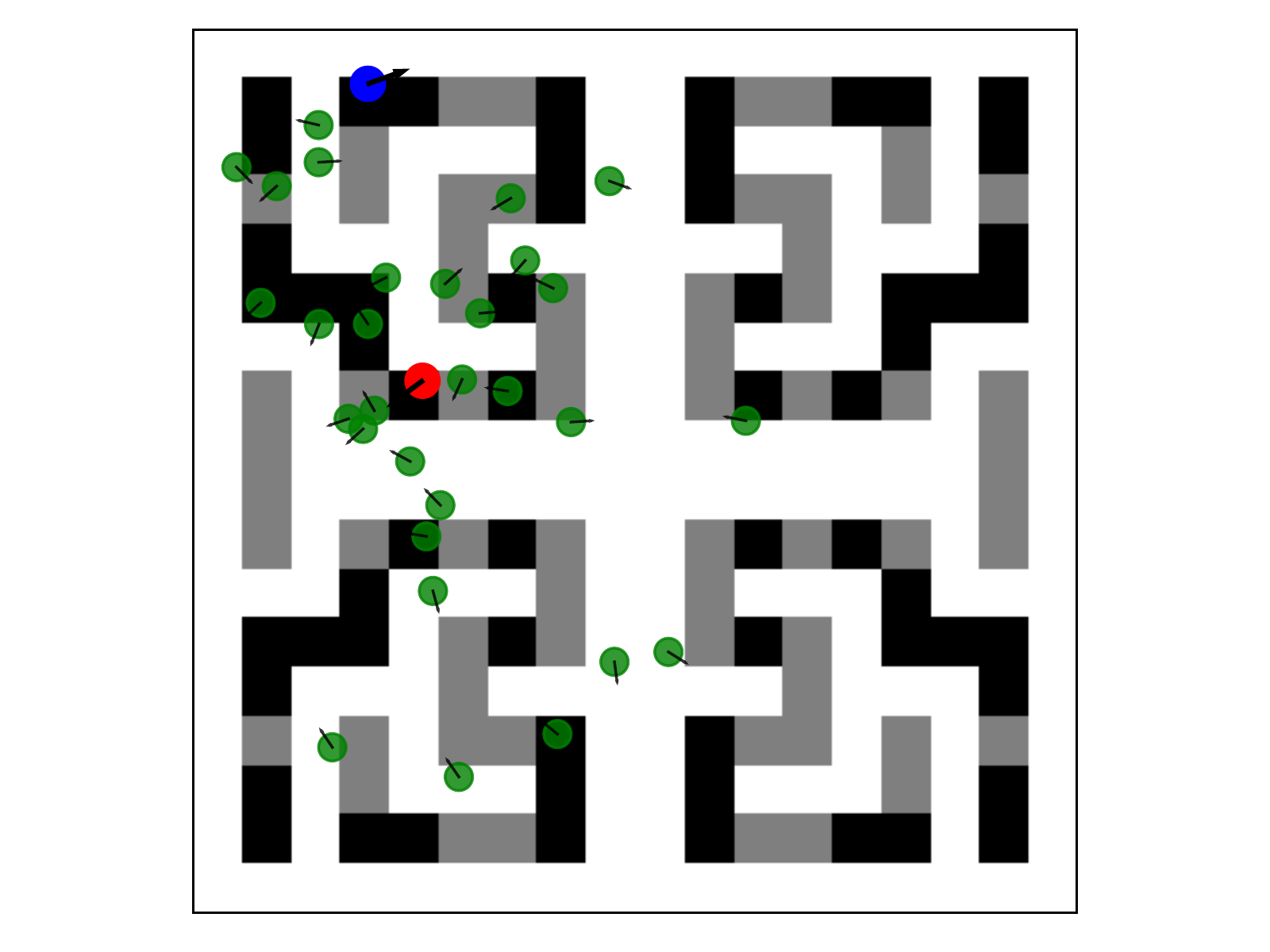} &
		\includegraphics[width=0.18\linewidth]{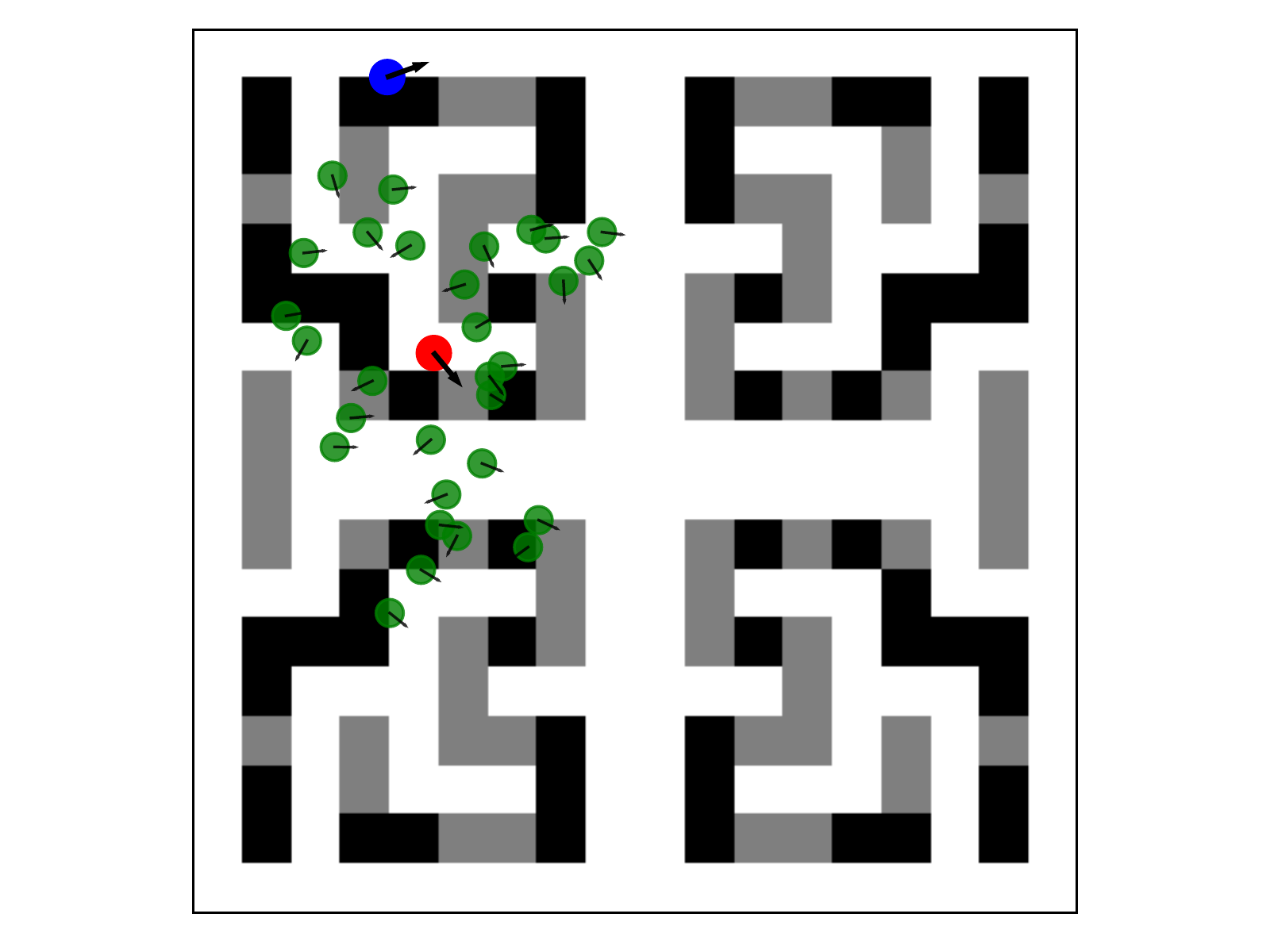} &
		\includegraphics[width=0.18\linewidth]{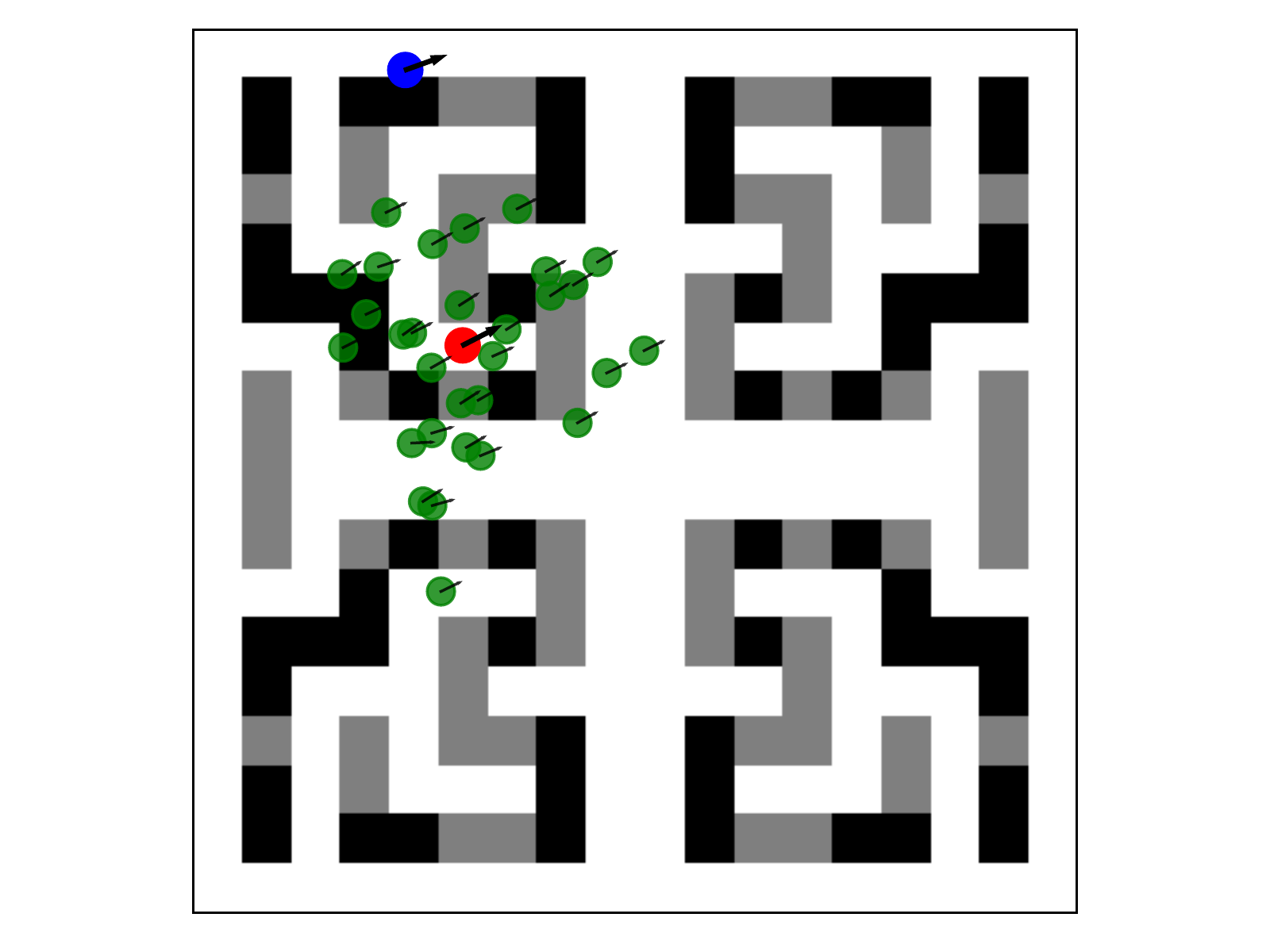} &
		\includegraphics[width=0.18\linewidth]{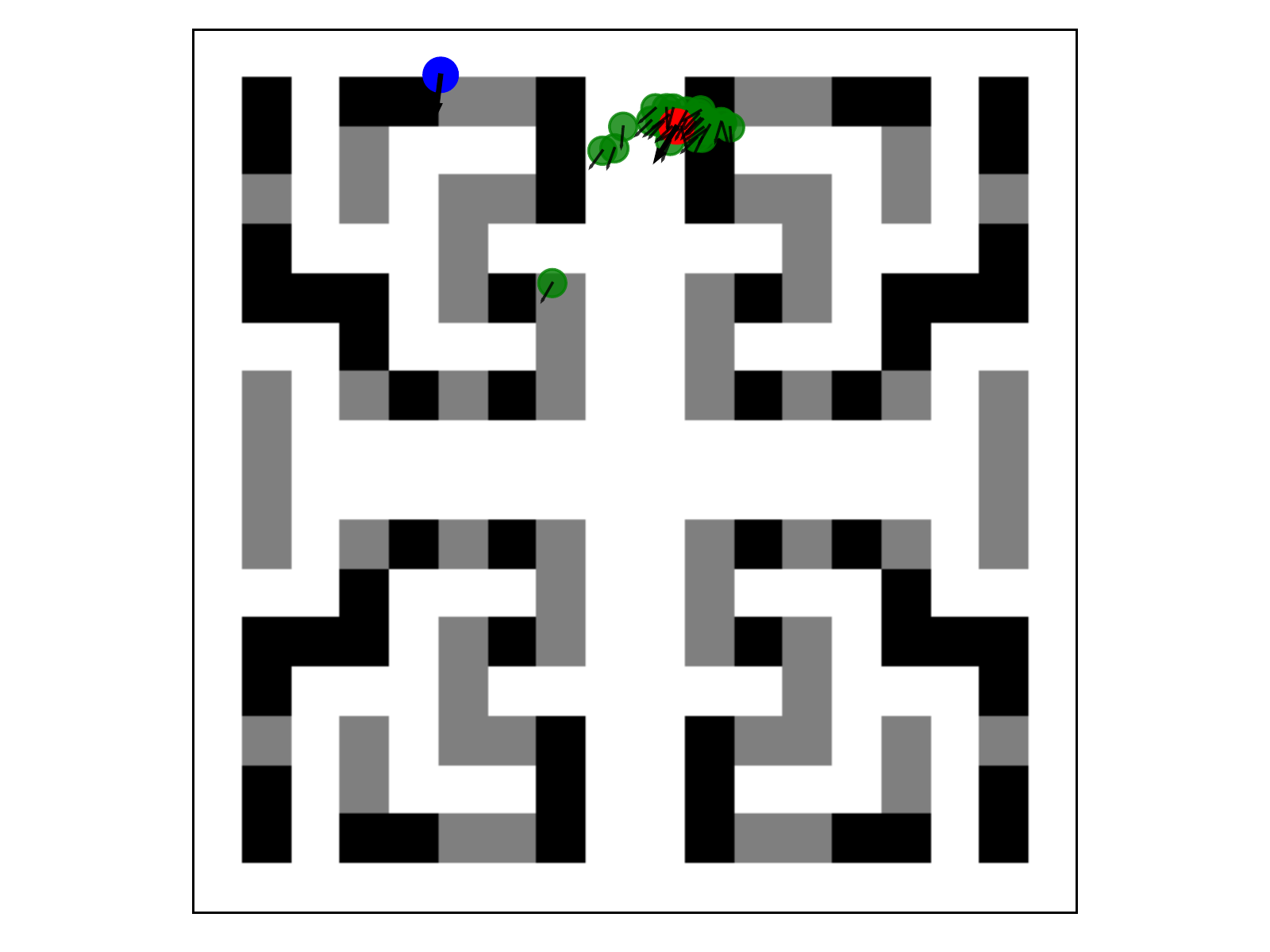} &
		\includegraphics[width=0.18\linewidth]{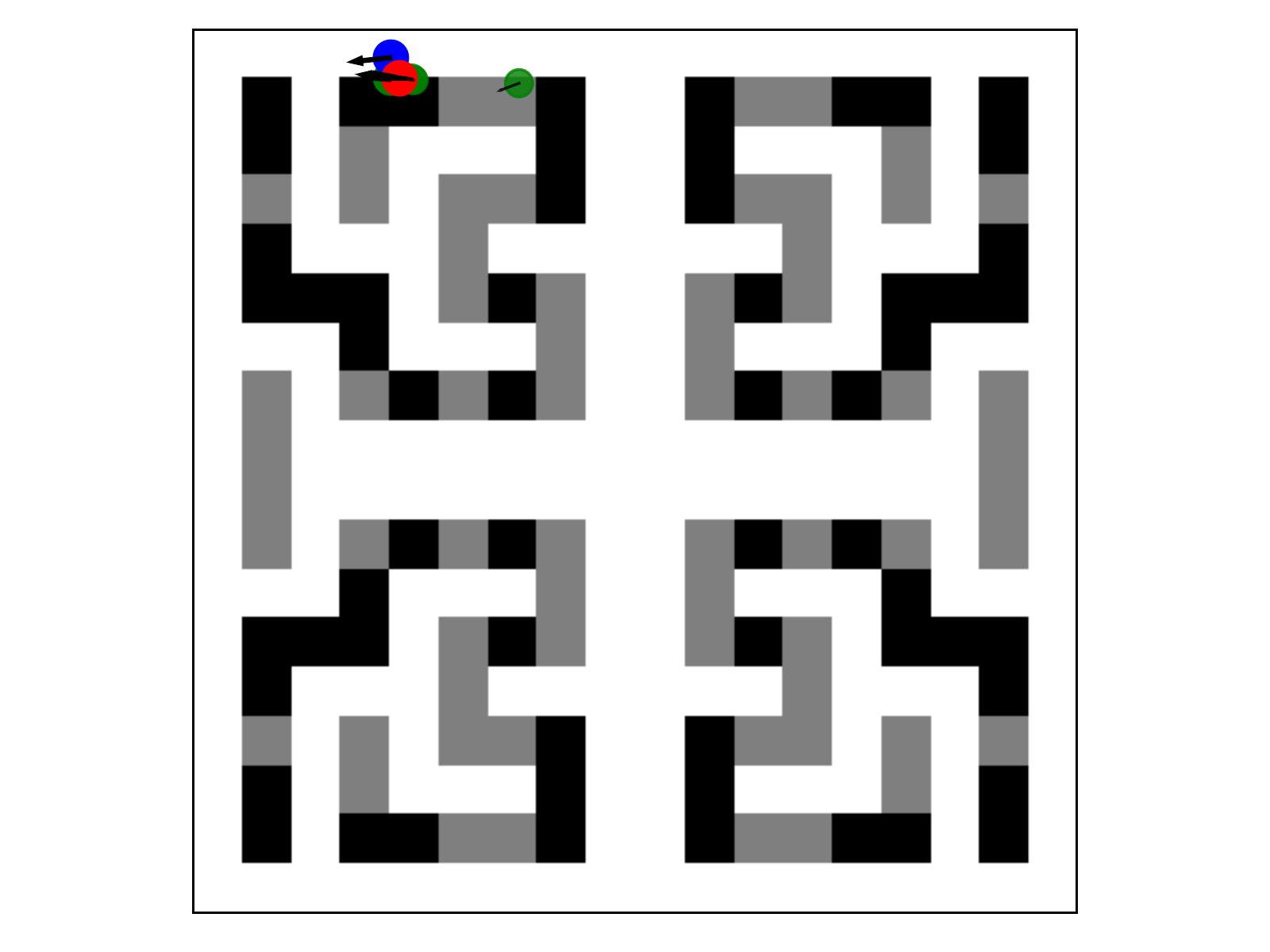} \\
		$t=0$ & $t=2$ & $t=4$ & $t=10$ & $t=20$
	\end{tabular}
\end{figure*}
\begin{figure*}[t]
	\centering
	\begin{tabular}{ccccc}
		\includegraphics[width=0.18\linewidth]{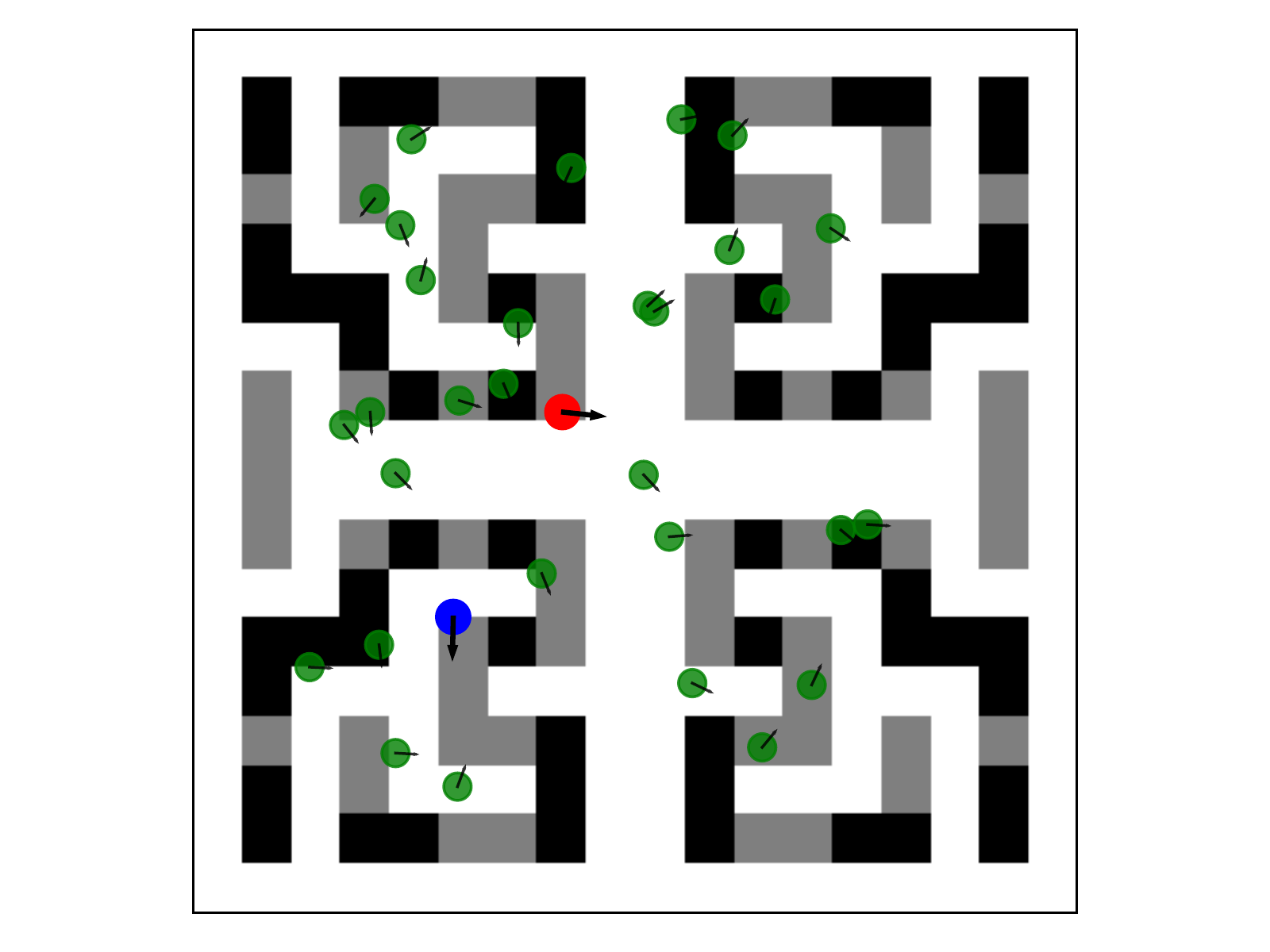} &
		\includegraphics[width=0.18\linewidth]{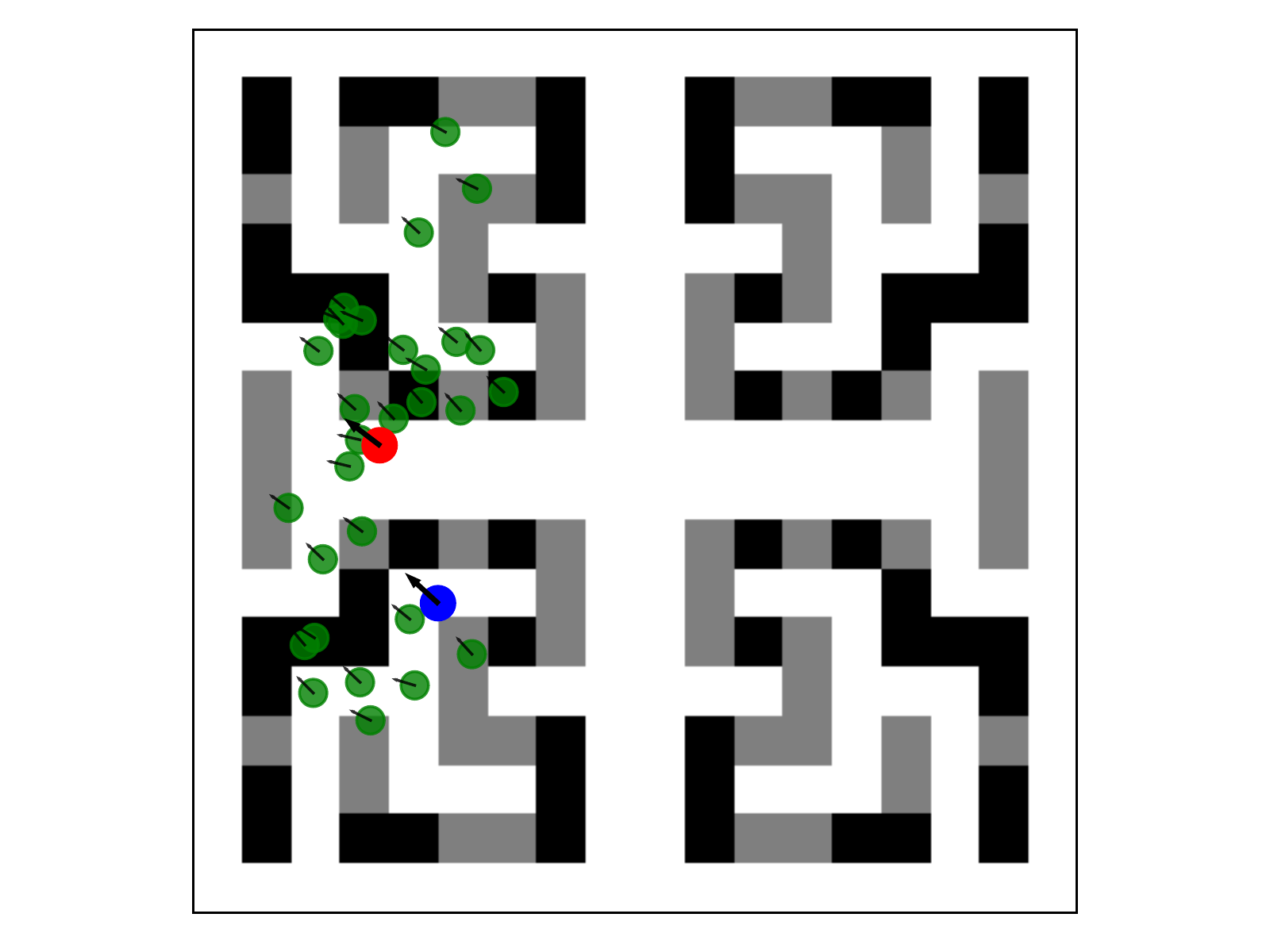} &
		\includegraphics[width=0.18\linewidth]{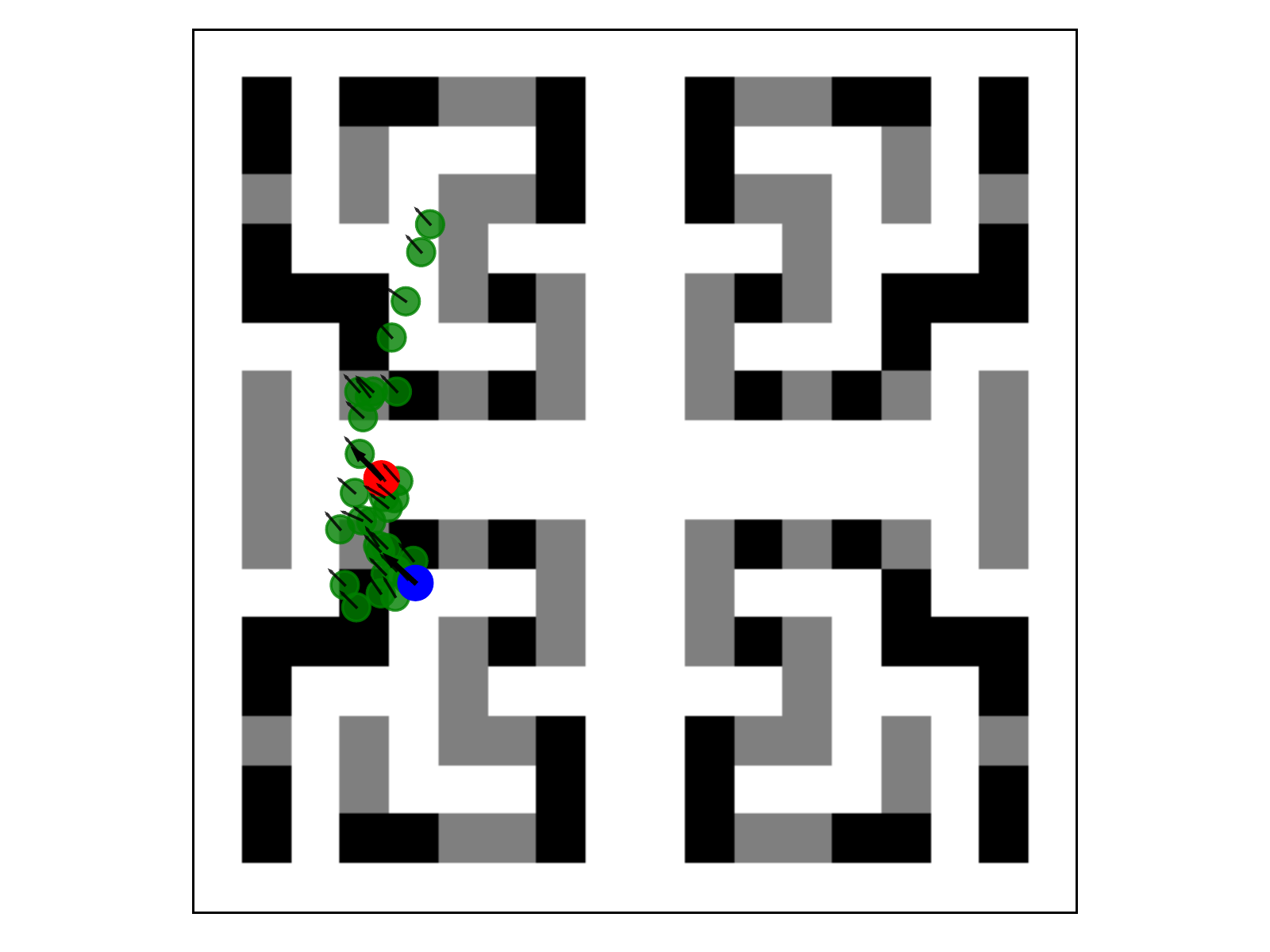} &
		\includegraphics[width=0.18\linewidth]{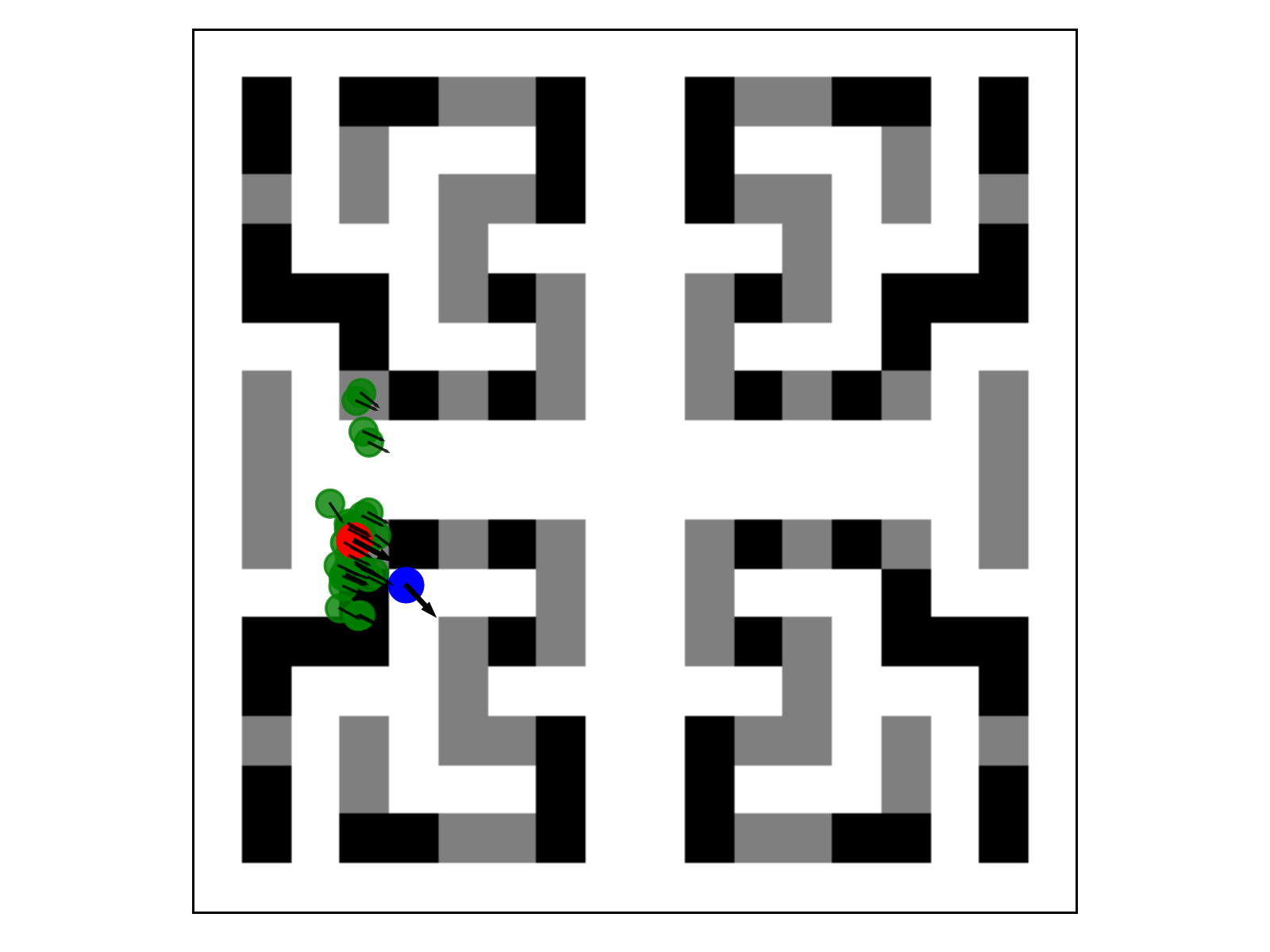} &
		\includegraphics[width=0.18\linewidth]{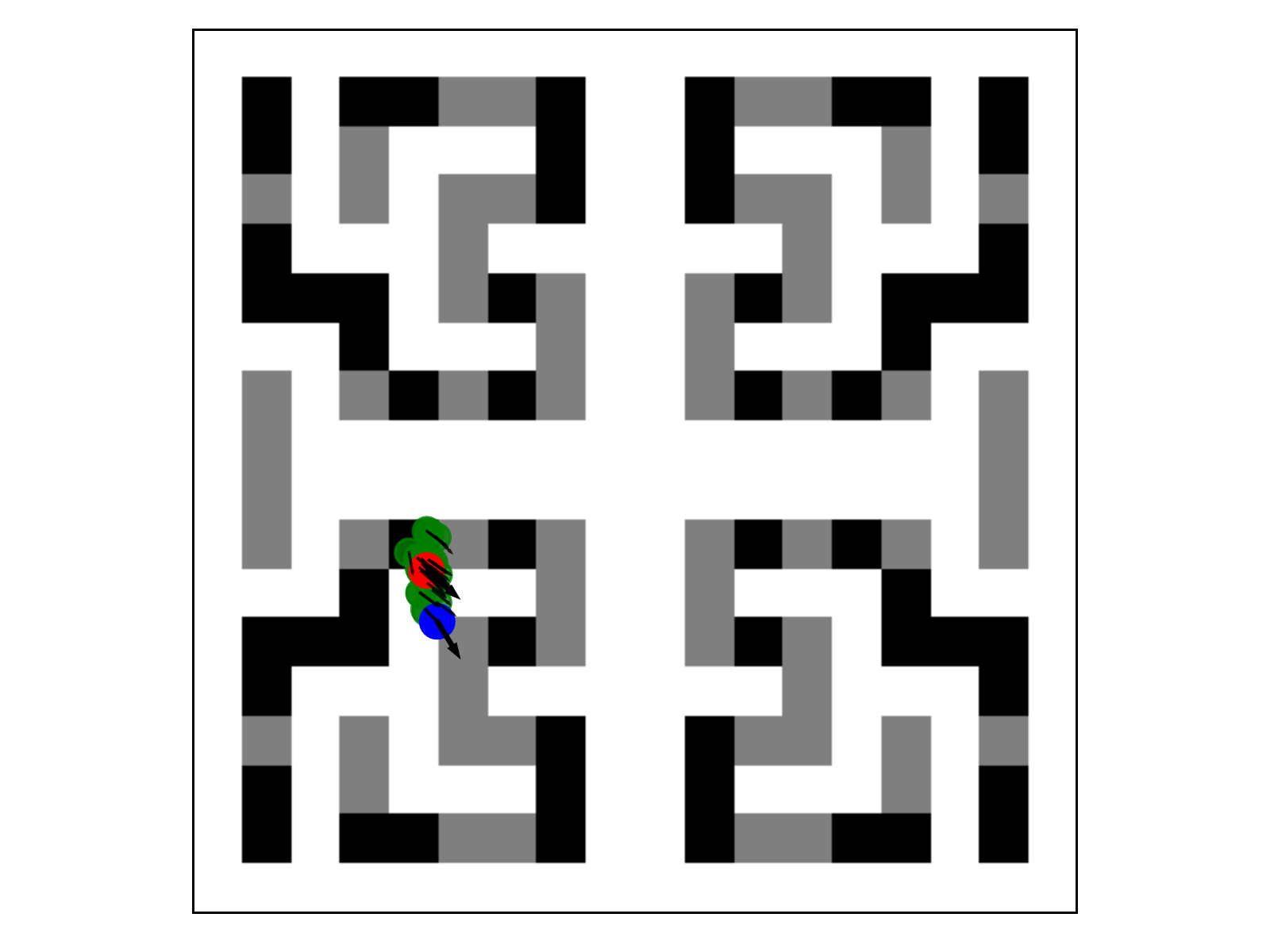} \\
		$t=0$ & $t=2$ & $t=5$ & $t=10$ & $t=15$
	\end{tabular}
\end{figure*}
\begin{figure*}[t]
	\centering
	\begin{tabular}{ccccc}
		\includegraphics[width=0.18\linewidth]{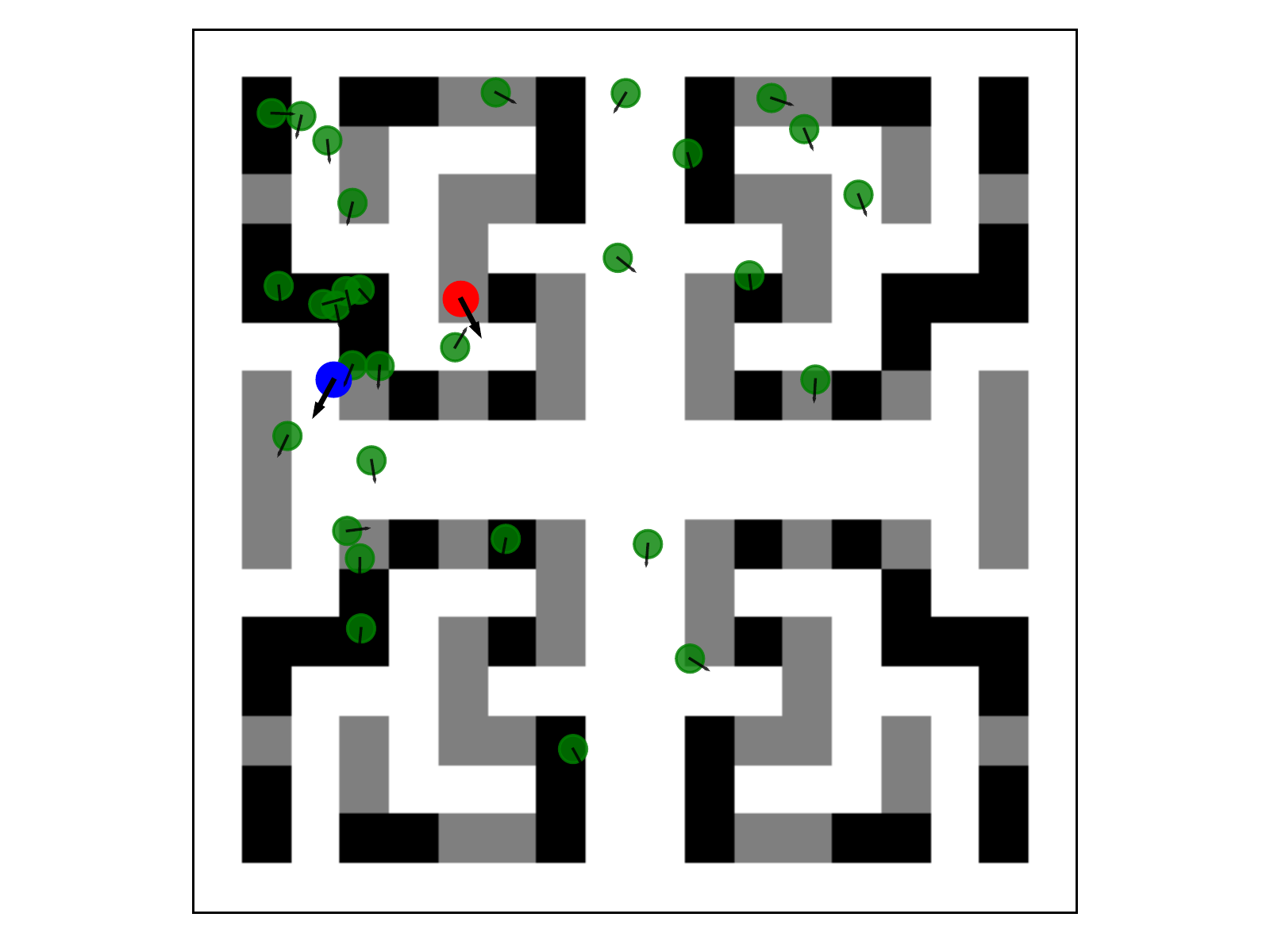} &
		\includegraphics[width=0.18\linewidth]{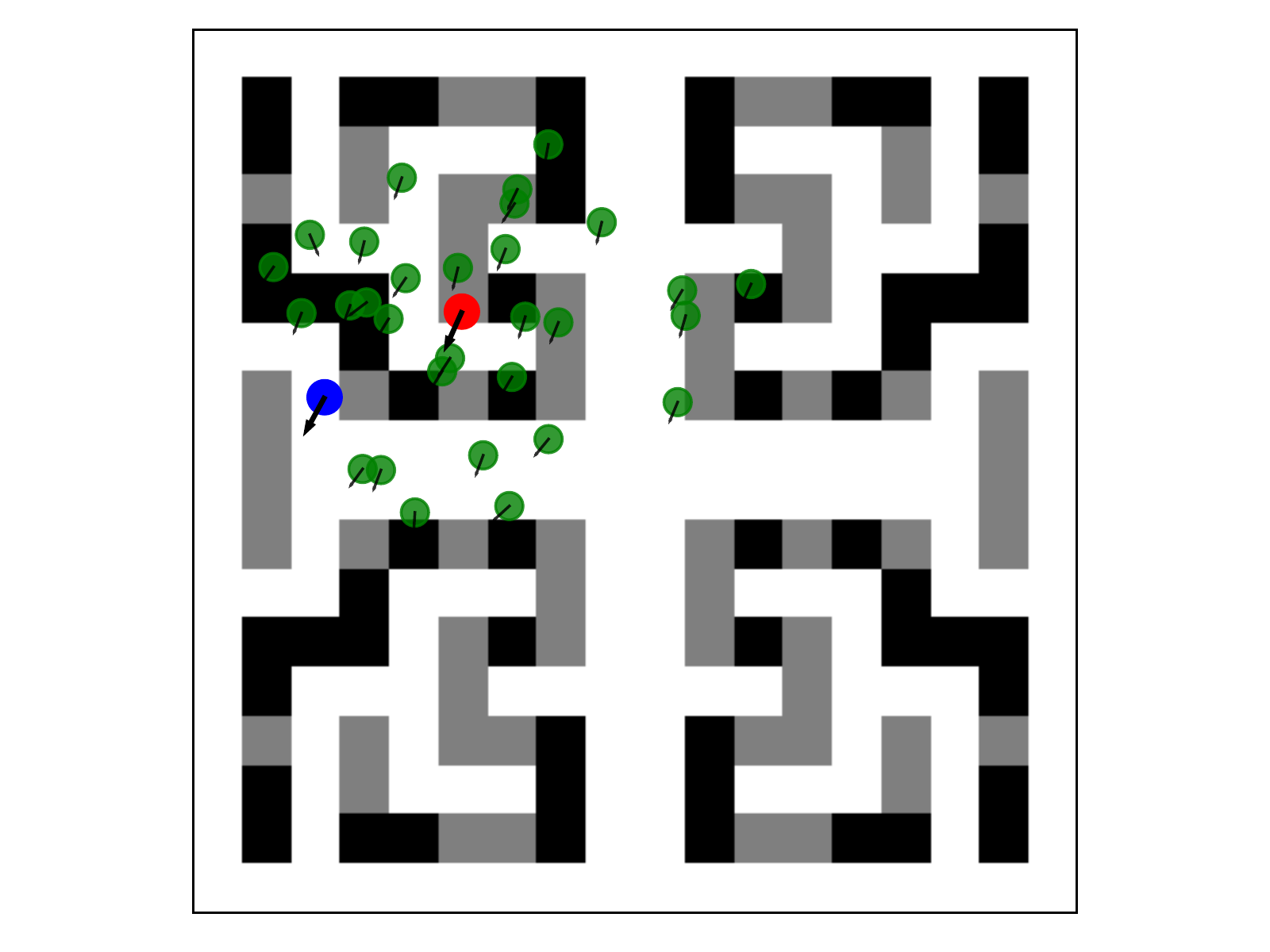} &
		\includegraphics[width=0.18\linewidth]{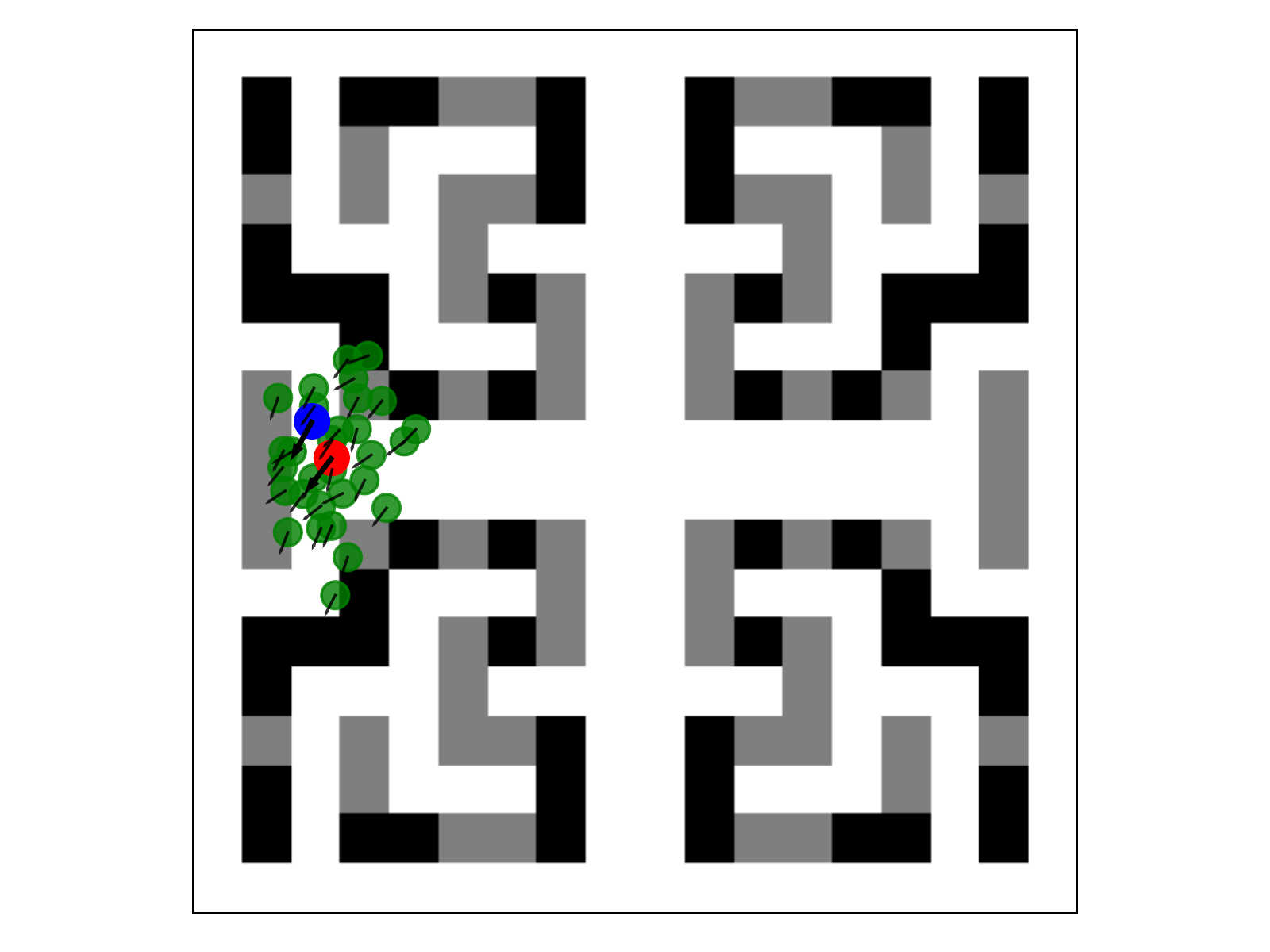} &
		\includegraphics[width=0.18\linewidth]{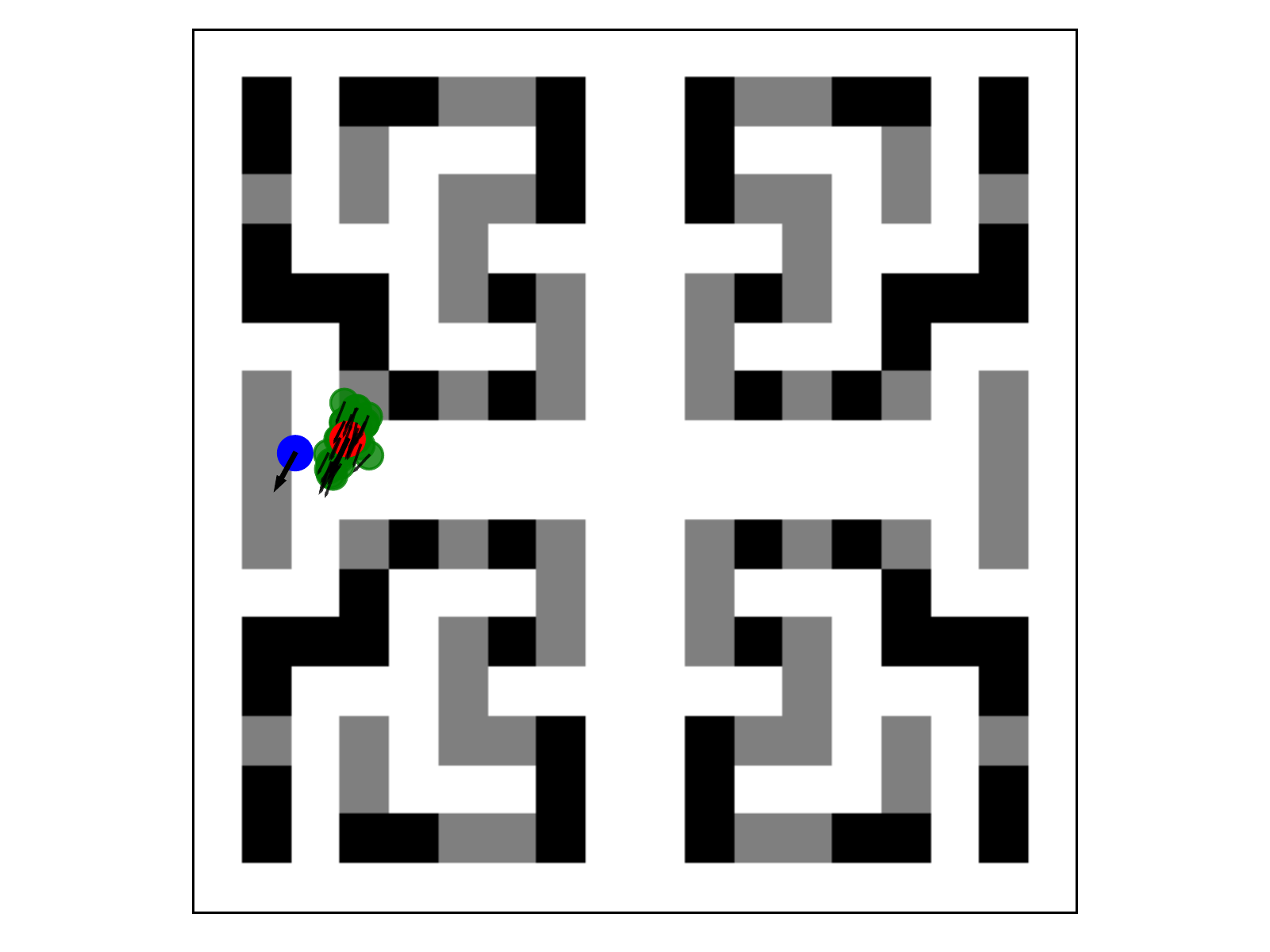} &
		\includegraphics[width=0.18\linewidth]{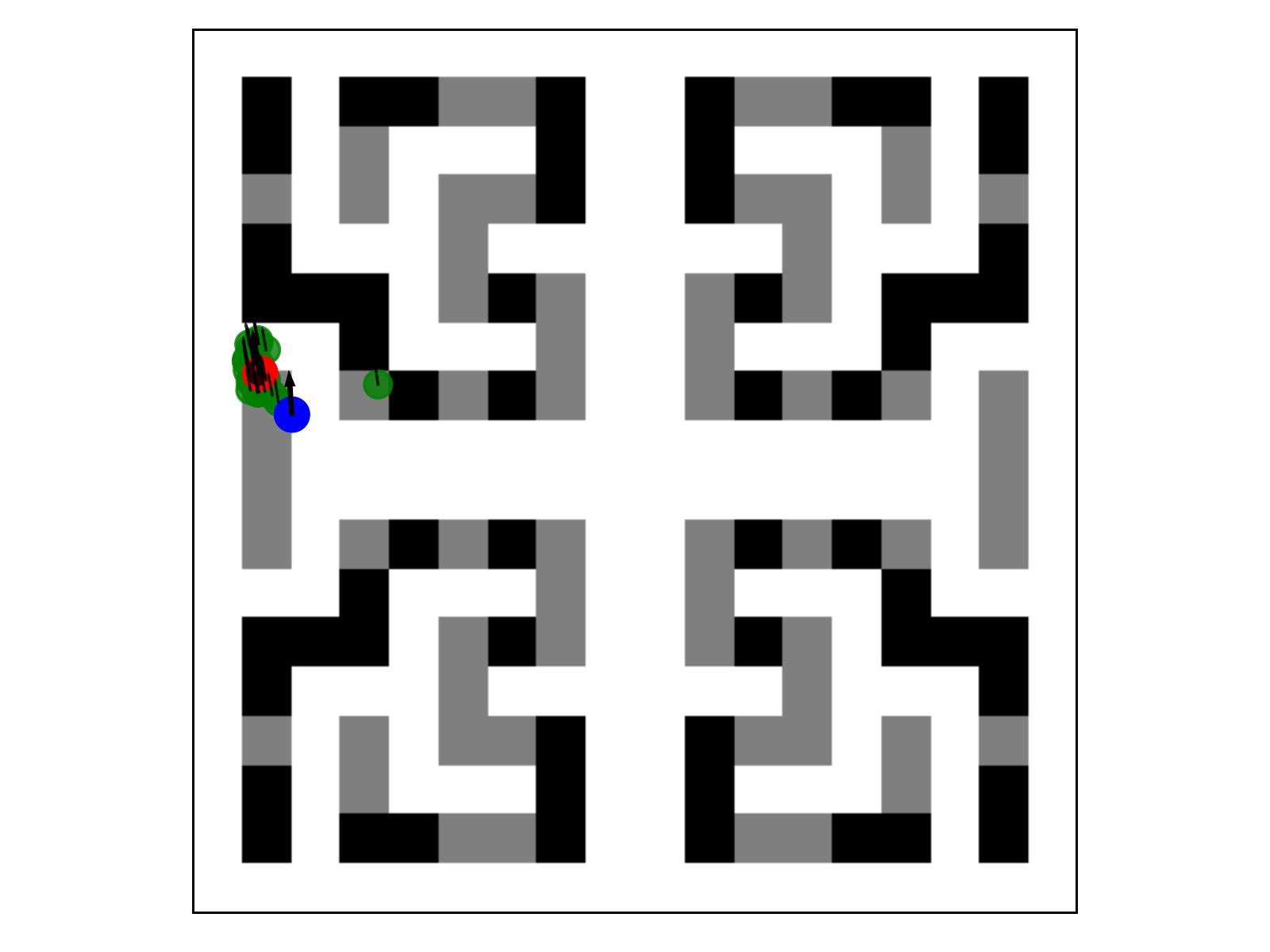} \\
		$t=0$ & $t=2$ & $t=5$ & $t=9$ & $t=13$
	\end{tabular}
\end{figure*}
\begin{figure*}[t]
	\centering
	\begin{tabular}{ccccc}
		\includegraphics[width=0.18\linewidth]{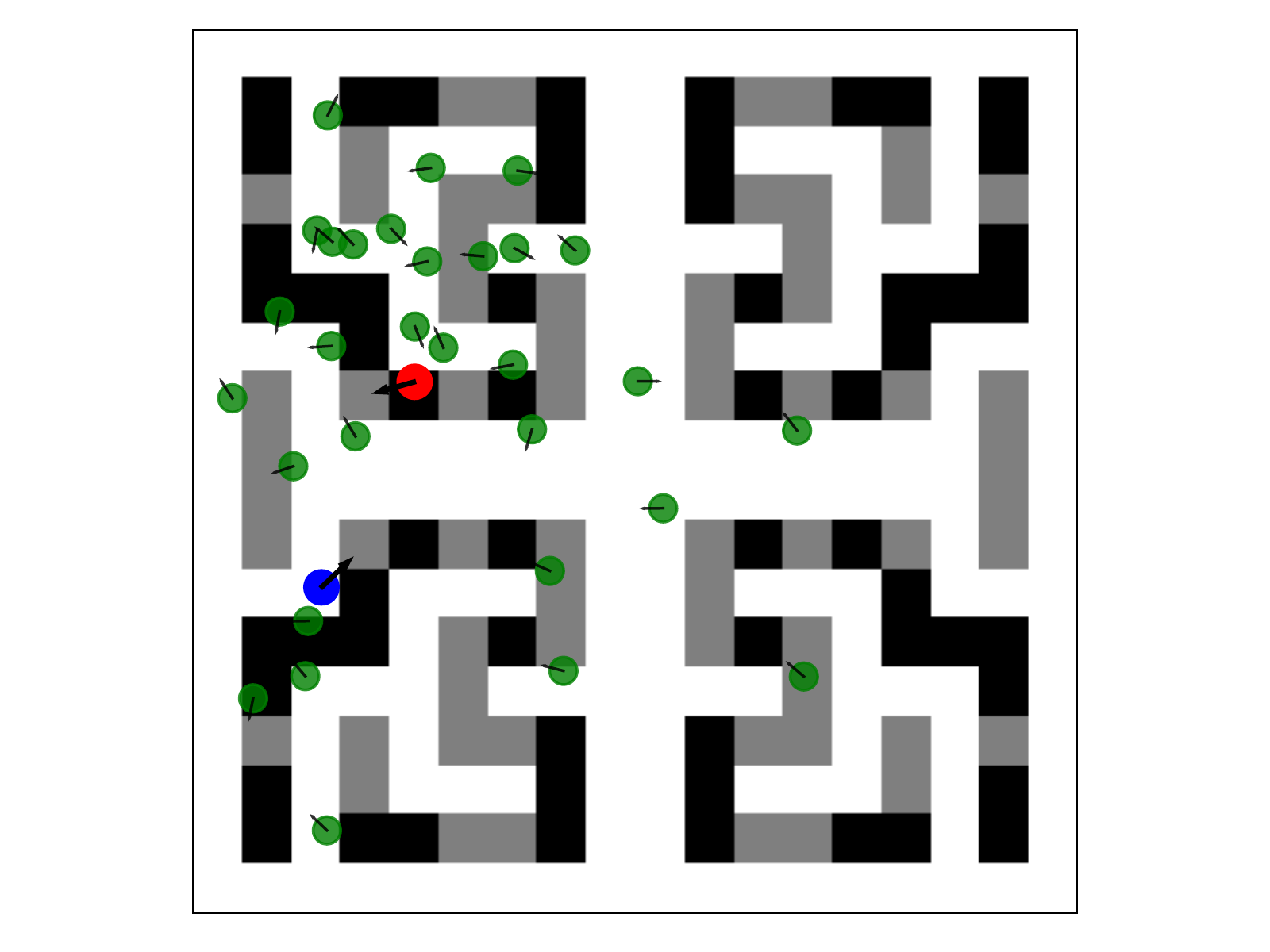} &
		\includegraphics[width=0.18\linewidth]{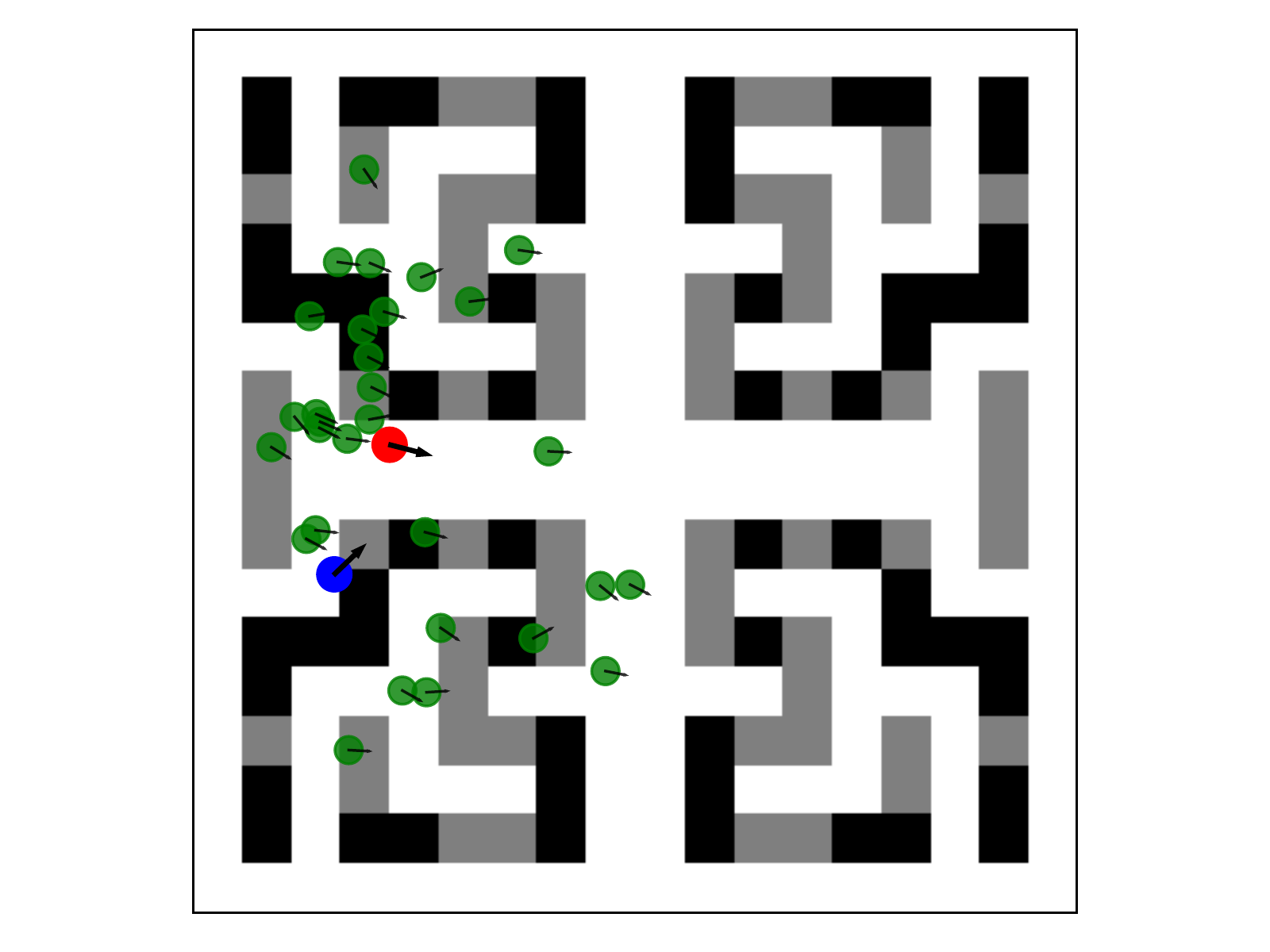} &
		\includegraphics[width=0.18\linewidth]{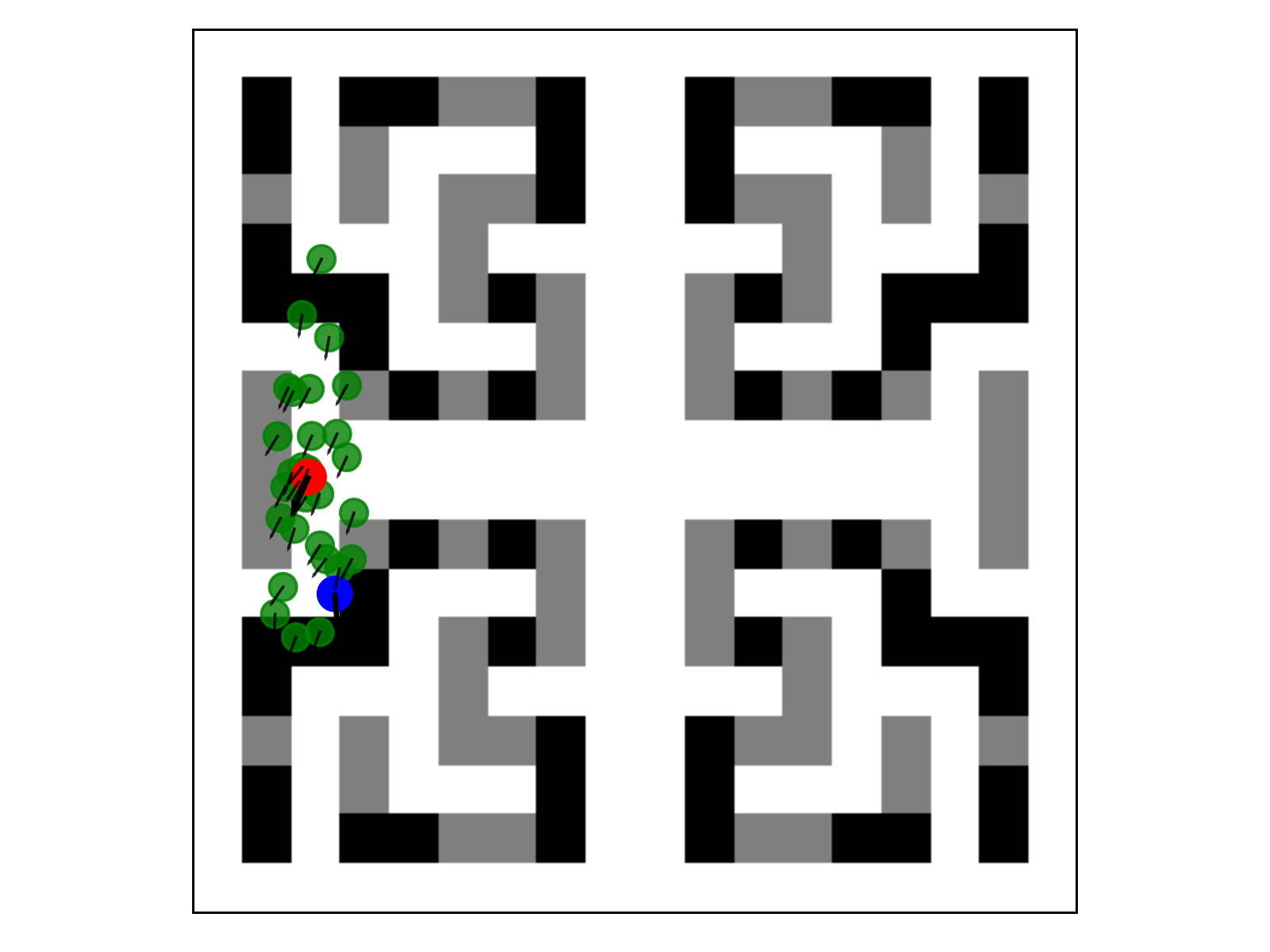} &
		\includegraphics[width=0.18\linewidth]{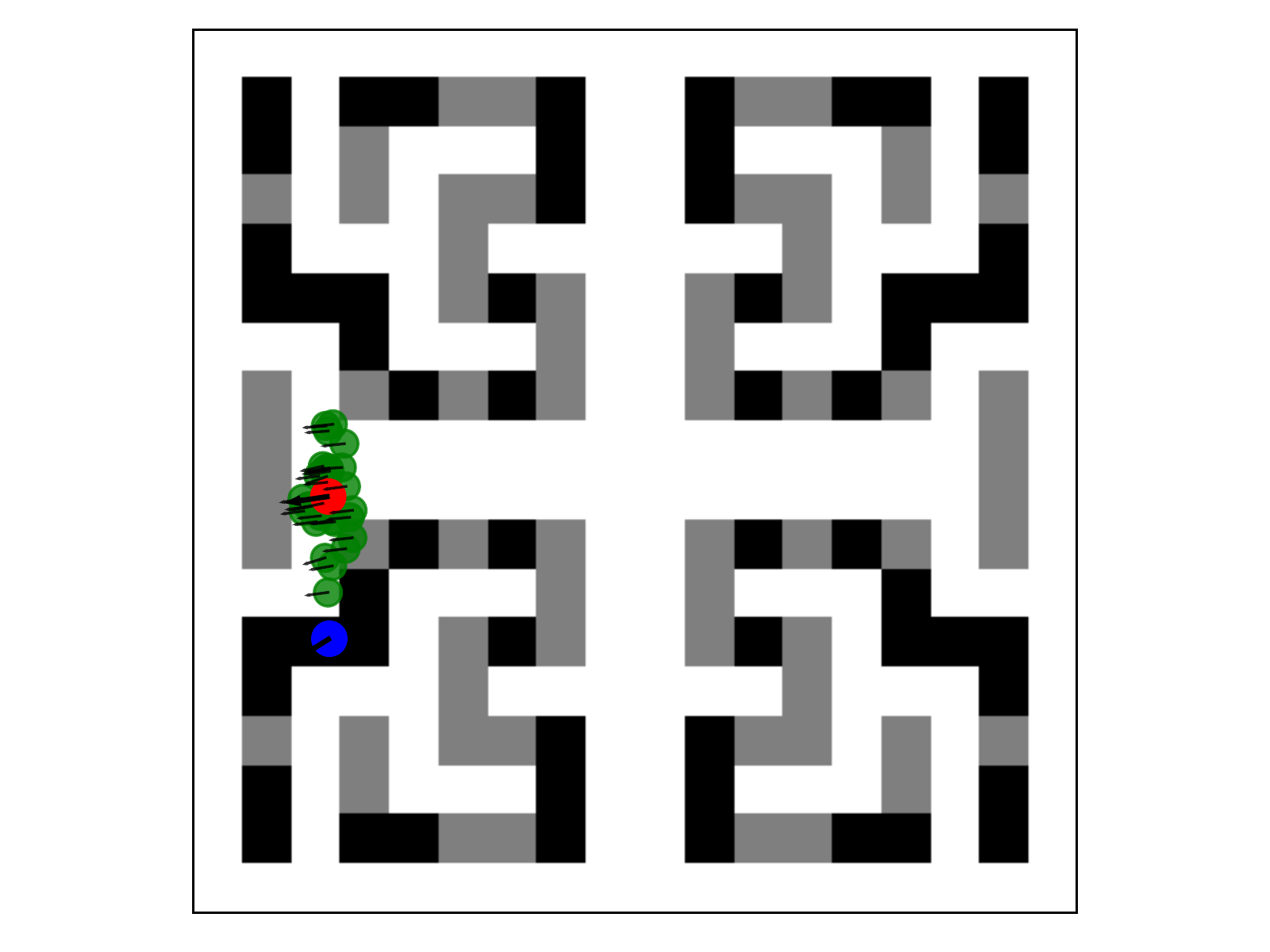} &
		\includegraphics[width=0.18\linewidth]{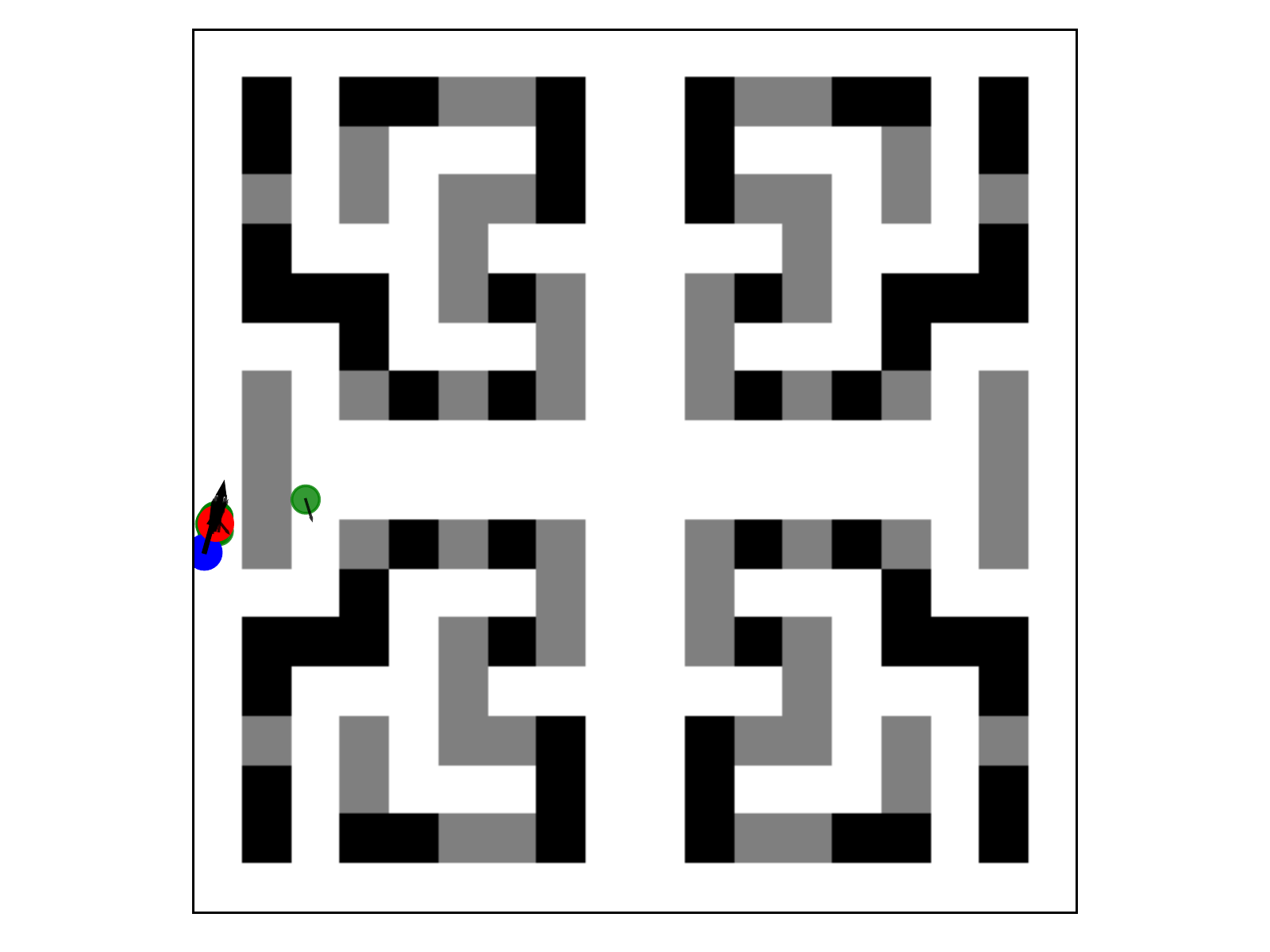} \\
		$t=0$ & $t=2$ & $t=4$ & $t=9$ & $t=15$
	\end{tabular}
\end{figure*}
\begin{figure*}[t]
	\centering
	\begin{tabular}{ccccc}
		\includegraphics[width=0.18\linewidth]{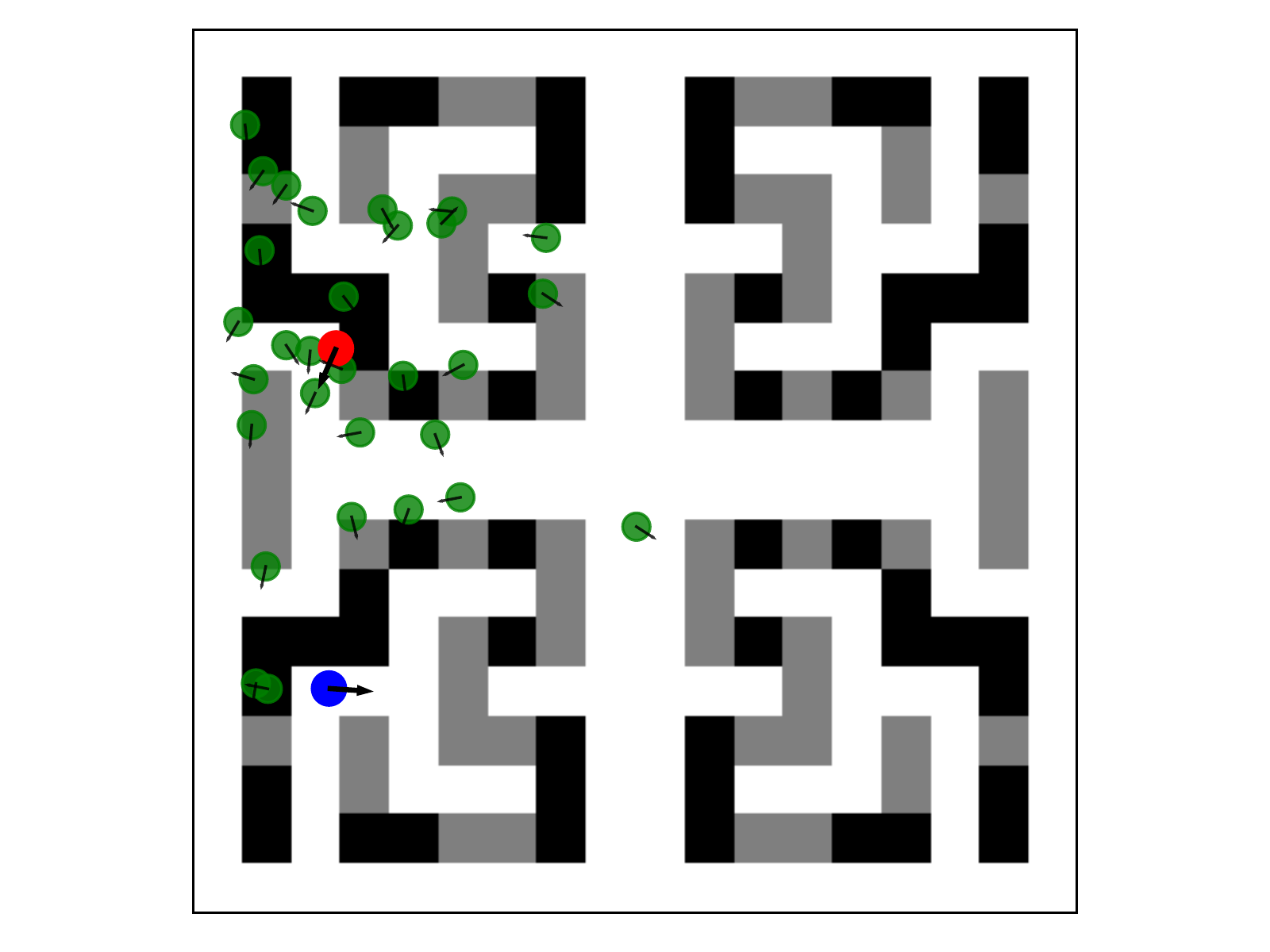} &
		\includegraphics[width=0.18\linewidth]{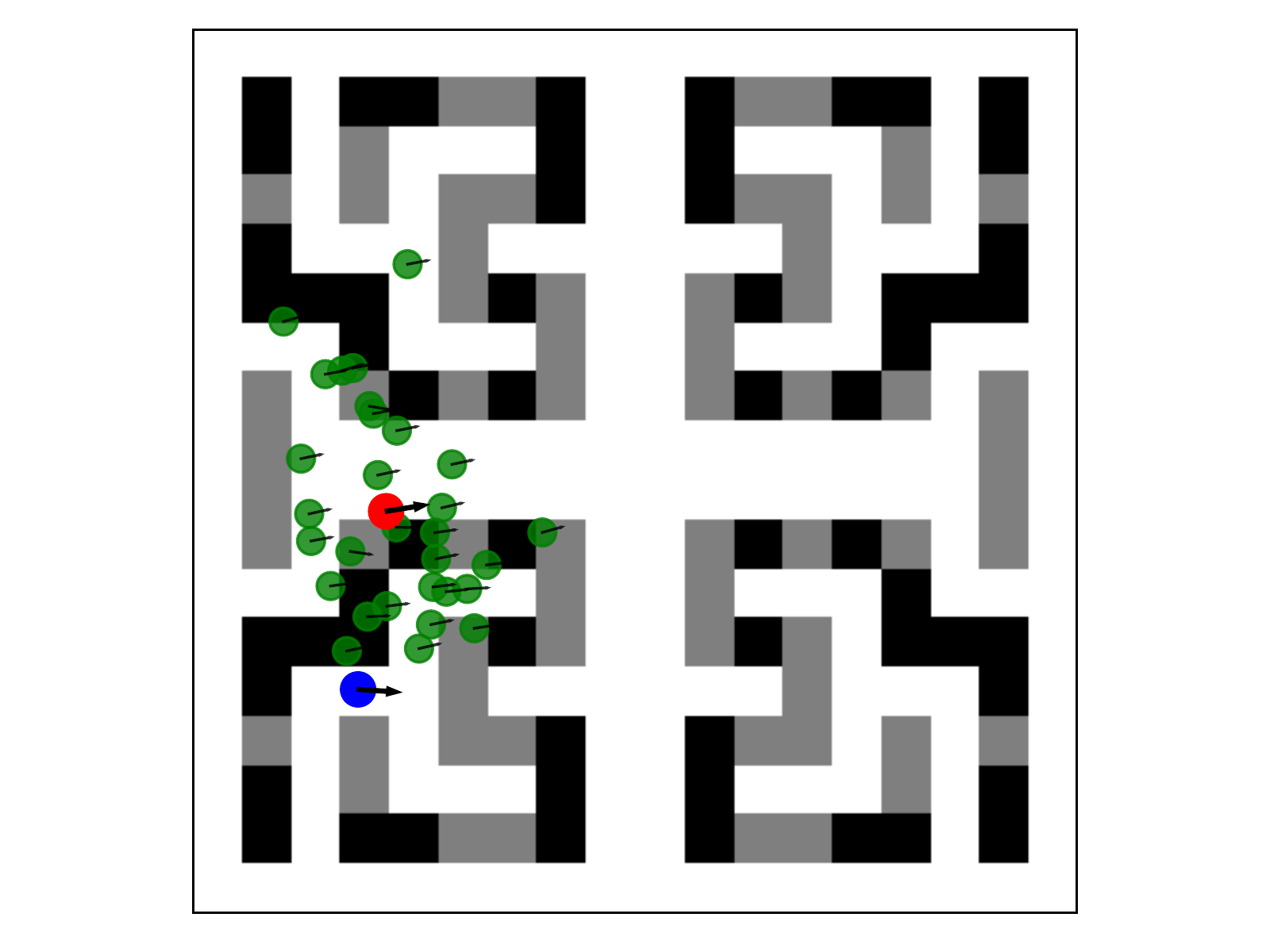} &
		\includegraphics[width=0.18\linewidth]{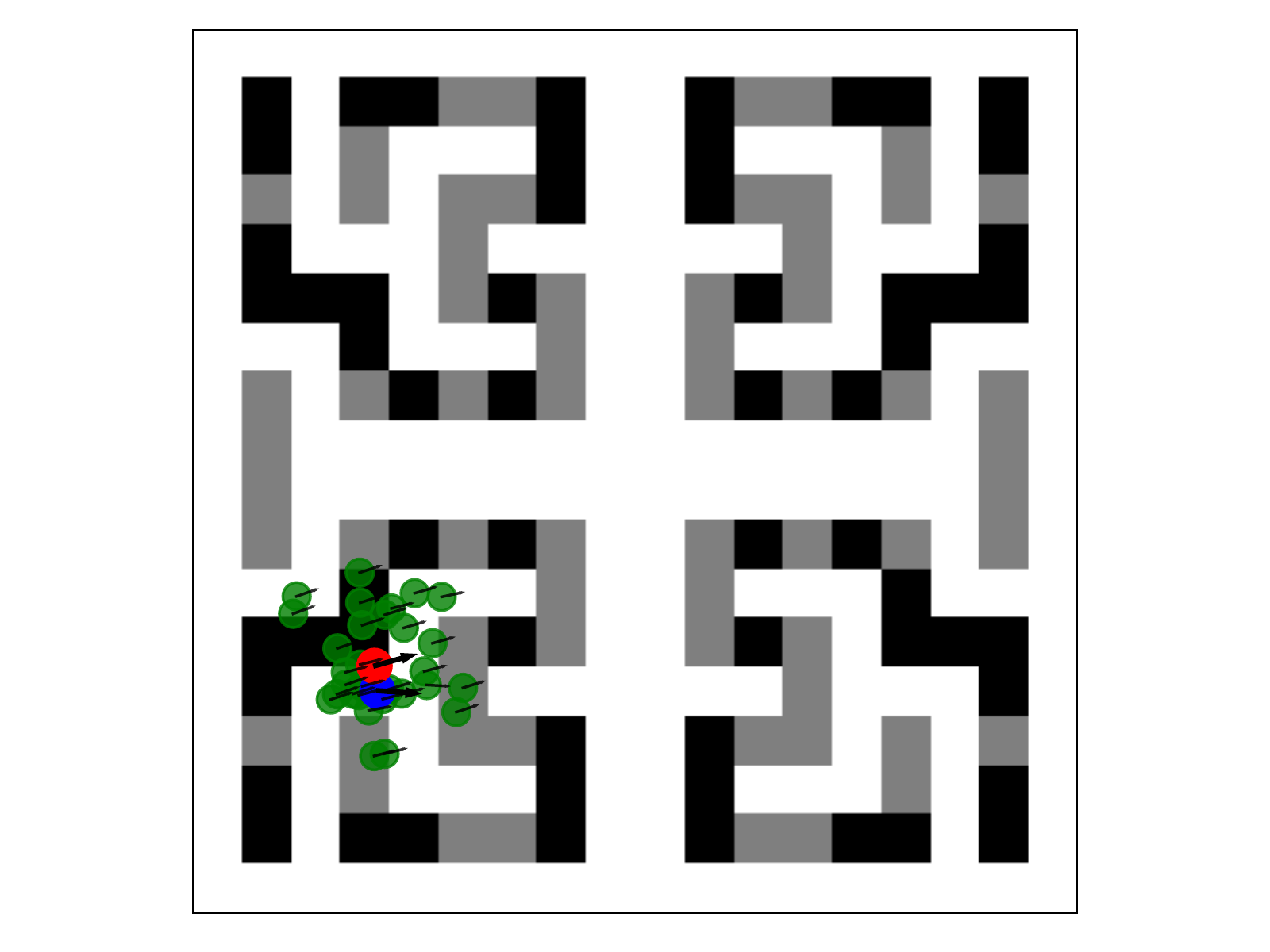} &
		\includegraphics[width=0.18\linewidth]{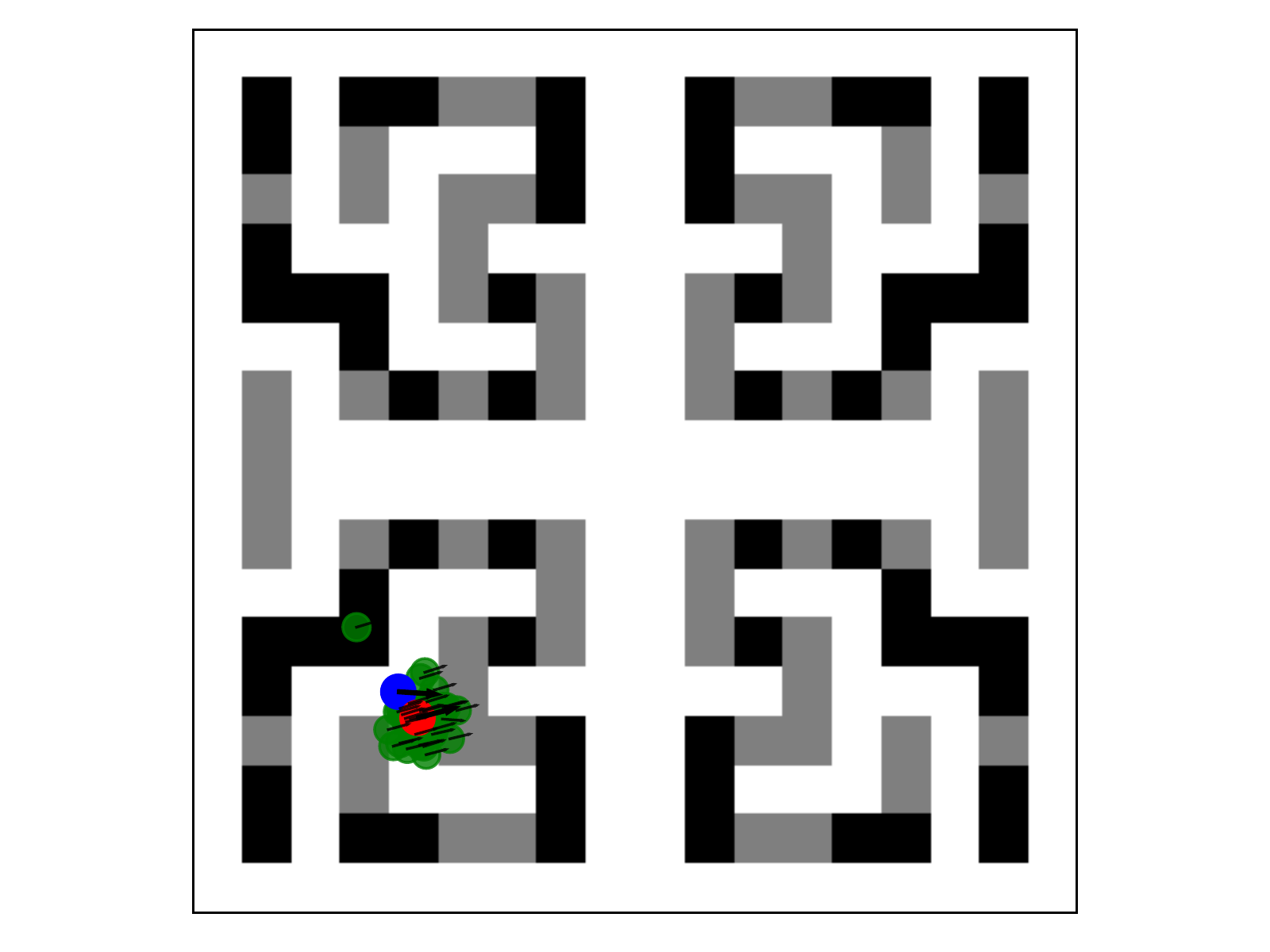} &
		\includegraphics[width=0.18\linewidth]{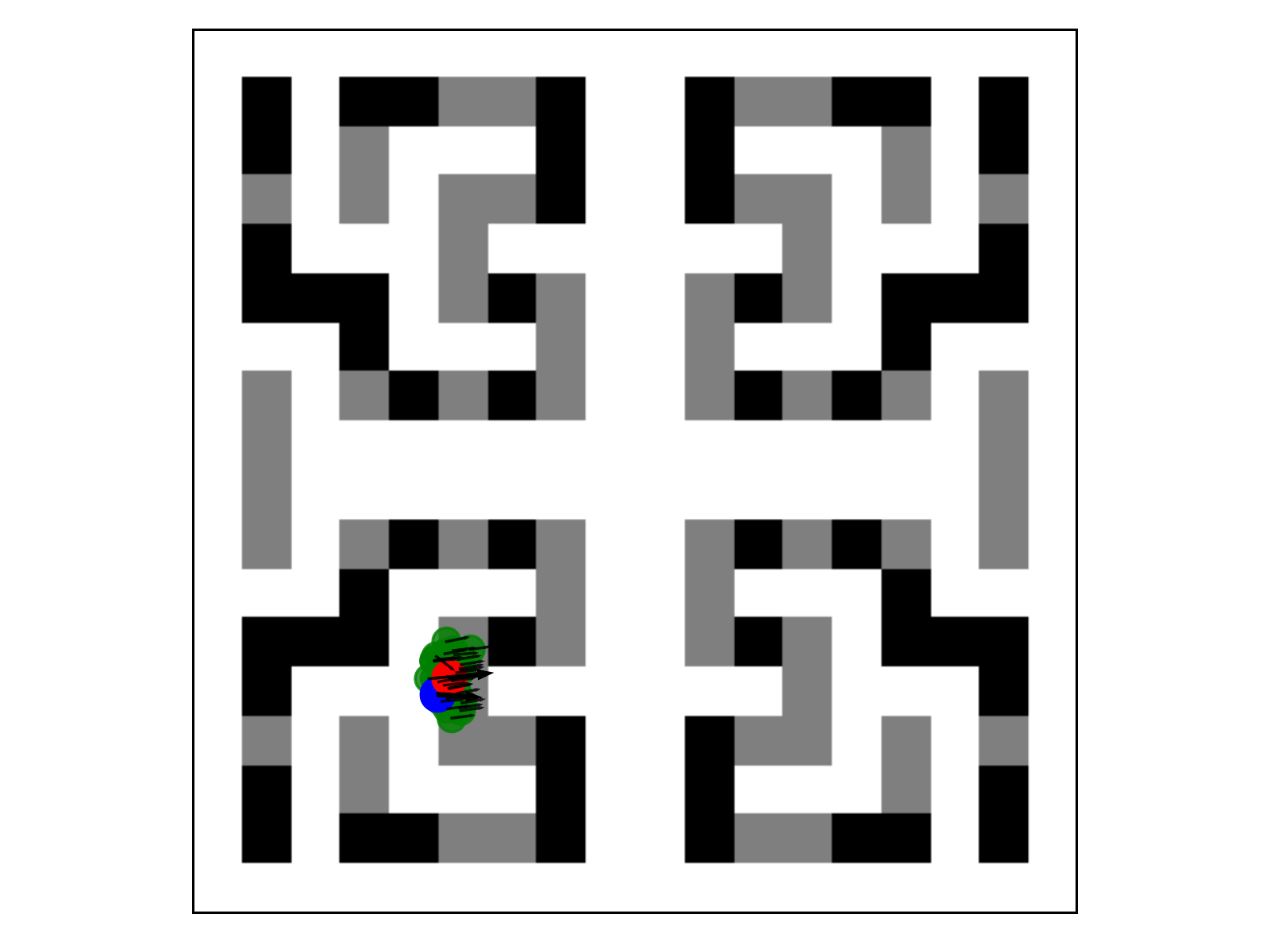} \\
		$t=0$ & $t=3$ & $t=5$ & $t=7$ & $t=11$
	\end{tabular}
\end{figure*}

\end{document}